\documentclass[lettersize,journal]{IEEEtran}
\usepackage{amsmath,amsfonts}
\usepackage{algorithm}
\usepackage{array}
\usepackage[caption=false,font=normalsize,labelfont=sf,textfont=sf]{subfig}
\usepackage{textcomp}
\usepackage{stfloats}

\usepackage{verbatim}
\usepackage{graphicx}
\usepackage[dvipsnames]{xcolor}

\usepackage{amssymb}
\usepackage{booktabs}
\usepackage{multirow}
\usepackage{bm}
\usepackage{enumitem}
\usepackage{colortbl}
\definecolor{mygray}{gray}{.9}
\usepackage{mathtools}
\usepackage{xcolor}

\usepackage{algpseudocode}
\usepackage{tabularx}
\usepackage{url}
\usepackage{color,soul}
\usepackage{epsfig}

\usepackage{subcaption}
\usepackage{multirow}
\usepackage{makecell}
\usepackage{pifont}
\usepackage[switch]{lineno}
\usepackage[misc]{ifsym}

\usepackage[colorlinks,citecolor=blue,urlcolor=blue,linkcolor=blue]{hyperref}
\usepackage{cleveref}
\usepackage{url}
\usepackage[dvipsnames]{xcolor}
\usepackage{tabularx}
\usepackage{multicol}
\usepackage{subcaption}
\usepackage{subfig}

\definecolor{shapecolor}{rgb}{0.0,0.5,0.0}

\definecolor{mygray}{HTML}{E9F1F6}
\usepackage{cite}

\usepackage[most]{tcolorbox}
\tcbuselibrary{theorems}

\newtcbtheorem[auto counter, number within = section]{cmt}{Comment}{
	colbacktitle = black!60!white, colframe = black!60!white,
	colback = black!5!white,
	fonttitle=\bfseries,
}{t}

\newtcolorbox{cmtbox}[1][]{
    colbacktitle = black!60!white,
    colframe = black!60!white,
    colback = black!5!white,
    fonttitle = \bfseries,
    title = Comment,
    #1
}

\title{Mamba-FSCIL: Dynamic Adaptation with Selective State Space Model for Few-Shot Class Incremental Learning}

\author{
Xiaojie Li, \quad
Yibo Yang, \quad
Jianlong Wu,~\IEEEmembership{Member,~IEEE,} \quad
Yue Yu,~\IEEEmembership{Member,~IEEE,} \\
Ming-Hsuan Yang,~\IEEEmembership{Fellow,~IEEE,} \quad
Liqiang Nie,~\IEEEmembership{Senior Member,~IEEE,} \quad
Min Zhang

\IEEEcompsocitemizethanks{
\IEEEcompsocthanksitem Xiaojie Li, Jianlong Wu, Liqiang Nie, and Min Zhang are with the School of Computer Science and Technology, Harbin Institute of Technology (Shenzhen), Shenzhen, China. E-mail: xiaojieli0903@gmail.com, wujianlong@hit.edu.cn, nieliqiang@gmail.com, zhangmin2021@hit.edu.cn.
\IEEEcompsocthanksitem Xiaojie Li, Jianlong Wu, and Yue Yu are with Pengcheng Laboratory, Shenzhen, China. E-mail: yuy@pcl.ac.cn.
\IEEEcompsocthanksitem Yibo Yang is with King Abdullah University of Science and Technology, Jeddah, Saudi Arabia. E-mail: yibo.yang93@gmail.com.
\IEEEcompsocthanksitem Ming-Hsuan Yang is with the Department of Computer Science and Engineering, University of California, Merced, CA, USA. E-mail: myang37@ucmerced.edu.
}}

\begin{document}
\IEEEtitleabstractindextext{
\begin{abstract}
Few-shot class-incremental learning (FSCIL) aims to incrementally learn novel classes from limited examples while preserving knowledge of previously learned classes. Existing methods face a critical dilemma: static architectures rely on a constant parameter space to learn from data that arrive sequentially, making them prone to overfitting to the current session, while dynamic architectures continually expand the parameter space, leading to increased complexity. 
In this study, we explore the potential of Selective State Space Models (SSMs) for FSCIL. Mamba leverages its input-dependent parameters to dynamically adjust its processing patterns and generate content-aware scan patterns without session-wise projector expansion. This enables it to configure distinct processing for base and novel classes, helping preserve existing knowledge while adapting to new ones.
To leverage Mamba's potential for FSCIL, we design two key modules:
First, we propose a dual selective SSM projector that generates input-conditioned state-space parameters from intermediate features for dynamic adaptation. The dual design structurally decouples base and novel-class processing, employing a frozen base branch to maintain stable base-class features and a dynamic incremental branch that adaptively learns distinctive feature shifts for novel classes. Second, we develop a class-sensitive selective scan mechanism to guide dynamic adaptation of the incremental branch. It reduces the disruption to base-class representations caused by training on novel data, and meanwhile, encourages the selective scan to perform in distinct patterns between base and novel classes.
Extensive experiments on miniImageNet, CIFAR-100, and CUB-200 demonstrate that Mamba-FSCIL achieves state-of-the-art performance. The code is available at \url{https://github.com/xiaojieli0903/Mamba-FSCIL}.
\end{abstract}

\begin{IEEEkeywords}
Few-Shot Class-Incremental Learning, Selective State Space Models, Stability-Plasticity Balance, Continual Learning
\end{IEEEkeywords}
\vspace{-2mm}
}
    
\maketitle
\IEEEdisplaynontitleabstractindextext
\IEEEpeerreviewmaketitle
\section{Introduction}
\label{sec:intro}
\begin{figure*}
 \centering
 \includegraphics[width=1\linewidth]{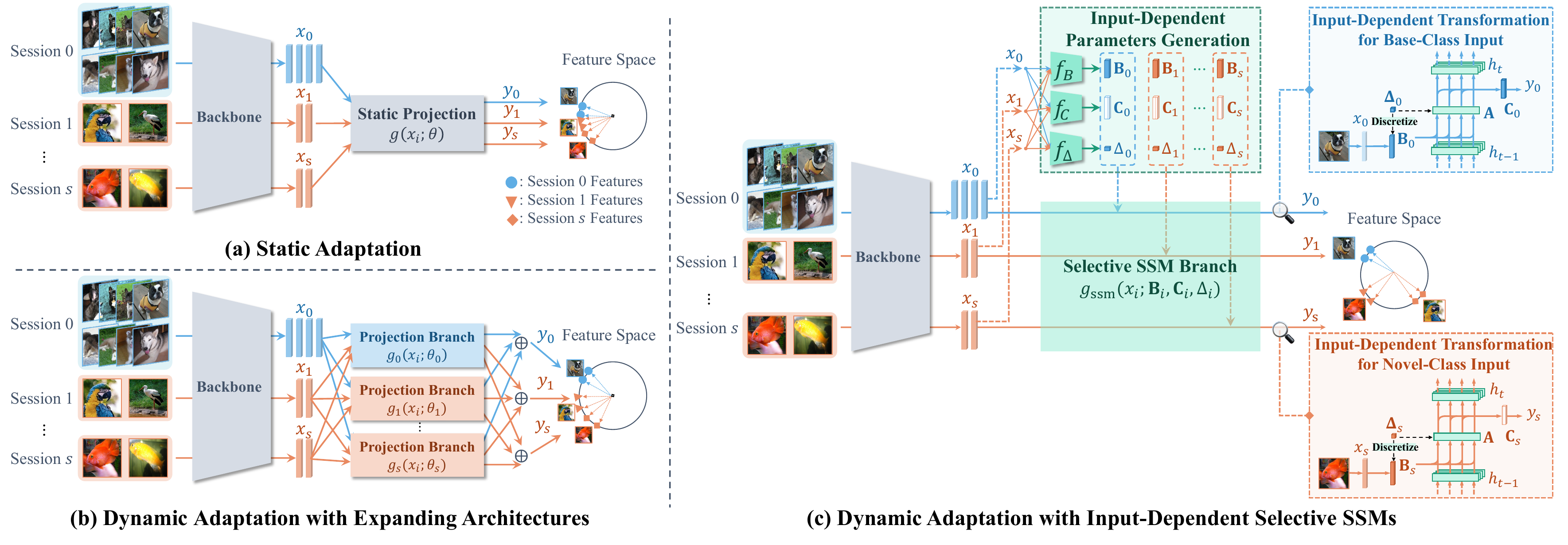}
 \vspace{-6mm}
  \caption{
  \textbf{Motivation and Comparison of Adaptation Paradigms in FSCIL}. \textbf{(a) Static Adaptation:} A single, fixed-capacity projector $g(x_i;\theta)$ is shared across all sessions. Updates for new classes (orange) may overwrite parameters encoding old ones (black), causing feature interference and catastrophic forgetting. \textbf{(b) Dynamic Adaptation with Expanding Architectures:} Each session adds a new projection branch $g_s(x_i;\theta_s)$ while retaining previous ones, resulting in session-wise growth in parameters and inference cost. \textbf{(c) Dynamic Adaptation with Input-Dependent Selective SSM Branch (Ours)}: A Mamba-based projector $g_{\text{ssm}}(\cdot)$ reused across sessions. By generating input-conditioned parameters $(\mathbf{B}_i, \mathbf{C}_i, \mathbf{\Delta}_i)$ for each input $x_i$ via $\mathbf{B}_i=f_B(x_i), \mathbf{C}_i=f_C(x_i), \mathbf{\Delta}_i=f_{\Delta}(x_i)$, this enables distinct transformations for base and novel classes, preserving base-class representations while performing discriminative adaptation for novel classes without session-wise parameter expansion.}
 \label{fig:motivation}
 \vspace{-4mm}
\end{figure*}

\IEEEPARstart{F}{ew}-Shot Class-Incremental Learning (FSCIL)~\cite{tao2020few,zhang2021few,zhou2022forward} addresses the challenge of continuously learning new concepts from limited data without forgetting previously acquired knowledge. This capability is critical for real-world applications, such as robotics and autonomous systems, where data arrive incrementally and new categories contain only a few examples. In FSCIL, a model first learns a set of base classes with sufficient data and then adapts to novel classes with limited samples, typically without storing or replaying raw image exemplars from previous sessions. The key challenge is to maintain high performance across all learned classes by balancing \textbf{plasticity} (learning new knowledge) and \textbf{stability} (retaining old knowledge)~\cite{zhang2023few,tian2024survey,wang2024comprehensive}.

Existing FSCIL methods typically follow two main paradigms.
\textbf{Static adaptation} paradigm~\cite{peng2022few,hersche2022constrained,yang2023neural_arxiv} employs a fixed-capacity projector shared across all sessions. As shown in Fig.~\ref{fig:motivation}~(a), a multi-layer perceptron (MLP) projector is used to map backbone features to classifier space. Since the shared weights are updated repeatedly without protecting established representations, learning new classes may overwrite the parameters encoding base knowledge, leading to feature interference and catastrophic forgetting.
\textbf{Dynamic adaptation with expanding architectures} paradigm~\cite{tao2020few,zhang2021few,yoon2020xtarnet,yang2021learnable,yan2021dynamically,yang2022dynamic,wang2022foster,park2024pre}, shown in Fig.~\ref{fig:motivation}~(b), allocates new parameters (e.g., additional branches or modules) for each incremental session while retaining those learned in previous sessions. Although effective in preserving stability, this paradigm introduces session-wise growth in model size and computational cost, limiting scalability in long-term learning or resource-constrained environments.
This presents a critical question: \textbf{how to dynamically adapt to new classes and preserve old knowledge without session-wise parameter expansion?}

To answer this question, we explore a new paradigm: \textbf{dynamic computation within a fixed-capacity architecture}. Instead of expanding model parameters, we exploit input-dependent processing to make a fixed-size model adapt its behavior based on the sample it receives. The objective is to realize class-sensitive adaptation without session-wise parameter expansion: performing stable transformations for base classes while generating adaptive, discriminative transformations for novel classes according to their input context.
Selective State Space Models (Mamba)~\cite{gu2023mamba} provide a promising foundation for this paradigm. Mamba’s selective scan mechanism enables \textbf{input-dependent dynamic computation} by generating core parameters $(\mathbf{B}_i, \mathbf{C}_i, \mathbf{\Delta}_i)$ \textbf{on the fly}, conditioned on each input $x_i$. 
This mechanism allows a Mamba module to perform input-specific transformations, providing a basis for processing base and novel inputs differently without session-wise parameter expansion (Fig.~\ref{fig:motivation}~(c)). Additionally, Vision Mamba’s~\cite{liu2024vmamba,zhu2024vision,li2024videomamba,xing2024segmamba,ma2024u,han2024demystify} mechanism of generating distinct parameters for each visual token enables fine-grained selectivity. This capability is particularly beneficial when novel classes have limited training samples, where models are prone to overfitting on environmental noise. This token-level selectivity can help emphasize discriminative foreground patches and suppress irrelevant background information, supporting more robust feature extraction for novel classes under scarce-data conditions.

However, bridging Mamba's capabilities with FSCIL's requirements presents two challenges:
\textbf{(1) The structural challenge: preserving base-class knowledge.}
FSCIL requires base-class knowledge to be preserved across incremental sessions to reduce catastrophic forgetting. Although Mamba inherently performs dynamic, input-dependent computation, applying it to FSCIL requires an explicit mechanism to reduce interference between base and novel class processing. We need to harness Mamba's dynamic nature within a framework of explicit structural decoupling, so that the plasticity required for novel classes can be leveraged while reducing perturbation to established base-class representations.
\textbf{(2) The regularization challenge: enforcing explicit class-sensitivity.}
FSCIL demands a mechanism that actively suppresses changes for established classes while promoting distinctive adaptation for novel ones. While Mamba naturally varies its parameters based on input context, this intrinsic mechanism is \textit{content-aware} but not necessarily class-sensitive. To fully harness this potential, we must elevate implicit input-dependence to explicit class-sensitive behavior, encouraging the generated parameters for novel classes to be discriminative while reducing interference with base-class representations.

To address these challenges, we propose \textbf{Mamba-FSCIL}, the first framework that leverages Mamba’s input-dependent dynamic computation to effectively balance stability and plasticity within a fixed-capacity model. Our approach introduces two synergistic innovations:

First, to address the \textbf{structural challenge}, we introduce the \textbf{dual selective SSM projector}. This innovation decouples the projection architecture into two Mamba-based branches to structurally support class-sensitive processing.
The \textbf{base branch} ($g_{\text{base}}$) serves as a stability anchor. It is trained only in the base session and then frozen to preserve foundational representations. Unlike the static frozen modules commonly used in expansion-based parameter-isolation methods, $g_{\text{base}}$ remains a computationally dynamic Mamba module. Its selective scan still generates input-conditioned parameters even after freezing, providing stable but context-aware transformations that protect base-class knowledge while supporting continual adaptation.
The \textbf{incremental branch}\textbf{ ($g_{\text{inc}}$)} acts as a plasticity enabler. It is a single, fixed-capacity Mamba module reused across all incremental sessions. Rather than expanding parameters for each session, $g_{\text{inc}}$ leverages Mamba's input-dependent dynamic computation to capture residual feature shifts for novel classes on top of the frozen anchor, enabling efficient adaptation without session-wise parameter growth.

Second, to resolve the \textbf{regularization challenge}, we introduce the \textbf{class-sensitive selective scan mechanism}. This mechanism explicitly regulates the dynamic adaptation of the incremental branch ($g_{\text{inc}}$) to enforce class-sensitive behaviors through two complementary objectives. The \textbf{suppression loss} encourages stability by minimizing the activation of $g_{\text{inc}}$ on base-class inputs. This effectively silences the incremental branch for base-class inputs, ensuring their representations remain governed by the frozen stability anchor to prevent interference. The \textbf{separation loss} enhances plasticity by maximizing the dissimilarity between the generated parameters of base and novel classes. This drives $g_{\text{inc}}$ to form distinct, discriminative transformation pathways for new concepts, ensuring that the model captures novel-specific feature shifts while minimizing the overlap with base-class processing logic.

Mamba-FSCIL combines dual-branch structural separation and class-sensitive parameter regularization, transforming Mamba into a dynamic adaptation framework tailored to FSCIL. Our key contributions are as follows:
\begin{itemize}
\item  We introduce Mamba-FSCIL, the first framework that leverages Mamba’s input-dependent dynamic computation for FSCIL. By generating input-dependent parameters, it achieves class-sensitive adaptation, reducing interference with base-class representations while enhancing plasticity for novel classes within a fixed-capacity model.
\item We propose the dual selective SSM projector which separates base and novel class processing via a frozen base selective SSM branch and a dynamic selective SSM incremental branch. This structure helps protect base-class representations while enabling input-dependent dynamic adaptation for novel classes.
\item  We introduce a class-sensitive selective scan mechanism that regularizes the incremental branch’s dynamic adaptation. The suppression loss minimizes its impact on base representations to maintain stability, while the separation loss promotes distinct operator-generation patterns for base and novel classes to enhance plasticity.
\item Extensive experiments on miniImageNet, CIFAR-100, and CUB-200 demonstrate that Mamba-FSCIL achieves state-of-the-art performance. It preserves base-class accuracy, adapts effectively to novel classes, and does not add new projector branches across incremental sessions, offering a practical and scalable alternative to parameter-sharing and parameter-expansion paradigms.
\end{itemize}

\section{Related Work}
\label{sec:related}
\subsection{Few-Shot Learning (FSL)}
FSL aims to enable models to recognize new classes from only a few labeled examples~\cite{ravi2017optimization,chen2019closer,ye2020heterogeneous,wang2020generalizing,zou2024flatten,zou2024closer,zou2024attention} and has been extensively explored through data augmentation, metric learning, model design, and optimization strategies~\cite{jadon2020overview}.
\textit{Data augmentation} approaches enhance generalization by synthesizing diverse samples using transformations or generative models~\cite{kong2022few}. 
\textit{Metric-based methods} learn a shared embedding space for distance-based classification, exemplified by Siamese Networks~\cite{koch2015siamese}, Matching Networks~\cite{vinyals2016matching}, and Prototypical Networks~\cite{snell2017prototypical}. 
\textit{Model-based approaches} introduce specialized architectures or refine existing modules, such as Memory-Augmented Networks~\cite{santoro2016meta}, re-designed CLS tokens~\cite{zou2024closer}, attention calibration~\cite{zou2024attention}, and token continuity modeling~\cite{yi2025revisiting}.
\textit{Optimization-based strategies} facilitate rapid adaptation through meta-learned initializations like MAML~\cite{finn2017model} and Reptile~\cite{nichol2018first}, or by flattening the loss landscape to improve generalization~\cite{zou2024flatten}.

\subsection{Class-Incremental Learning (CIL)}
CIL enables models to learn new classes sequentially without forgetting previously acquired knowledge~\cite{cauwenberghs2000incremental,li2017learning,rebuffi2017icarl,de2019continual,masana2020class,tao2020topology,wang2022foster,zhou2024class}. Existing methods are broadly categorized into exemplar-based and non-exemplar-based approaches.
\textit{Exemplar-based approaches} mitigate forgetting by retaining a subset of past samples. iCaRL~\cite{rebuffi2017icarl} combines rehearsal with knowledge distillation, while BiC~\cite{wu2019large} and GEM~\cite{lopez2017gradient} calibrate predictions or constrain gradients using exemplars. Recent works have also explored embedding-level storage~\cite{iscen2020memory} and generative replay~\cite{xiang2019incremental}.
\textit{Non-exemplar-based approaches} prevent forgetting by constraining parameter updates. Regularization methods such as EWC~\cite{kirkpatrick2017overcoming}, MAS~\cite{aljundi2018memory}, and SI~\cite{zenke2017continual} penalize changes to important weights, while others modify the architecture via dynamic expansion~\cite{xu2018reinforced,yoon2018lifelong}.

\subsection{Few-Shot Class-Incremental Learning (FSCIL)}
\label{sec:related_fscil}
FSCIL~\cite{zhang2023few,tian2024survey} integrates the data-scarce setting of FSL with the continual learning objective of CIL, requiring models to incrementally integrate new classes with limited examples while retaining old knowledge. This setting introduces unique challenges: (1) \textit{Catastrophic Forgetting}, where adapting to new classes may erode previously acquired knowledge~\cite{mccloskey1989catastrophic,goodfellow2013empirical,yang2024corda}. (2) \textit{Overfitting}, as limited data hinders the learning of robust representations~\cite{snell2017prototypical,sung2018learning}. (3) \textit{Stability--Plasticity Dilemma}, which demands a delicate balance between maintaining old knowledge and adapting to new information~\cite{hebb2005organization,grossberg2012studies,mermillod2013stability}. 

Various methods have been proposed to address these challenges. 
\textit{Replay- or data-based methods} mitigate forgetting by replaying real or synthetic data~\cite{liu2022few,agarwal2022semantics,peng2022few}. 
Other methods improve FSCIL through meta-learning~\cite{yoon2020xtarnet,chi2022metafscil,zhou2022few}, pseudo-feature generation~\cite{cheraghian2021synthesized,zhou2022forward}, advanced loss functions or regularizers~\cite{tao2020topology,joseph2022energy,lu2022geometer,chen2021incremental,akyurek2021subspace,song2023learning,ahmed2024orco,zou2024compositional,goswami2024calibrating,zhou2024delve,oh2024closer}, well-designed classifiers~\cite{hersche2022constrained,yang2023neural,yang2023neural_arxiv,zhu2021self,wang2024few}, and knowledge distillation~\cite{tao2020few,dong2021few,cheraghian2021semantic,zhao2023few}. 
For example, NC-FSCIL~\cite{yang2023neural} introduces a fixed equiangular tight frame classifier and trains an MLP-based projection layer to align backbone features with ideal prototypes. Comp-FSCIL~\cite{zou2024compositional} designs a cognitive-inspired compositional reasoning mechanism to reuse visual primitives from base classes.
However, many of these methods rely on \textit{static adaptation}, where input-agnostic parameters may limit the model's flexibility in balancing old-class retention and novel-class adaptation. 

\textit{Dynamic adaptation with expanding architectures} methods expand model capacity by adding new parameters for each new session while preserving previous ones~\cite{tao2020few,yang2021learnable,yan2021dynamically,yang2022dynamic,zhang2021few,ahmad2022few}. For instance, TOPIC~\cite{tao2020few} integrates new nodes into a neural gas network, while CEC~\cite{zhang2021few} evolves the classifier’s topology using a dynamic graph. Other methods like LEC-Net~\cite{yang2021learnable}, DER~\cite{yan2021dynamically}, DSN~\cite{yang2022dynamic}, and FeSSSS~\cite{ahmad2022few} introduce task-specific parameters for new tasks. While enhancing plasticity, their continual expansion increases resource costs, especially as the number of incremental sessions grows.

Different from expansion-based methods that rely on parameter isolation through session-wise architectural growth, Mamba-FSCIL aims to improve the stability--plasticity balance without adding new projector branches across incremental sessions.
\textbf{First}, rather than allocating new parameters for every incoming session, we employ a single reused incremental branch. Its plasticity stems from Mamba's input-dependent dynamic computation, enabling adaptation to novel classes without session-wise parameter expansion.
\textbf{Second}, in contrast to static frozen modules such as MLPs, our frozen base branch remains computationally dynamic. It continues to generate input-conditioned parameters on the fly, providing context-aware transformations after its trainable weights are frozen.
\textbf{Third}, we further guide the incremental branch through class-sensitive regularization. This encourages the incremental branch to generate different operator patterns for base and novel classes, reducing interference with established knowledge during adaptation.

\subsection{State Space Models and Mamba}
\label{sec:related_ssm}
Sequence modeling architectures are useful for capturing token dependencies in spatial feature maps. Recurrent networks such as LSTM~\cite{hochreiter1997long} and GRU~\cite{cho2014learning} capture sequential dependencies through recurrence but suffer from limited long-range memory and inflexibility in adapting to new input distributions due to parameter sharing. Transformers~\cite{vaswani2017attention,dosovitskiy2020image,liu2021swin} overcome this with global self-attention, but their application to long sequences or high-resolution images is hindered by the quadratic complexity and memory cost of attention.

State Space Models (SSMs) provide an efficient alternative for long-sequence modeling by using recurrent state updates with favorable scaling properties~\cite{gu2021efficiently,gu2021combining, sun2023retentive,poli2023hyena,smith2023simplified,peng2023rwkv,fu2022hungry,wang2022pretraining,wang2023state,lu2024structured}. SSMs involve recurrent updates over a sequence through hidden states, maintaining context in hidden states, and updating outputs by integrating these states with incoming inputs. Structured state space models (S4)~\cite{gu2021efficiently,gu2021combining} exemplify this efficiency with a parameterized linear recurrent neural network. Subsequent works further improve their computational efficiency and expressiveness across domains~\cite{sun2023retentive,poli2023hyena,smith2023simplified,peng2023rwkv}.

Mamba~\cite{gu2023mamba} extends S4 by introducing an input-dependent selective scan mechanism and proposes hardware-aware algorithms to boost efficiency. This allows the model to dynamically control information flow based on input content, selectively propagating or suppressing information across the sequence to adapt its internal dynamics. Mamba has also been adopted to process non-sequential input such as images and videos in vision tasks~\cite{liu2024vmamba,zhu2024vision,ruan2024vm,liu2024swin,xing2024segmamba,ma2024u,li2024videomamba,han2024demystify}. Among them, VMamba~\cite{liu2024vmamba} introduces a 2D selective scan (SS2D) mechanism that processes images in both horizontal and vertical dimensions, allowing feature map elements to integrate information from all directions and providing a global receptive field while maintaining linear complexity with respect to the number of tokens.

Although Mamba has been applied in various domains such as language~\cite{gu2023mamba,fu2022hungry,gu2021efficiently,waleffe2024empirical}, vision~\cite{zhu2024vision,liu2024vmamba,li2024videomamba,lv2024decision}, and audio~\cite{goel2022sashimi}, its potential for mitigating forgetting and adapting to continually shifting distributions has not yet been explored.
To the best of our knowledge, we are the first to introduce Mamba into FSCIL, combining structural decoupling with class-sensitive regularization to exploit its dynamic adaptation capability for FSCIL.

\section{Preliminaries}
\label{sec:preliminaries}
\subsection{Few-Shot Class-Incremental Learning}
\label{sec:preliminaries_fscil}
FSCIL trains a model incrementally in multiple sessions $\{\mathcal{D}^{(0)}, \mathcal{D}^{(1)}, \dots, \mathcal{D}^{(s)}, \dots, \mathcal{D}^{(S)}\}, s \in \{0,1,\dots, S\}$, where $\mathcal{D}^{(s)}=\{(\mathbf{x}_i,y_i)\}_{i=1}^{|\mathcal{D}^{(s)}|}$ represents the training set for session $s$. $S$ is the number of incremental sessions. The process begins with $\mathcal{D}^{(0)}$, the base session, which typically features a large label space $\mathcal{C}^{(0)}$ and sufficient training data for each class $c \in \mathcal{C}^{(0)}$. In each incremental session $\mathcal{D}^{(s)}$, $s>0$, there are only a few labeled images, $|\mathcal{D}^{(s)}|=pq$, where $p$ is the number of classes and $q$ is the number of samples per novel class, known as $p$-way $q$-shot. There is no overlap in the label space between sessions \emph{i.e.,} $\mathcal{C}^{(s)}\cap\mathcal{C}^{(s')}=\emptyset$ for all $s'\neq s$. Raw images from previous sessions are typically not stored or replayed, which requires the model to generalize to new data without forgetting previously learned information. Evaluation in session $s$ involves test data from all classes encountered up to that session, \emph{i.e.,} the label space of $\cup_{i=0}^{s}\mathcal{C}^{(i)}$. In this paper, we use \textit{base classes} to denote classes in $\mathcal{C}^{(0)}$, and \textit{novel classes} to denote all classes learned after the base session, i.e., classes in $\cup_{j=1}^{s}\mathcal{C}^{(j)}$ at session $s$.

\subsection{State Space Models (SSMs)}
\label{sec:preliminaries_ssms}
State Space Models (SSMs) process sequential data by mapping a 1-D input signal $x(t) \in \mathbb{R}$ to an output $y(t) \in \mathbb{R}$ through a hidden state $h(t) \in \mathbb{R}^{d_{\text{state}}}$, where $d_{\text{state}}$ is the state dimension. A continuous Linear Time-Invariant (LTI) SSM is computed as:
\begin{equation}
\label{equ:ssm_continuous}
 h'(t) = \mathbf{A} h(t) + \mathbf{B} x(t), \quad y(t) = \mathbf{C} h(t),
\end{equation}
where $\mathbf{A} \in \mathbb{R}^{{d_{\text{state}}} \times {d_{\text{state}}}}$ governs the evolution of the hidden state over time, $\mathbf{B} \in \mathbb{R}^{{d_{\text{state}}} \times 1}$ projects the input into the state space, and $\mathbf{C} \in \mathbb{R}^{1 \times {d_{\text{state}}}}$ maps the hidden state to the output.

To process discrete sequences (e.g., text tokens), this system is discretized using the Zero-Order Hold (ZOH) rule, which assumes the input holds constant within a discrete interval of timescale $\mathbf{\Delta} \in \mathbb{R}^+$. The continuous parameters $(\mathbf{A}, \mathbf{B})$ are transformed into discrete parameters $(\overline{\mathbf{A}}, \overline{\mathbf{B}})$:
\begin{equation}
\label{equ:ssm_param_discretization}
\overline{\mathbf{A}} = \exp(\mathbf{\Delta} \mathbf{A}), \quad
\overline{\mathbf{B}} = (\mathbf{\Delta} \mathbf{A})^{-1}(\exp(\mathbf{\Delta} \mathbf{A}) - \mathbf{I}) \cdot \mathbf{\Delta} \mathbf{B}.
\end{equation}

Given a discrete input sequence $\mathbf{x} = (x_1, \dots, x_L)$ of length $L$, the discretized SSM computes the hidden state $h_t$ and output $y_t$ as:
\begin{equation}
\label{equ:ssm_discretized}
h_t = \overline{\mathbf{A}} h_{t-1} + \overline{\mathbf{B}} x_t, \quad
y_t  = \mathbf{C} h_t.
\end{equation}

For standard LTI SSMs (e.g., S4~\cite{gu2021efficiently}), the system matrices $(\mathbf{A},\mathbf{B},\mathbf{C})$ and the step size $\mathbf{\Delta}$ remain fixed across all timesteps. Under this time-invariant setting, Eq.~\ref{equ:ssm_discretized} can be rewritten as a global convolution:
\begin{equation}
\label{equ:mamba_kernel}
\mathbf{y} = \mathbf{x} * \overline{\mathbf{K}}, \quad  \overline{\mathbf{K}} = (\mathbf{C} \overline{\mathbf{B}}, \mathbf{C} \overline{\mathbf{A}}\overline{\mathbf{B}}, \dots, \mathbf{C} \overline{\mathbf{A}}^{L-1}\overline{\mathbf{B}})
\end{equation}
where $*$ denotes convolution and $\overline{\mathbf{K}}\in\mathbb{R}^{L}$ is the full sequence-length kernel.
This convolutional form enables parallel training while still allowing the model to switch to the recurrent formulation in Eq.~\ref{equ:ssm_discretized} for efficient autoregressive inference.
The state matrix $\mathbf{A}$ is typically initialized using the HiPPO framework~\cite{gu2020hippo} to enhance the model with long-range memory.

\subsection{Selective State Space Models (Mamba)} 
\label{sec:preliminaries_mamba}
While the LTI formulation of SSMs is efficient, its fixed system matrices are input-independent, limiting its ability to perform content-aware reasoning (e.g., selectively focusing on or filtering out specific inputs). Mamba~\cite{gu2023mamba} introduces the Selective State Space Model (S6), which parameterizes the system matrices as functions of the input.

Given an input sequence $\mathbf{X} \in \mathbb{R}^{L \times D}$, where each token $x_t \in \mathbb{R}^D$, the SSM is applied independently to each of the $D$ channels. The matrices $\mathbf{B}_t$, $\mathbf{C}_t$, and step size $\mathbf{\Delta}_t$ are conditioned on the input $x_t$ at each timestep $t$:
\begin{equation}
\label{equ:mamba_parameter_generation}
\mathbf{B}_t = f_B(x_t),\quad
\mathbf{C}_t = f_C(x_t),\quad
\mathbf{\Delta}_t = f_{\Delta}(x_t),
\end{equation}
where $f_B(\cdot)$, $f_C(\cdot)$, and $f_\Delta(\cdot)$ are learnable projections, e.g., linear layers, with output dimensions $\mathbf{B}_t, \mathbf{C}_t \in \mathbb{R}^{d_{\text{state}}}$ and $\mathbf{\Delta}_t \in \mathbb{R}^D$.

Consequently, the discretized parameters become input-dependent ($\overline{\mathbf{A}}_t, \overline{\mathbf{B}}_t$), and the selective state space model is written as:
\begin{equation}
\label{equ:S6}
h_t = \overline{\mathbf{A}}_t h_{t-1} + \overline{\mathbf{B}}_t x_t,  \quad
y_t = \mathbf{C}_t h_t,
\end{equation}
where $\overline{\mathbf{A}}_t$ and $\overline{\mathbf{B}}_t$ are computed using Eq.~\ref{equ:ssm_param_discretization}.
Mamba employs a hardware-aware parallel associative scan algorithm to compute the sequence efficiently. This is formalized as the sequence-level selective scan operator $\texttt{S6}(\cdot)$:
\begin{equation}
\label{equ:S6_operator}
\mathbf{Y} = \texttt{S6}(\mathbf{X}; \mathbf{B}, \mathbf{C}, \mathbf{\Delta}),
\end{equation}
where $\mathbf{Y} \in \mathbb{R}^{L \times D}$. This operator processes the entire sequence with dynamic parameters, enabling parallel hidden state propagation ($h_0, \dots, h_L$) and achieving linear complexity $O(L)$ with dynamic adaptability.

\section{Mamba-FSCIL}
\label{sec:method}
\begin{figure*}[t]
    \centering
    \includegraphics[width=1\linewidth]{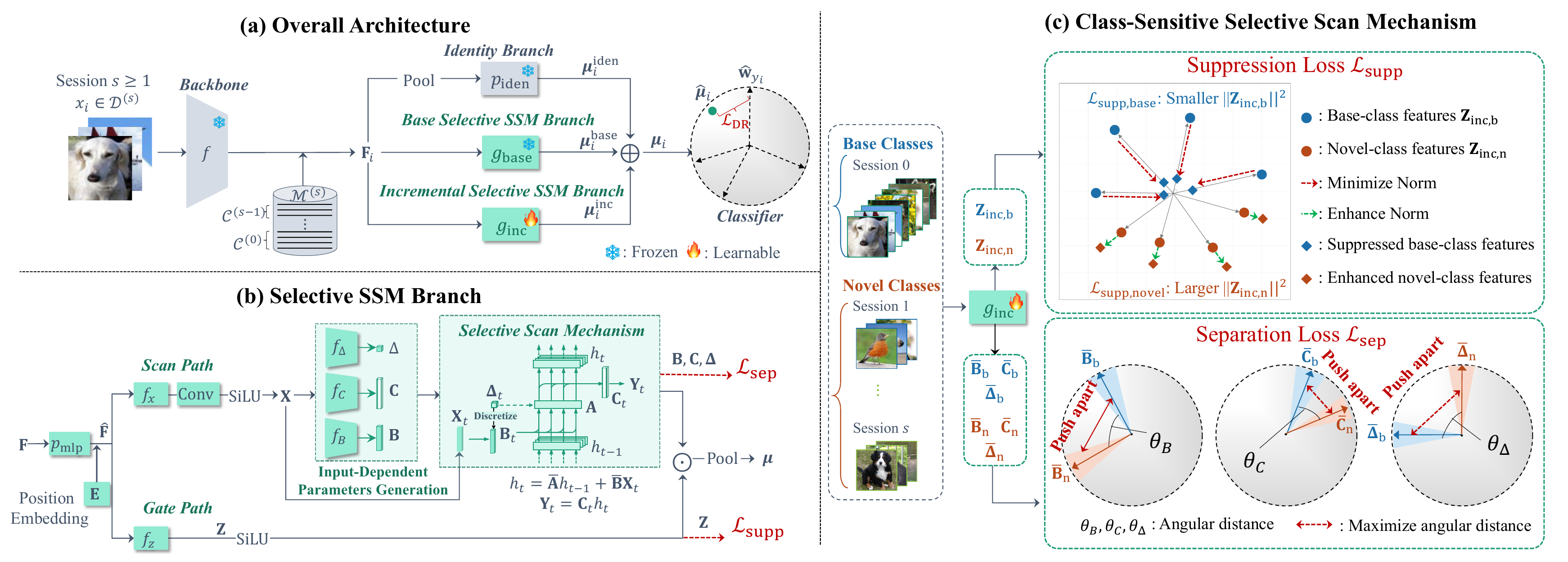}
    \caption{\textbf{The Framework of Mamba-FSCIL.}
    \textbf{(a) Overall Architecture:} The model comprises a frozen backbone, a \textbf{Dual Selective SSM Projector}, and a fixed classifier. The projector consists of: (1) an Identity Branch $p_{\text{iden}}$ for general feature preservation; (2) a Base Selective SSM Branch $g_{\text{base}}$ serving as a \textit{Stability Anchor}; and (3) an Incremental Selective SSM Branch $g_{\text{inc}}$ acting as a \textit{Plasticity Enabler}. Outputs are aggregated for the fixed ETF classifier.
    \textbf{(b) Selective SSM Branch Architecture:} Details the internal structure shared by $g_{\text{base}}$ and $g_{\text{inc}}$. It features a scan path that dynamically generates input-dependent parameters ($\mathbf{B}, \mathbf{C}, \mathbf{\Delta}$) to drive the selective state-space model, and a gate path producing features $\mathbf{Z}$ to modulate the branch output.
    \textbf{(c) Class-Sensitive Selective Scan Mechanism:} Regulates the incremental branch ($g_{\text{inc}}$) using two loss terms. Suppression loss ($\mathcal{L}_{\text{supp}}$) acts on the gating features $\mathbf{Z}_{\text{inc}}$: it penalizes the norm for base-class inputs ($\mathbf{Z}_{\text{inc, b}}$, black) to prevent interference (stability) while encouraging a larger norm for novel-class inputs ($\mathbf{Z}_{\text{inc, n}}$, orange) to facilitate adaptation (plasticity). Separation loss ($\mathcal{L}_{\text{sep}}$) acts on the generated parameters. It reduces the absolute cosine similarity between the averaged operators of base classes $(\bar{\mathbf{B}}_\text{b},\bar{\mathbf{C}}_\text{b},\bar{\mathbf{\Delta}}_\text{b})$ and novel classes $(\bar{\mathbf{B}}_\text{n},\bar{\mathbf{C}}_\text{n},\bar{\mathbf{\Delta}}_\text{n})$, encouraging the model to form distinct processing patterns without session-wise projector expansion.
}
    \label{fig:framework}
    \vspace{-2mm}
\end{figure*}

\subsection{Overall Architecture}
\label{sec:method_overall_arch}
Mamba-FSCIL aims to continuously learn new classes while reducing forgetting within a projector architecture that does not add new session-specific branches. To achieve this, our framework is constructed to resolve the two fundamental challenges: the structural challenge (preserving base-class knowledge while reducing interference) and the regularization challenge (enforcing explicit class-sensitivity in dynamic computation). 
As shown in Fig.~\ref{fig:framework}~(a), the overall architecture comprises the following components:

\textbf{Backbone ($f$)} serves as the feature extractor and is trained once using the abundant base-class data. It learns mid-level semantic features that form the input to all projection branches. Following standard FSCIL practice~\cite{zhang2021few,yang2023neural,zou2022margin,zou2024compositional}, $f$ is frozen after the base session so that the backbone mapping remains fixed throughout all incremental sessions.

\textbf{Dual selective SSM projector} (Sec.~\ref{sec:method_dual_projector})
addresses the structural challenge by decoupling the learning process into two branches. The base selective SSM branch ($g_{\text{base}}$) acts as a stability anchor, which is optimized for base classes and then frozen to preserve foundational knowledge. 
The incremental selective SSM branch ($g_{\text{inc}}$) acts as a plasticity enabler, which is a single, fixed-capacity module reused across sessions that leverages Mamba's input-dependent parameter generation to capture discriminative feature shifts for novel classes.

\textbf{Class-Sensitive Selective Scan Mechanism} 
(Sec.~\ref{sec:method_selective_scan}) addresses the regularization challenge by explicitly guiding the dynamic behavior of $g_{\text{inc}}$. 
It employs a suppression loss to minimize the activation of $g_{\text{inc}}$ on base-class inputs (preserving stability) and a separation loss to reduce the similarity between generated operators for base and novel classes (enhancing plasticity).

\textbf{Classifier} adopts a fixed simplex equiangular tight frame (ETF) following \cite{yang2023neural,zhong2023understanding,li2023no,seo2024learning,xiao2024targeted,ma2023digeo}. This structure ensures maximal equiangular separation among classes \cite{yang2022inducing,papyan2020prevalence}, enabling the projector to generate distinct features for both base and novel classes. The optimization process across sessions is detailed in Sec.~\ref{sec:method_optim}.

\textbf{Feature Extraction and Batch Construction.}
Given a raw image $\mathbf{x}$, the backbone $f$ extracts an intermediate spatial feature map $\mathbf{f}=f(\mathbf{x})\in\mathbb{R}^{D\times H\times W}$, where $D$ is the channel dimension and $H\times W$ is the spatial resolution. 
During the base session ($s=0$), the model is trained with raw-image mini-batches $\mathcal{B}^{(0)}$ sampled from $\mathcal{D}^{(0)}$, and the corresponding feature-level batch is:
\begin{equation}
\label{equ:features_backbone_base}
    \mathbf{F}^{(0)}
    =
    \{ f(\mathbf{x}) \mid (\mathbf{x},y)\in\mathcal{B}^{(0)} \}.
\end{equation}

During incremental sessions, Mamba-FSCIL follows an \textit{image-exemplar-free} protocol: it does not store or replay raw images from previous sessions, but maintains a compact feature-level memory, a practice widely validated in prior FSCIL studies \cite{hersche2022constrained,chen2021incremental,akyurek2021subspace}. 
Specifically, the memory stores one class-mean backbone feature map $\bar{\mathbf{F}}_c$ for each class learned before session $s$:
\begin{equation}
\label{equ:memory}
\mathcal{M}^{(s)}
=
\{\bar{\mathbf{F}}_c \mid c\in\cup_{j=0}^{s-1}\mathcal{C}^{(j)}\},
\quad 
\bar{\mathbf{F}}_c 
= 
\frac{1}{N_c}\sum_{i:y_i=c} f(\mathbf{x}_i).
\end{equation}
where $\mathcal{C}^{(j)}$ is the label set of session $j$, $N_c$ is the number of training samples from class $c$, and $\bar{\mathbf{F}}_c\in\mathbb{R}^{D\times H\times W}$. Each stored prototype is associated with its class label $c$. After each session, the class-mean features of the newly learned classes are computed from the available training images in that session and added to the memory for subsequent sessions.

For an incremental session $s\ge1$, the backbone $f$ is frozen. Given a current raw-image mini-batch $\mathcal{B}^{(s)}$ sampled from $\mathcal{D}^{(s)}$, we first extract the current-session features $  \mathbf{F}_{\text{cur}}^{(s)}=\{f(\mathbf{x}) \mid (\mathbf{x},y)\in\mathcal{B}^{(s)}\}$. The projector input is then constructed at the feature level by combining current-session features with stored prototypes from previously learned classes: $\mathbf{F}^{(s)} = \mathbf{F}_{\text{cur}}^{(s)} \cup \mathcal{M}^{(s)}$. Thus, raw images from previous sessions are never replayed or passed through the backbone during incremental training.

For the Class-Sensitive Scan losses, we partition the feature batch $\mathbf{F}^{(s)}$ into base-class and novel-class features:
\begin{equation}
\label{equ:batch_split}
\begin{aligned}
    &\mathbf{F}^{(s)}_{\mathrm{base}}
    =
    \{\bar{\mathbf{F}}_c\in\mathcal{M}^{(s)}
    \mid c\in\mathcal{C}^{(0)}\}, \\
    &\mathbf{F}^{(s)}_{\mathrm{novel}}
    =
    \mathbf{F}_{\text{cur}}^{(s)}
    \cup
    \{\bar{\mathbf{F}}_c\in\mathcal{M}^{(s)}
    \mid c\in\cup_{j=1}^{s-1}\mathcal{C}^{(j)}\},
\end{aligned}
\end{equation}
where $\mathbf{F}^{(s)}_{\mathrm{novel}}$ denotes the novel-class feature group at session $s$. It includes current-session novel-class features and stored feature prototypes from previously learned novel classes. The resulting features are fed into the dual-branch projector.

\subsection{Dual Selective SSM Projector}
\label{sec:method_dual_projector}
The dual selective SSM projector structurally decouples stability and plasticity via two parallel branches. Both branches utilize the same Selective SSM architecture (detailed in Sec.~\ref{sec:method_ssm_branch}) but differ in their optimization protocols:
\begin{itemize}
    \item \textbf{Base selective SSM branch} is optimized solely during the base session and subsequently frozen to help preserve foundational representations during incremental learning.
    \item \textbf{Incremental selective SSM branch} is a single, fixed-capacity module reused across all incremental sessions. By leveraging input-dependent dynamic computation, it adapts to novel classes without parameter expansion.
\end{itemize}

This design supports long-term stability through the frozen $g_{\text{base}}$ while enabling dynamic adaptation via $g_{\text{inc}}$. The following section details the shared Selective SSM Branch architecture (Sec.~\ref{sec:method_ssm_branch}) used in these branches, and its instantiation during Base (Sec.~\ref{sec:method_base_session}) and Incremental Sessions (Sec.~\ref{sec:method_incremental_session}).

\subsubsection{Selective SSM Branch}
\label{sec:method_ssm_branch}
Both $g_{\text{base}}$ and $g_{\text{inc}}$ utilize the same Selective SSM architecture, based on the 2D Selective Scan mechanism~\cite{liu2024vmamba}, which extends Mamba’s input-conditioned selective scan to 2D visual features, processing along four spatial directions to enable global context aggregation. As shown in Fig.~\ref{fig:framework}~(b), the process proceeds as follows:

\textbf{\textit{Tokenization and Positional Encoding.}}
We reshape the backbone feature map $\mathbf{F}$ into a sequence of $L = H \times W$ spatial tokens $\texttt{Reshape}(\mathbf{F}) \in \mathbb{R}^{N \times L \times D}$. Learnable positional embeddings $\mathbf{E} \in \mathbb{R}^{1 \times L \times D'}$ are added after projecting the tokens to dimension $D'$ via an MLP:
\begin{equation}
\label{equ:ssm_branch_backbone_addpos}
\hat{\mathbf{F}} = p_\text{mlp}(\texttt{Reshape}(\mathbf{F})) + \mathbf{E}, {\quad \hat{\mathbf{F}} \in \mathbb{R}^{N \times L \times D'}}.
\end{equation}

\textbf{\textit{Scan and Gating Stream Construction.}}
The token sequence is split into two parallel streams: the \textit{scan stream} $\mathbf{X}$, driving the selective scanning, and the \textit{gate stream} $\mathbf{Z}$, modulating the output. These are computed as:
\begin{equation}
\label{equ:ssm_branch_scan_and_gate}
\mathbf{X} = \texttt{SiLU}(\texttt{Conv}(p_x(\hat{\mathbf{F}}))), \quad
\mathbf{Z} = p_z(\hat{\mathbf{F}}),
\end{equation}
where $\mathbf{X}, \mathbf{Z} \in \mathbb{R}^{N \times L \times D'}$, and $p_x, p_z$ are linear projections. The scan stream $\mathbf{X}$ undergoes a linear projection $p_x(\cdot)$ followed by group convolution $\texttt{Conv}$ and element-wise SiLU activation $\texttt{SiLU}$ to capture local contexts.

\textbf{\textit{Directional Scanning.}}
To capture global context, the cross-scan module traverses the 2D feature map $\mathbf{X}$ along $K = 4$ directions (top-left $\rightarrow$ bottom-right, bottom-right $\rightarrow$ top-left, top-right $\rightarrow$ bottom-left, and bottom-left $\rightarrow$ top-right). For each direction $k \in \{1,\dots,K\}$, the $\texttt{ScanExpand}$ operator reorders the tokens into a direction-specific sequence $\mathbf{X}_{k}$:
\begin{equation}
\label{equ:ssm_branch_directional_scanning}
\mathbf{X}_k = \texttt{ScanExpand}(\mathbf{X}, k) \in \mathbb{R}^{N \times L \times D'}.
\end{equation}

\textbf{\textit{Input-Conditioned Parameter Generation.}}
Following Mamba’s parameter-generation rule (Eq.~\ref{equ:mamba_parameter_generation}), state-space parameters are generated for each ordered token sequence $\mathbf{X}_k$:
\begin{equation}
\begin{aligned}
\label{equ:ssm_branch_parameter_generation}
\mathbf{B}_k = f_B(\mathbf{X}_k), \quad
\mathbf{C}_k = f_C(\mathbf{X}_k), \quad
\mathbf{\Delta}_k = f_{\Delta}(\mathbf{X}_k),
\end{aligned}
\end{equation}
where $f_B$ and $f_C$ are linear transformations mapping the input dimension $D'$ to $D_B$ and $D_C$, respectively. We set $D_B = D_C = d_{\text{state}}$, where $d_{\text{state}}$ is the SSM state dimension. For $f_\Delta$, the input is first projected from $D'$ to a low-rank intermediate dimension $D_\Delta$ (set to $D_\Delta=d_{\text{state}}$), and then projected back to $D'$, followed by a Softplus activation to ensure positivity. This produces the direction-wise parameter sets:
\begin{equation}
\label{equ:ssm_parameters}
\mathbf{B} \in \mathbb{R}^{N \times K \times L \times D_B}, \mathbf{C} \in \mathbb{R}^{N \times K \times L \times D_C}, \mathbf{\Delta} \in \mathbb{R}^{N \times K \times L \times D'}.
\end{equation}

\textbf{\textit{Selective SSM updates.}}
For each direction $k$, the sequence $\mathbf{X}_k$ is independently processed by the selective scan mechanism (Eq.~\ref{equ:S6_operator}) using the generated parameters:
\begin{equation}
\label{equ:ssm_branch_s6_recurrence}
\mathbf{Y}_{k} = \texttt{S6}(\mathbf{X}_{k}; \mathbf{B}_{k}, \mathbf{C}_{k}, \mathbf{\Delta}_{k}) \in \mathbb{R}^{N \times L \times D'}.
\end{equation}

\textbf{\textit{Output Gating.}}
The output sequences are restored to the original 2D arrangement via $\texttt{InverseScan}$ and summed. Finally, the result is gated by $\mathbf{Z}$ and pooled:
\begin{equation}
\label{equ:ssm_branch_gated_output}
\begin{aligned}
\boldsymbol{\mu} 
&= g_{\text{ssm}}(\mathbf{F}) \\[2pt]
&= \texttt{Pool}\!\left( 
    \left( \sum_{k=1}^{K} \texttt{InverseScan}(\mathbf{Y}_{k}, k) \right)
    \odot 
    \texttt{SiLU}(\mathbf{Z})
\right).
\end{aligned}
\end{equation}
where $\odot$ denotes element-wise multiplication. The function $g_{\text{ssm}}(\cdot)$ represents the generic selective SSM branch. In subsequent sections, we instantiate it as the Base Branch $g_{\text{base}}$ and the Incremental Branch $g_{\text{inc}}$. These branches share the same functional form but utilize distinct learnable parameter sets, denoted as $\Theta_{\text{base}}$ and $\Theta_{\text{inc}}$:
\begin{equation}
\label{equ:g_base_g_inc_definition}
g_{\text{base}}(\mathbf{F}) \triangleq g_{\text{ssm}}(\mathbf{F}; \Theta_{\text{base}}), 
\quad
g_{\text{inc}}(\mathbf{F})   \triangleq g_{\text{ssm}}(\mathbf{F}; \Theta_{\text{inc}}).
\end{equation}

\subsubsection{Base Session ($s=0$)}
\label{sec:method_base_session}
During the base session, Mamba-FSCIL learns a strong, discriminative representation space from the abundant base-class dataset $\mathcal{D}^{(0)}$. This stage aims to: (1) establish a stable and expressive foundation for incremental sessions, and (2) form the structural separation for incremental learning by fully training the base branch ($g_{\text{base}}$) before freezing it. Thus, the backbone $f$, the identity branch $p_{\text{iden}}$, and the base selective SSM branch $g_{\text{base}}$ are jointly optimized:

\textbf{Backbone Network ($f$)} learns robust and general-purpose visual representations, producing the base-session feature batch $\mathbf{F}^{(0)}$ as defined in Eq.~\ref{equ:features_backbone_base}, which serves as the input to both projection branches.

\textbf{Identity Branch ($p_{\text{iden}}$)} provides a direct, minimally transformed path from the backbone to the classifier, preserving backbone semantics and offering consistent residual information across sessions. It transforms $\mathbf{F}^{(0)}$ to $\boldsymbol{\mu}_{\text{iden}}^{(0)}$ via average pooling followed by a linear projection $p_{\text{iden}}$:
\begin{equation}
\label{equ:mu_iden_0}
    \boldsymbol{\mu}_{\text{iden}}^{(0)} = p_{\text{iden}}(\texttt{Pool}(\mathbf{F}^{(0)})) \in \mathbb{R}^{N \times D'}.
\end{equation}

\textbf{Base Selective SSM Branch ($g_\text{base}$)} is optimized to extract highly discriminative, input-dependent representations for base classes. Using the selective scan architecture defined in Sec.~\ref{sec:method_ssm_branch}, it performs input-conditioned state-space updates to produce globally contextualized representations $\boldsymbol{\mu}_{\text{base}}^{(0)}$:
\begin{equation}
\label{equ:mu_base_0}
\begin{aligned}
  \boldsymbol{\mu}_{\text{base}}^{(0)} &= g_\text{base}({\mathbf{F}}^{(0)}) \in \mathbb{R}^{N \times D'}.
\end{aligned}
\end{equation}
After training in the base session, $g_\text{base}$ is frozen for preserving the base-class representation space and reducing parameter overwriting during incremental learning. Although its parameters are fixed, its selective SSM mechanism remains computationally dynamic, continuing to generate input-conditioned parameters that supply stable foundational representations for both base and novel inputs in later sessions.
    
The final base-session representation is the sum of outputs from the identity and base branches:
\begin{equation}
\label{equ:mu_0}
    \boldsymbol{\mu}^{(0)} = \boldsymbol{\mu}_{\text{iden}}^{(0)} + \boldsymbol{\mu}_{\text{base}}^{(0)} \in \mathbb{R}^{N \times D'}.
\end{equation}
where $\boldsymbol{\mu}^{(0)}$ becomes the stability anchor that all incremental sessions must retain, and the foundation upon which the incremental branch later learns novel-class residuals.

\subsubsection{Incremental Sessions ($s \ge 1$)}
\label{sec:method_incremental_session}
Incremental sessions integrate novel classes under few-shot constraints while preserving the representation space learned in the base session. This stage addresses the stability–plasticity dilemma through: (1) structural decoupling, which protects base knowledge via frozen components, and (2) input-dependent plasticity, enabling flexible adaptation to novel classes through the Incremental SSM branch $g_{\text{inc}}$.

To preserve the base representation space, the backbone $f$, the identity branch $p_{\text{iden}}$, and the base selective SSM branch $g_{\text{base}}$ are all frozen after the base session (black in Fig.~\ref{fig:framework}~(a)).
This structural constraint helps keep the feature structure learned during the base stage stable, sustaining reliable base-class representations. It provides a stable feature foundation for novel-class adaptation, on top of which $g_{\text{inc}}$ learns the necessary novel-class feature shifts. As a result, these frozen components consistently produce features for any input $\mathbf{F}^{(s)}$:
 \begin{equation}
 \begin{aligned}
 \label{equ:mu_iden_base_t}
    \boldsymbol{\mu}_{\text{iden}}^{(s)} &= p_{\text{iden}}(\texttt{Pool}(\mathbf{F}^{(s)})) \in \mathbb{R}^{N \times D'}, \\
    \boldsymbol{\mu}_{\text{base}}^{(s)} &= g_\text{base}({\mathbf{F}}^{(s)}) \in \mathbb{R}^{N \times D'}.
\end{aligned}
\end{equation}

\textbf{Incremental Selective SSM Branch ($g_{\text{inc}}$).}
This branch (orange modules in Fig.~\ref{fig:framework}~(a)) is the only component updated during incremental sessions. It operates on top of the frozen stability components to capture \emph{residual feature shifts}—nuanced attributes required for novel classes that are not fully encoded by the general-purpose frozen features ($\boldsymbol{\mu}_{\text{iden}}^{(s)} + \boldsymbol{\mu}_{\text{base}}^{(s)}$).

Unlike static projectors that apply a fixed transformation to all inputs (risking catastrophic forgetting) or expansion-based methods that add new parameters for every session (causing session-wise growth), $g_{\text{inc}}$ balances stability and plasticity without adding session-specific projector branches. This capability stems from Mamba’s input-dependent dynamic computation. As described in Sec.~\ref{sec:method_ssm_branch}, the state-space parameters are generated conditioned on the input sequence. Since the feature statistics of base-class tokens ($\mathbf{X}_\text{b}$) differ from those of novel-class tokens ($\mathbf{X}_\text{n}$), $g_{\text{inc}}$, which leverages Mamba's input-dependent mechanism, provides the architectural capacity to generate distinct dynamic parameters:
\begin{equation}
\label{equ:ssm_branch_parameter_generation_basenovel}
\begin{aligned}
\{\mathbf{B}_\text{b}, \mathbf{C}_\text{b}, \mathbf{\Delta}_\text{b}\} &= \{f_B(\mathbf{X}_\text{b}), f_C(\mathbf{X}_\text{b}), f_{\Delta}(\mathbf{X}_\text{b})\}, \\
\{\mathbf{B}_\text{n}, \mathbf{C}_\text{n}, \mathbf{\Delta}_\text{n}\} &= \{f_B(\mathbf{X}_\text{n}), f_C(\mathbf{X}_\text{n}), f_{\Delta}(\mathbf{X}_\text{n})\}.
\end{aligned}
\end{equation}
This input-conditioned divergence empowers a single set of weights to execute distinct transformations based on the input context, giving $g_{\text{inc}}$ the potential for class-sensitive adaptation:
\begin{itemize}
    \item \textbf{Plasticity}: Novel-class tokens can induce adaptive parameter patterns that capture feature shifts missed by the frozen $g_{\text{base}}$, which is critical to form precise decision boundaries from only a few samples.
    \item \textbf{Stability}: Base-class inputs can be guided to produce limited incremental-branch responses, reducing interference with the frozen $g_{\text{base}}$.
    \item \textbf{Scalability}: As adaptation arises from dynamic parameter generation, the projector is reused across sessions without session-wise parameter expansion.
\end{itemize}

To transform this intrinsic potential into a reliable and controllable mechanism, we further introduce the Class-Sensitive Selective Scan Mechanism (Sec.~\ref{sec:method_selective_scan}) to explicitly regularize the parameter generation process.

The output of $g_{\text{inc}}$ for $\mathbf{F}^{(s)}$ is computed as:
\begin{equation}
\begin{aligned}
\label{equ:mu_inc_t}
  \boldsymbol{\mu}_{\text{inc}}^{(s)} &= g_\text{inc}({\mathbf{F}}^{(s)}) \in \mathbb{R}^{N \times D'}.
\end{aligned}
\end{equation}
     
The full representation at session $s$ combines the stable frozen features with the adaptive residual features from $g_{\text{inc}}$:
\begin{equation}
\label{equ:mu_t}
\boldsymbol{\mu}^{(s)} = \boldsymbol{\mu}_{\text{iden}}^{(s)} + \boldsymbol{\mu}_{\text{base}}^{(s)} + \boldsymbol{\mu}_{\text{inc}}^{(s)} \in \mathbb{R}^{N \times D'}.
\end{equation}

\subsection{Class-Sensitive Selective Scan Mechanism}
\label{sec:method_selective_scan}
The dual selective SSM projector separates stability ($g_\text{base}$) and plasticity ($g_\text{inc}$) through distinct branches. As described in Sec.~\ref{sec:method_incremental_session}, $g_\text{inc}$ generates input-dependent state-space parameters through the projection functions $f_B$, $f_C$, and $f_\Delta$, which are shared across base-class and novel-class inputs but produce different parameters according to the input features. To further make this dynamic parameter generation class-sensitive, we introduce the \textbf{Class-Sensitive Selective Scan Mechanism}. It regularizes the input-conditioned parameter generation in $g_\text{inc}$ to produce more separated operator patterns for base-class and novel-class inputs, encouraging limited responses on base-class inputs and stronger adaptive responses on novel-class inputs. This mechanism is implemented through two complementary objectives.

\subsubsection{Suppression Loss ($\mathcal{L}_{\text{supp}}$)}
\label{sec:method_selective_scan_loss_suppression}
For base-class inputs, the frozen components ($p_{\text{iden}}, g_{\text{base}}$) already provide a stable and discriminative feature anchor (Eq.~\ref{equ:mu_iden_base_t}). To preserve base-class performance, the incremental branch should avoid perturbing this frozen anchor. We introduce the \textbf{Suppression Loss} to encourage $g_{\text{inc}}$ to contribute minimally on base-class inputs while remaining responsive to novel-class inputs.

We achieve this by regulating the gating stream $\mathbf{Z}_{\text{inc}}$ (Eq.~\ref{equ:ssm_branch_scan_and_gate}) from $g_\text{inc}$, as shown in the upper part of Fig.~\ref{fig:framework}~(c). Since $\mathbf{Z}_{\text{inc}}$ modulates the output of $g_{\text{inc}}$ through element-wise gating (Eq.~\ref{equ:ssm_branch_gated_output}), controlling its magnitude directly regulates the contribution of $\boldsymbol{\mu}_{\text{inc}}$:
\begin{itemize}
    \item \textbf{For base classes:} We minimize $\|\mathbf{Z}_{\text{inc,b}}\|^2$ to suppress the incremental branch, making the residual term $\boldsymbol{\mu}_{\text{inc}}$ in Eq.~\ref{equ:mu_t} small, so that the final representation remains close to the base-session form in Eq.~\ref{equ:mu_0}.
    \item \textbf{For novel classes:} We encourage a larger squared norm of the gating features from $g_{\text{inc}}$, denoted as $\|\mathbf{Z}_{\text{inc,n}}\|^2$. This allows $g_{\text{inc}}$ to contribute more to the representation, capturing novel-specific feature shifts required for few-shot discrimination.
\end{itemize}

Given the partitioned feature batches $\mathbf{F}^{(s)}_{\text{base}}$ and $\mathbf{F}^{(s)}_{\text{novel}}$ defined in Eq.~\ref{equ:batch_split}, let $N_{\text{base}} = |\mathbf{F}^{(s)}_{\text{base}}|$ and $N_{\text{novel}} = |\mathbf{F}^{(s)}_{\text{novel}}|$. The suppression loss is formulated as:
\begin{equation}
\begin{aligned}
\label{equ:loss_supp}
\mathcal{L}^{(s)}_{\text{supp, base}} &= \frac{1}{N_{\text{base}} \cdot L \cdot D'} \sum_{i=1}^{N_{\text{base}}} \|\mathbf{Z}_{\text{inc}, \text{b}, i}\|^2, \\
\mathcal{L}^{(s)}_{\text{supp, novel}} &= -\frac{1}{N_{\text{novel}} \cdot L \cdot D'} \sum_{j=1}^{N_{\text{novel}}} \|\mathbf{Z}_{\text{inc}, \text{n}, j}\|^2, \\
\mathcal{L}_{\text{supp}}^{(s)} &= \lambda_1 \mathcal{L}^{(s)}_{\text{supp, base}} + \lambda_2 \mathcal{L}^{(s)}_{\text{supp, novel}},
\end{aligned}
\end{equation}
where $\mathbf{Z}_{\text{inc},\text{b},i}$ and $\mathbf{Z}_{\text{inc},\text{n},j}$ are the gating vectors derived from the $i$-th base feature and the $j$-th novel-class feature, respectively. $\lambda_1$ and $\lambda_2$ balance the suppression strength for base-class inputs and the activation strength for novel-class inputs. This loss guides $g_{\text{inc}}$ to maintain base-class stability while promoting adaptability to novel classes.

\subsubsection{Separation Loss ($\mathcal{L}_{\text{sep}}$)}
\label{sec:method_selective_scan_loss_separation}
As discussed in Sec.~\ref{sec:method_incremental_session}, $g_{\text{inc}}$ can produce input-dependent parameter sets for different input groups. To make this class-sensitive behavior more explicit and better exploit its plasticity for novel-class learning, we introduce the \textbf{Separation Loss}. It encourages $g_{\text{inc}}$ to form distinct transformation patterns for base and novel inputs by reducing the cosine similarity between their generated state-space parameters. This provides additional flexibility for novel-class adaptation while reducing interference with base-class representations.

We first compute the averaged input-dependent parameters for base-class features $\mathbf{F}^{(s)}_{\text{base}}$ and novel-class features $\mathbf{F}^{(s)}_{\text{novel}}$ by aggregating their generated parameters over the batch, scanning directions, and spatial tokens:
\begin{equation}
\begin{aligned}
\label{equ:parameter_base_novel}
\bar{\mathbf{P}}_\text{b} 
&= 
\frac{1}{|\mathbf{F}^{(s)}_{\text{base}}| \cdot K \cdot L}
\sum_{\mathbf{F}_i \in \mathbf{F}^{(s)}_{\text{base}}}
\sum_{k=1}^{K}
\sum_{\ell=1}^{L}
\mathbf{P}_{i,k,\ell,:}, \\
\bar{\mathbf{P}}_\text{n} 
&= 
\frac{1}{|\mathbf{F}^{(s)}_{\text{novel}}| \cdot K \cdot L}
\sum_{\mathbf{F}_i \in \mathbf{F}^{(s)}_{\text{novel}}}
\sum_{k=1}^{K}
\sum_{\ell=1}^{L}
\mathbf{P}_{i,k,\ell,:},
\end{aligned}
\end{equation}
where $\bar{\mathbf{P}}_\text{b}$ and $\bar{\mathbf{P}}_\text{n}$ denote the globally aggregated parameters for base-class and novel-class features, respectively. $\mathbf{P}$ corresponds to one of the generated parameter types $\mathbf{B}$, $\mathbf{C}$, or $\mathbf{\Delta}$. For each feature map, $\mathbf{P}_{i,k,\ell,:}\in\mathbb{R}^{D_P}$ is the parameter vector generated by $g_{\text{inc}}$ at scanning direction $k$ and spatial token $\ell$, where $D_P$ denotes the dimensionality of the corresponding parameter type.

We then minimize the absolute cosine similarity between these averaged parameters. This geometric constraint drives the parameter vectors toward orthogonality, reducing their representational overlap without encouraging a strong negative correlation. The separation loss is:
\begin{equation}
\label{equ:loss_sep}
\mathcal{L}_{\text{sep}}^{(s)} = 
\left|\frac{\bar{\mathbf{B}}_\text{b} \cdot \bar{\mathbf{B}}_\text{n}}{\|\bar{\mathbf{B}}_\text{b}\| \|\bar{\mathbf{B}}_\text{n}\|}\right| 
+ 
\left|\frac{\bar{\mathbf{C}}_\text{b} \cdot \bar{\mathbf{C}}_\text{n}}{\|\bar{\mathbf{C}}_\text{b}\| \|\bar{\mathbf{C}}_\text{n}\|}\right| 
+ 
\left|\frac{\bar{\mathbf{\Delta}}_\text{b} \cdot \bar{\mathbf{\Delta}}_\text{n}}{\|\bar{\mathbf{\Delta}}_\text{b}\| \|\bar{\mathbf{\Delta}}_\text{n}\|}\right|.
\end{equation}

As shown in the lower part of Fig.~\ref{fig:framework}~(c), minimizing $\mathcal{L}_\text{sep}$ reduces the cosine similarity between the averaged novel-class parameters $(\bar{\mathbf{B}}_\text{n}, \bar{\mathbf{C}}_\text{n}, \bar{\mathbf{\Delta}}_\text{n})$ and their base-class counterparts $(\bar{\mathbf{B}}_\text{b}, \bar{\mathbf{C}}_\text{b}, \bar{\mathbf{\Delta}}_\text{b})$. This encourages more separated parameter-generation patterns for the two input groups and enables $g_{\text{inc}}$ to learn novel-specific transformations while reducing interference with base-class representations.

\subsection{Optimization}
\label{sec:method_optim}
Given the final representation from Eq.~\ref{equ:mu_0} for the base session or Eq.~\ref{equ:mu_t} for incremental sessions, we perform classification using the fixed ETF classifier. Following \cite{yang2023neural,seo2024learning}, we use the \textit{dot regression} (DR) loss \cite{yang2022inducing} as the training objective, as it is shown to outperform cross-entropy loss when using a fixed ETF classifier under the class imbalance typical in FSCIL. Given the $l_2$-normalized representation $\hat{\boldsymbol{\mu}}_i = \frac{\boldsymbol{\mu}_i}{\|\boldsymbol{\mu}_i\|_2}$ and its corresponding label $y_i$, the DR loss is defined as:
\begin{equation}
\label{equ:loss_dr}
\mathcal{L}_{\text{DR}}\left(\hat{\boldsymbol{\mu}}_i, \hat{\mathbf{W}}_{\text{ETF}}\right) = \frac{1}{2} \left( \hat{\mathbf{w}}_{y_i}^\top \hat{\boldsymbol{\mu}}_i - 1 \right)^2,
\end{equation}
where $\hat{\mathbf{w}}_{y_i}$ denotes the fixed prototype vector for class $y_i$ in the ETF classifier $\hat{\mathbf{W}}_{\text{ETF}}$.

\subsubsection{Base Session Training}
\label{sec:method_optim_base}
During the base session ($s=0$), the goal is to learn a stable representation space from the base-session training data. We jointly optimize the backbone $f$, the identity branch $p_{\text{iden}}$, and the base selective SSM branch $g_{\text{base}}$ using the base-session feature representations defined in Eq.~\ref{equ:features_backbone_base}. The objective is to minimize the DR loss:
\begin{align}
\label{equ:optim_base}
\min_{f, p_{\text{iden}}, g_{\text{base}}}
\frac{1}{|\mathbf{F}^{(0)}|}
\sum_{i=1}^{|\mathbf{F}^{(0)}|}
\mathcal{L}_\text{DR}\left(
\hat{\boldsymbol{\mu}}_i^{(0)}, 
\hat{\mathbf{W}}_{\rm ETF}
\right).
\end{align}
After this stage, $f$, $p_{\text{iden}}$, and $g_{\text{base}}$ are frozen to preserve the learned base representation space.

\subsubsection{Incremental Sessions Training}
\label{sec:method_optim_inc}
For each incremental session ($s \ge 1$), the goal is to adapt to the current-session classes while preserving performance on previously learned classes. Only the incremental branch $g_{\text{inc}}$ is optimized, while $f$, $p_{\text{iden}}$, and $g_{\text{base}}$ remain frozen. Given the mixed feature-level batch $\mathbf{F}^{(s)}$ constructed in Sec.~\ref{sec:method_overall_arch}, we compute the final representation $\boldsymbol{\mu}^{(s)}$ using Eq.~\ref{equ:mu_t} and minimize the DR classification loss:

\begin{equation}
\label{equ:optim_inc_cls}
\mathcal{L}_\text{cls}^{(s)}
=
\frac{1}{|\mathbf{F}^{(s)}|}
\sum_{i=1}^{|\mathbf{F}^{(s)}|}
\mathcal{L}_\text{DR}\left(
\hat{\boldsymbol{\mu}}_i^{(s)}, 
\hat{\mathbf{W}}_{\rm ETF}
\right).
\end{equation}

To enforce class-sensitive separation, we incorporate the class-sensitive regularization losses defined in Sec.~\ref{sec:method_selective_scan}. The total objective for session $s$ is:
\begin{equation}
\label{equ:optim_inc}
\min_{g_{\text{inc}}} \left( \mathcal{L}_{\text{cls}}^{(s)} + \underbrace{\lambda_1 \mathcal{L}_{\text{supp, base}}^{(s)} + \lambda_2 \mathcal{L}_{\text{supp, novel}}^{(s)}}_{\text{Suppression Loss}} + \underbrace{\lambda_3 \mathcal{L}_{\text{sep}}^{(s)}}_{\text{Separation Loss}} \right),
\end{equation}
where $\lambda_1, \lambda_2, \lambda_3$ are hyperparameters.

\section{Experiments}
\label{sec:exps}

\subsection{Performance on Benchmarks}
\label{sec:exps_overall}

\begin{table*}[t!]
\renewcommand\arraystretch{1.1}
\begin{center}
\centering
\caption{Performance on miniImageNet across sessions. }
\vspace{-2mm}
\resizebox{0.95\textwidth}{!}{
\begin{tabular}{lccccccccccccl}
\toprule
\multicolumn{1}{l}{\multirow{2}{*}{\bf Methods}}&\multicolumn{1}{c}{\multirow{2}{*}{\bf Venue}} &\multirow{2}{*}{\bf Backbone}&\multicolumn{9}{c}{\bf Accuracy in each session (\%) }&\multirow{2}{*}{\bf \textsc{AVG}}&  \multirow{2}{*}{\textbf{\textsc{PD}}}\\ 
\cmidrule{4-12}
&  &&\bf 0&\bf 1&\bf 2&\bf 3&\bf 4&\bf 5&\bf 6&\bf 7&\bf 8&&    \\ 
\midrule
TOPIC~\cite{tao2020few}&CVPR'20&R-18&61.31&50.09&45.17&41.16&37.48&35.52&32.19&29.46&24.42&39.64&  36.89\\
LEC-Net~\cite{yang2021learnable}&arXiv'22&R-18&61.31&35.37&36.66&38.59&33.90&35.89&36.12&32.97&30.55&37.92&  30.76\\
SFbFSCIL~\cite{cheraghian2021synthesized}&ICCV'21&R-18& 61.40&59.80&54.20&51.69&49.45&48.00&45.20&43.80&42.10&50.63&  19.30\\
SPPR~\cite{zhu2021self}&CVPR'21&R-18&61.45&63.80&59.53&55.53&52.50&49.60&46.69&43.79&41.92&52.76&  19.53\\
IDLVQ~\cite{chen2021incremental}&ICLR'21&R-18&64.77&59.87&55.93&52.62&49.88&47.55&44.83&43.14&41.84&51.16&  22.93\\
DSN~\cite{yang2022dynamic}&TPAMI'22&R-18&68.95&63.46&59.78&55.64&52.85&51.23&48.90&46.78&45.89&54.83&  23.06\\
Data-free~\cite{liu2022few}& ECCV'22&R-18&71.84&67.12&63.21&59.77&57.01&53.95&51.55&49.52&48.21& 58.02&  23.63\\
CEC~\cite{zhang2021few}&CVPR'21&R-18&72.00&66.83&62.97&59.43&56.70&53.73&51.19&49.24&47.63&57.75&  24.37\\
MetaFSCIL~\cite{chi2022metafscil}&CVPR'22&R-18&72.04&67.94&63.77&60.29&57.58&55.16&52.90&50.79&49.19&58.85&  22.85\\
LIMIT~\cite{zhou2022few}&TPAMI'22&R-18&72.32&68.47&64.30&60.78&57.95&55.07&52.70&50.72&49.19& 59.06&  23.13\\
FACT~\cite{zhou2022forward}&CVPR'22&R-18&72.56&69.63&66.38&62.77&60.60&57.33&54.34&52.16&50.49&60.70&  22.07\\
TEEN~\cite{wang2024few}&NeurIPS'23&R-18&73.53& 70.55& 66.37& 63.23& 60.53& 57.95& 55.24& 53.44& 52.08& 61.44&  21.45\\
CABD~\cite{zhao2023few}&CVPR'23&R-18&74.65&70.43&66.29&62.77&60.75&57.24&54.79&53.65&52.22&61.42&  22.43\\
\rowcolor{mygray}\textbf{Mamba-FSCIL}&&R-18&74.72&71.18&67.96&64.33&61.54&58.09&55.47&53.53&52.11&62.10&  22.61\\ \midrule
CLOSER~\cite{oh2024closer}&ECCV'24&R-18& 76.02& 71.61& 67.99& 64.69& 61.70& 58.94& 56.23& 54.52& 53.33& 62.78&  22.69\\
C-FSCIL~\cite{hersche2022constrained}&CVPR'22&R-12 &76.40&71.14&66.46&63.29&60.42&57.46&54.78&53.11&51.41&61.61&  24.99\\
Regularizer~\cite{akyurek2021subspace}&ICLR'22&R-18&80.37& 74.68& 69.39& 65.51& 62.38& 59.03& 56.36& 53.95& 51.73& 63.71&  28.64\\
ALICE~\cite{peng2022few}&ECCV'22&R-18&80.60& 70.60& 67.40& 64.50& 62.50& 60.00& 57.80& 56.80& 55.70& 63.99&  24.90\\
\rowcolor{mygray}\textbf{Mamba-FSCIL}&&R-18&80.62&74.40&69.81&67.49&64.57&62.24&58.59&56.84&55.80&65.60&  24.82\\ \midrule
SAVC~\cite{song2023learning}& CVPR'23&R-18& 81.12&76.14&72.43&68.92&66.48&62.95&59.92&58.39&57.11&67.05&   24.01\\
ALFSCIL~\cite{li2024analogical}&TCSVT'24&R-18&81.27&75.97&70.97&66.53&63.46&59.95&56.93&54.81&53.31&64.80&  27.96\\
FeSSSS~\cite{ahmad2022few}&CVPR'22&R-18+R-50&81.50&77.04&72.92&69.56&67.27&64.34&62.07&60.55&58.87&68.23&  22.63\\
KRRM~\cite{wang2023improved}&TCSVT'24&R-18&82.65&77.82&73.59&70.24&67.74&64.82&61.91&59.96&58.35&68.56&  24.30\\
\rowcolor{mygray}\textbf{Mamba-FSCIL}& &R-18&82.77&78.77&73.87&70.59&68.17&65.19&62.00&60.24&58.69&68.92&  24.08\\ \midrule
YourSelf~\cite{tang2024rethinking}& ECCV'24&ViT (10.0M)&84.00&77.60&73.70&70.00&68.00&64.90&62.10&59.80&59.00&68.80&  25.00\\
NC-FSCIL~\cite{yang2023neural}& ICLR'23&R-12 &84.02&76.80&72.00&67.83&66.35&64.04&61.46&59.54&58.31&67.82&  25.71\\
\rowcolor{mygray}\textbf{Mamba-FSCIL}
& & R-12
& \makecell{84.54\\{\scriptsize $\pm$0.15}}
& \makecell{79.23\\{\scriptsize $\pm$0.14}}
& \makecell{74.68\\{\scriptsize $\pm$0.90}}
& \makecell{71.42\\{\scriptsize $\pm$0.60}}
& \makecell{68.85\\{\scriptsize $\pm$0.31}}
& \makecell{65.89\\{\scriptsize $\pm$0.29}}
& \makecell{62.77\\{\scriptsize $\pm$0.10}}
& \makecell{61.01\\{\scriptsize $\pm$0.32}}
& \makecell{59.29\\{\scriptsize $\pm$0.15}}
& \makecell{69.74\\{\scriptsize $\pm$0.21}}
& \makecell{25.25\\{\scriptsize $\pm$0.29}}
\\
\bottomrule
\end{tabular}
}
\label{table:imgnet}
\end{center}
\vspace{-4mm}
\end{table*}

\begin{table*}[t!]
\renewcommand\arraystretch{1.1}
\begin{center}
\centering
\caption{Performance on CIFAR-100 across sessions. }
\vspace{-2mm}
\resizebox{0.95\textwidth}{!}{
\begin{tabular}{lccccccccccccl}
\toprule
\multicolumn{1}{l}{\multirow{2}{*}{\bf Methods}}&\multicolumn{1}{l}{\multirow{2}{*}{\bf Venue}} &\multirow{2}{*}{\bf Backbone}&\multicolumn{9}{c}{\bf Accuracy in each session (\%) }&\multirow{2}{*}{\bf \textsc{AVG}}&  \multirow{2}{*}{\textbf{\textsc{PD}}}\\ 
\cmidrule{4-12}
&  &&\bf 0&\bf 1&\bf 2&\bf 3&\bf 4&\bf 5&\bf 6&\bf 7&\bf 8&&  \\ 
\midrule
TOPIC~\cite{tao2020few}& CVPR'20&R-18&64.10&55.88&47.07&45.16&40.11&36.38&33.96&31.55&29.37&42.62&  34.73\\
LEC-Net~\cite{yang2021learnable}& arXiv'22&R-18&64.10&53.23&44.19&41.87&38.54&39.54&37.34&34.73&34.73&43.14&  29.37\\
DSN~\cite{yang2022dynamic}&TPAMI'22&R-18&73.00&68.83&64.82&62.24&59.16&56.96&54.04&51.57&49.35&60.00&  23.65\\
CEC~\cite{zhang2021few}& CVPR'21&R-20 &73.07&68.88&65.26&61.19&58.09&55.57&53.22&51.34&49.14&59.53&  23.93\\
LIMIT~\cite{zhou2022few}& TPAMI'22&R-20 &73.81&72.09&67.87&63.89&60.70&57.77&55.67&53.52&51.23&61.84&  22.58\\
\rowcolor{mygray}\textbf{Mamba-FSCIL}& &R-18&73.58&72.18&68.39&64.17&61.86&58.19&56.47&54.85&52.50& 62.47&  21.08\\ \midrule
Data-free~\cite{liu2022few}&ECCV'22&R-20 &74.40&70.20&66.54&62.51&59.71&56.58&54.52&52.39&50.14& 60.78&  24.26\\
MetaFSCIL~\cite{chi2022metafscil}&CVPR'22&R-20 &74.50& 70.10& 66.84& 62.77& 59.48& 56.52& 54.36& 52.56& 49.97& 60.79&  24.53\\
FACT~\cite{zhou2022forward}& CVPR'22&R-20 &74.60&72.09&67.56&63.52&61.38&58.36&56.28&54.24&52.10&62.24&  22.50\\
TEEN~\cite{wang2024few}&NeurIPS'23&R-18&74.92& 72.65& 68.74& 65.01& 62.01& 59.29& 57.90& 54.76& 52.64& 63.10&  22.28\\
\rowcolor{mygray}\textbf{Mamba-FSCIL}& &R-18&74.92&72.58&69.21&65.04&63.06&59.94&58.16&55.75&52.74&63.49&  22.18\\ \midrule
FeSSSS~\cite{ahmad2022few}& CVPR'22&R-18+R-50&75.35&70.81&66.70&62.73&59.62&56.45&54.33&52.10&50.23&60.92&  25.12\\
C-FSCIL~\cite{hersche2022constrained}&CVPR'22&R-12 &77.47& 72.40& 67.47& 63.25& 59.84& 56.95& 54.42& 52.47& 50.47& 61.64&  27.00\\
SAVC~\cite{song2023learning}&CVPR'23&R-20 &78.77&73.31&69.31&64.93&61.70&59.25&57.13&55.19&53.12&63.63&  25.65\\
ALICE~\cite{peng2022few}& ECCV'22&R-18& 79.00&70.50&67.10&63.40&61.20&59.20&58.10&56.30&54.10&63.21&   24.90\\
CABD~\cite{zhao2023few}& CVPR'23&R-18&79.45&75.38&71.84&67.95&64.96&61.95&60.16&57.67&55.88&66.14&  23.57\\
ALFSCIL~\cite{li2024analogical}& TCSVT'24&R-18&80.75&77.88&72.94&68.79&65.33&62.15&60.02&57.68&55.17&66.75&  25.58\\
\rowcolor{mygray}\textbf{Mamba-FSCIL}& &R-18&80.50&77.25&73.73&68.90&65.41&62.44&61.03&58.34&56.59&67.13&  23.91\\ \midrule
NC-FSCIL~\cite{yang2023neural}& ICLR'23&R-12 &82.52&76.82&73.34&69.68&66.19&62.85&60.96&59.02&56.11&67.50&  26.41\\
YourSelf~\cite{tang2024rethinking}& ECCV'24&ViT (10.0M)&82.90&76.30&72.90&67.80&65.20&62.00&60.70&58.80&56.60&67.02&  26.30\\
\rowcolor{mygray}\textbf{Mamba-FSCIL}& &R-18&82.78&77.82&74.17&69.77&66.51&62.81&61.24&59.64&57.45&68.02&  25.33\\
\rowcolor{mygray}\textbf{Mamba-FSCIL}
& & R-12
& \makecell{82.95\\{\scriptsize $\pm$0.13}}
& \makecell{77.87\\{\scriptsize $\pm$0.16}}
& \makecell{74.12\\{\scriptsize $\pm$0.37}}
& \makecell{69.67\\{\scriptsize $\pm$0.30}}
& \makecell{66.75\\{\scriptsize $\pm$0.17}}
& \makecell{63.62\\{\scriptsize $\pm$0.13}}
& \makecell{61.44\\{\scriptsize $\pm$0.13}}
& \makecell{59.80\\{\scriptsize $\pm$0.33}}
& \makecell{57.52\\{\scriptsize $\pm$0.23}}
& \makecell{68.19\\{\scriptsize $\pm$0.11}}
& \makecell{25.43\\{\scriptsize $\pm$0.24}}
\\
\bottomrule
\end{tabular}
}
\label{table:cifar}
\end{center}
\vspace{-4mm}
\end{table*}

\begin{table*}[ht!] 
\renewcommand\arraystretch{1.1}
\begin{center}
\centering
\caption{Performance on CUB-200 across sessions. }
\vspace{-2mm}
\resizebox{0.95\textwidth}{!}{
\begin{tabular}{lccccccccccccccl}
\toprule
\multicolumn{1}{l}{\multirow{2}{*}{\bf Methods}}&\multicolumn{1}{l}{\multirow{2}{*}{\bf Venue}} &\multirow{2}{*}{\bf Backbone}&\multicolumn{11}{c}{\bf Accuracy in each session (\%) }&\multirow{2}{*}{\bf \textsc{AVG}}&   \multirow{2}{*}{\textbf{\textsc{PD}}}\\ 
\cmidrule{4-14}
& &&\bf 0&\bf 1&\bf 2&\bf 3&\bf 4&\bf 5&\bf 6&\bf 7&\bf 8&\bf 9&\bf 10&&   \\ 
\midrule
SPPR~\cite{zhu2021self}& CVPR'21&R-18&68.68&61.85&57.43&52.68&50.19&46.88&44.65&43.07&40.17&39.63&37.33&49.32&   31.35\\
TOPIC~\cite{tao2020few}& CVPR'20&R-18&68.68&62.49&54.81&49.99&45.25&41.40&38.35&35.36&32.22&28.31&26.28&43.92&   42.40\\
SFbFSCIL~\cite{cheraghian2021synthesized}& ICCV'21&R-18&68.78&59.37&59.32&54.96&52.58&49.81&48.09&46.32&44.33&43.43&43.23&51.84&   25.55\\
LEC-Net~\cite{yang2021learnable}& arXiv'22&R-18&70.86&58.15&54.83&49.34&45.85&40.55&39.70&34.59&36.58&33.56&31.96&45.09&   38.90\\
\rowcolor{mygray}\textbf{Mamba-FSCIL}& &R-18&70.38&63.86&60.26&60.94&57.55&56.98&57.82&55.06&53.40&53.26&52.87&58.40&   17.51\\ \midrule
MgSvF~\cite{zhao2021mgsvf}& TPAMI'21&R-18&72.29&70.53&67.00&64.92&62.67&61.89&59.63&59.15&57.73&55.92&54.33&62.37&   17.96\\
CEC~\cite{zhang2021few}& CVPR'21&R-18&75.85&71.94&68.50&63.50&62.43&58.27&57.73&55.81&54.83&53.52&52.28&61.33&   23.57\\
Data-free~\cite{liu2022few}& ECCV'22&R-18&75.90&72.14&68.64&63.76&62.58&59.11&57.82&55.89&54.92&53.58&52.39&61.52&   23.51\\
MetaFSCIL~\cite{chi2022metafscil}& CVPR'22&R-18&75.90& 72.41& 68.78& 64.78& 62.96& 59.99& 58.30& 56.85& 54.78& 53.82& 52.64& 61.93&   23.26\\
FACT~\cite{zhou2022forward}&CVPR'22&R-18&75.90&73.23&70.84&66.13&65.56&62.15&61.74&59.83&58.41&57.89&56.94& 64.42&   18.96\\
DSN~\cite{yang2022dynamic}& TPAMI'22&R-18&76.06&72.18&69.57&66.68&64.42&62.12&60.16&58.94&56.99&55.10&54.21&63.31&   21.85\\
LIMIT~\cite{zhou2022few}& TPAMI'22&R-18&76.32&74.18&72.68&69.19&68.79&65.64&63.57&62.69&61.47&60.44&58.45&66.67&   17.87\\
TEEN~\cite{wang2024few}&NeurIPS'23&R-18&77.26&76.13&72.81&68.16&67.77&64.40&63.25&62.29&61.19&60.32&59.31&66.63&   17.95\\
IDLVQ~\cite{chen2021incremental}& ICLR'21&R-18& 77.37& 74.72&70.28&67.13&65.34&63.52&62.10&61.54&59.04&58.68&57.81&65.23&    19.56\\
ALICE~\cite{peng2022few}& ECCV'22&R-18&77.40&72.70&70.60&67.20&65.90&63.40&62.90&61.90&60.50&60.60&60.10&65.75&   17.30\\
\rowcolor{mygray}\textbf{Mamba-FSCIL}& &R-18& 77.34&74.67& 71.55&68.54&67.19&65.26&63.74&62.96&61.10&60.88&60.51&66.70&    16.83\\ \midrule
FeSSSS~\cite{ahmad2022few}&CVPR'22&R-18+R-50&79.60&73.46&70.32&66.38&63.97&59.63&58.19&57.56&55.01&54.31&52.98&62.85&   26.62\\
ALFSCIL~\cite{li2024analogical}& TCSVT'24&R-18&79.79&76.53&73.12&69.02&67.62&64.76&63.45&62.32&60.83&60.21&59.30&67.00&   20.49\\
NC-FSCIL~\cite{yang2023neural}& ICLR'23&R-18&80.45&75.98&72.30&70.28&68.17&65.16&64.43&63.25&60.66&60.01&59.44&67.28&   21.01\\
\rowcolor{mygray}\textbf{Mamba-FSCIL}
& & R-18
& \makecell{80.92\\{\scriptsize $\pm$0.04}}
& \makecell{76.28\\{\scriptsize $\pm$0.07}}
& \makecell{73.03\\{\scriptsize $\pm$0.08}}
& \makecell{70.12\\{\scriptsize $\pm$0.03}}
& \makecell{67.77\\{\scriptsize $\pm$0.12}}
& \makecell{65.72\\{\scriptsize $\pm$0.06}}
& \makecell{65.29\\{\scriptsize $\pm$0.13}}
& \makecell{64.05\\{\scriptsize $\pm$0.07}}
& \makecell{62.35\\{\scriptsize $\pm$0.04}}
& \makecell{62.14\\{\scriptsize $\pm$0.06}}
& \makecell{61.52\\{\scriptsize $\pm$0.11}}
& \makecell{68.11\\{\scriptsize $\pm$0.02}}
& \makecell{19.40\\{\scriptsize $\pm$0.14}}
\\

\bottomrule
\end{tabular}
}
\label{table:cub}
\end{center}
\vspace{-2mm}
\end{table*}

\textbf{Datasets.} We employ three benchmark datasets. miniImageNet~\cite{russakovsky2015imagenet} is a variant of ImageNet with 100 classes, each having 500 training and 100 testing images of $84 \times 84$ pixels. CIFAR-100~\cite{krizhevsky2009learning} has the same number of classes and images, and the image size is $32\times32$. CUB-200~\cite{wah2011caltech} is a fine-grained classification dataset consisting of 11,788 images in 200 classes, with an image resolution of $224 \times 224$. We follow the standard experimental settings in FSCIL~\cite{tao2020few,zhang2021few,yang2023neural}. For miniImageNet and CIFAR-100, the base session includes 60 classes, followed by 8 incremental sessions, each with a 5-way 5-shot setting (5 classes and 5 images per class). For CUB-200, the base session comprises 100 classes, with 10 incremental sessions, each following a 10-way 5-shot setting.

\textbf{Evaluation Metrics}. Following standard protocols~\cite{tao2020few,zhang2021few}, we denote by $\mathcal{A}_s$ the Top-1 accuracy at session $s$ among a total of $S$ sessions. We evaluate performance using: \textsc{\textbf{Base}} session accuracy ($\mathcal{A}_0$); \textsc{\textbf{Final}} session accuracy ($\mathcal{A}_S$); \textbf{Average Accuracy} (\textsc{AVG}), computed as the mean accuracy across all sessions; and \textbf{Performance Drop} (\textsc{PD}), defined as $\mathcal{A}_0 - \mathcal{A}_S$ to quantify forgetting. 

\textbf{Training Details.}
We conduct experiments on three benchmark datasets using both CNN-based and transformer-based vision backbones. Standard data augmentation techniques, including random resizing, horizontal flipping, and color jittering, are applied following prior FSCIL works~\cite{tao2020few,zhang2021few,peng2022few,yang2023neural}.
The batch size is set to 512 during the base session and 64 during incremental sessions. 
We use SGD with a momentum of 0.9 and a weight decay of 0.0005, and apply cosine annealing for the learning rate schedule.
The dimensions of the input-dependent parameters in Eq.~\ref{equ:ssm_branch_directional_scanning} are set to $D_B=D_C=D_{\Delta}=256$. The output dimension $D'$ of the SSM branch in Eq.~\ref{equ:ssm_branch_backbone_addpos} is set to 512 for CIFAR-100 and CUB-200, and 1024 for miniImageNet. All experiments are performed using 8 NVIDIA A100 GPUs.

Architectural choices and loss weights are selected using only the base-session training data. More details are provided in Appendix~\ref{supp:hyperparameter_protocol}.
For the final Mamba-FSCIL row in Tables~\ref{table:imgnet}, \ref{table:cifar}, \ref{table:cub} and \ref{table:cub_transformer}, we report mean and standard deviation over three independent seeds. Unless otherwise specified, all other experiments are conducted with a single fixed seed (seed 0).

\subsubsection{CNN Backbone Results}
\label{sec:exps_overall_cnn}
We evaluate Mamba-FSCIL with standard CNN backbones, including ResNet-12 and ResNet-18, on miniImageNet, CIFAR-100, and CUB-200. Implementation details are provided in Appendix~\ref{supp:details_cnn}. As shown in Tables~\ref{table:imgnet}, \ref{table:cifar}, and \ref{table:cub}, Mamba-FSCIL achieves strong final-session and average accuracy across different settings.

On \textbf{miniImageNet}, Mamba-FSCIL with ResNet-12 achieves the highest \textsc{AVG} accuracy ($69.74{\pm}0.21\%$) and \textsc{Final} accuracy ($59.29{\pm}0.15\%$), improving over NC-FSCIL by 1.92 and 0.98 points, respectively. Using ResNet-18 with a matched base accuracy (80.62\% vs. ALICE's 80.60\%), Mamba-FSCIL also obtains higher \textsc{Final} accuracy (55.80\% vs. 55.70\%) and \textsc{AVG} accuracy (65.60\% vs. 63.99\%).

On \textbf{CIFAR-100}, Mamba-FSCIL with ResNet-12 achieves the highest \textsc{Final} accuracy ($57.52{\pm}0.23\%$) and \textsc{AVG} accuracy ($68.19{\pm}0.11\%$). In final-session accuracy, it improves over NC-FSCIL and FeSSSS by 1.41 and 7.29 points, respectively. With ResNet-18 at a matched base accuracy (73.58\%, close to LIMIT's 73.81\%), Mamba-FSCIL achieves higher \textsc{Final} accuracy and lower \textsc{PD}.

On the fine-grained \textbf{CUB-200} dataset, Mamba-FSCIL achieves an \textsc{AVG} accuracy of $68.11{\pm}0.02\%$ and a \textsc{Final} accuracy of $61.52{\pm}0.11\%$ in the final three-seed setting. It improves over DSN by 7.31 points in final-session accuracy. With a matched base accuracy of 77.34\% compared to ALICE's 77.40\%, Mamba-FSCIL obtains higher \textsc{Final} accuracy (60.51\% vs. 60.10\%) and lower \textsc{PD} (16.83\% vs. 17.30\%).

\begin{table*}[t!] 
\renewcommand\arraystretch{1.1}
\begin{center}
\centering
\caption{Performance on CUB-200 using Transformer backbones. Results for CLOM~\cite{zou2022margin} are from Comp-FSCIL~\cite{zou2024compositional}. NC-FSCIL~\cite{yang2023neural} is our re-implementation.}
\vspace{-2mm}
\resizebox{0.95\textwidth}{!}{
\begin{tabular}{lllccccccccccccc}
\toprule
\multicolumn{1}{l}{\multirow{2}{*}{\bf Methods}}&\multicolumn{1}{l}{\multirow{2}{*}{\bf Venue}} &\multirow{2}{*}{\bf Backbone}&\multicolumn{11}{c}{\bf Accuracy in each session (\%) }&\multirow{2}{*}{\bf \textsc{AVG}}&\multirow{2}{*}{\bf \textsc{PD}}\\ 
\cmidrule{4-14}
& &&\bf 0&\bf 1&\bf 2&\bf 3&\bf 4&\bf 5&\bf 6&\bf 7&\bf 8&\bf 9&\bf 10&&\\ 
\midrule
CLOM~\cite{zou2022margin} & NeurIPS'22 &Swin-T (IN1K)&86.28 &82.85 &80.61 &77.79 &76.34 &74.64 &73.62 &72.82 &71.24 &71.33 &70.50 & 76.18&15.78\\
NC-FSCIL~\cite{yang2023neural}& ICLR'23 &Swin-T (IN1K)&87.53 &84.25 &81.72 &79.10 &77.21 &75.52 &74.51 &74.42 &72.26 &72.86 &72.49&77.44&15.04\\
Comp-FSCIL~\cite{zou2024compositional}& ICML'24 &Swin-T (IN1K)&87.67 &84.73 &83.03 &80.04 &77.73 &75.52 &74.32 &74.55 &73.35 &73.15 &72.80&77.90&14.87\\
CPE-CLIP~\cite{d2023multimodal}&  ICCVW'23&ViT-B/16 (CLIP)&81.58&78.52&76.68& 71.86&71.52&70.23&67.66&66.52&65.09&64.47&64.60& 70.79&16.98\\
PriViLege~\cite{park2024pre}&CVPR'24& ViT-B/16 (IN21K)& 82.21& 81.25& 80.45& 77.76& 77.78& 75.95& 75.69& 76.00& 75.19& 75.19&75.08&77.50&7.13\\
\rowcolor{mygray}\textbf{Mamba-FSCIL} &  &Swin-T (IN1K)&88.13&85.14&83.41& 80.77&77.23&75.73&75.70&75.32&74.18&74.26&74.13&78.55&14.00\\
\rowcolor{mygray}\textbf{Mamba-FSCIL}&& ViT-B/16 (CLIP)& 87.05& 84.06& 82.83& 81.01& 80.31& 78.64& 78.05& 78.46& 77.16& 77.16&76.92&80.15&10.13\\
\rowcolor{mygray}\textbf{Mamba-FSCIL}
&& ViT-B/16 (IN21K)
& \makecell{88.17\\{\scriptsize $\pm$0.07}}
& \makecell{86.96\\{\scriptsize $\pm$0.18}}
& \makecell{86.20\\{\scriptsize $\pm$0.26}}
& \makecell{84.09\\{\scriptsize $\pm$0.51}}
& \makecell{84.06\\{\scriptsize $\pm$0.68}}
& \makecell{82.71\\{\scriptsize $\pm$0.43}}
& \makecell{83.39\\{\scriptsize $\pm$0.57}}
& \makecell{83.40\\{\scriptsize $\pm$0.37}}
& \makecell{82.85\\{\scriptsize $\pm$0.81}}
& \makecell{82.99\\{\scriptsize $\pm$0.35}}
& \makecell{83.41\\{\scriptsize $\pm$0.39}}
& \makecell{84.39\\{\scriptsize $\pm$0.39}}
& \makecell{4.76\\{\scriptsize $\pm$0.38}}\\
\bottomrule
\end{tabular}
}
\label{table:cub_transformer}
\end{center}
\vspace{-4mm}
\end{table*}

\subsubsection{Transformer Backbone Results}
\label{sec:exps_overall_transformer}
We evaluate Mamba-FSCIL on the \textbf{CUB-200} dataset using three widely-used transformer backbones: \textbf{Swin-T (IN1K)}, \textbf{ViT-B/16 (CLIP)}, and \textbf{ViT-B/16 (IN21K)}, following recent works (e.g., Comp-FSCIL~\cite{zou2024compositional}, CPE-CLIP~\cite{d2023multimodal}, and PriViLege~\cite{park2024pre}). Our goal is to test whether the Selective SSM mechanism (1) remains effective on high-capacity pretrained transformers, and (2) can adapt general-purpose representations to the FSCIL paradigm while mitigating forgetting. Implementation details are provided in Appendix~\ref{supp:details_vit}.

As shown in Tab.~\ref{table:cub_transformer}, Mamba-FSCIL achieves the best or highly competitive performance across the evaluated transformer settings.
\textbf{With Swin-T (IN1K)}, it obtains the highest average accuracy (78.55\%) and lowest PD (14.00\%), surpassing Comp-FSCIL despite using a simpler training setup without compositional learning.  
\textbf{With CLIP-pretrained ViT-B/16}, it improves the average accuracy of CPE-CLIP by \textbf{+9.36\%} (80.15\% vs. 70.79\%) and reduces PD from \textbf{16.98\%} to \textbf{10.13\%}, without relying on multimodal prompts or text-driven regularization, demonstrating the effectiveness of our class-sensitive selective scanning on strong pretrained features. 
\textbf{With ViT-B/16 (IN21K)}, performance further increases to an average accuracy of $84.39{\pm}0.39\%$ with the lowest \textsc{PD} ($4.76{\pm}0.38\%$), showing that our method effectively leverages high-quality pretrained representations while preserving previously learned knowledge.

Overall, these results indicate that Mamba-FSCIL remains effective across different backbone architectures and can benefit from high-quality pretrained representations while maintaining low performance drop.

\subsubsection{Comparison with Dynamic Adaptation Method}
\label{sec:exps_overall_dsn}
\begin{figure}[t!]
    \centering
    \includegraphics[width=1\linewidth]{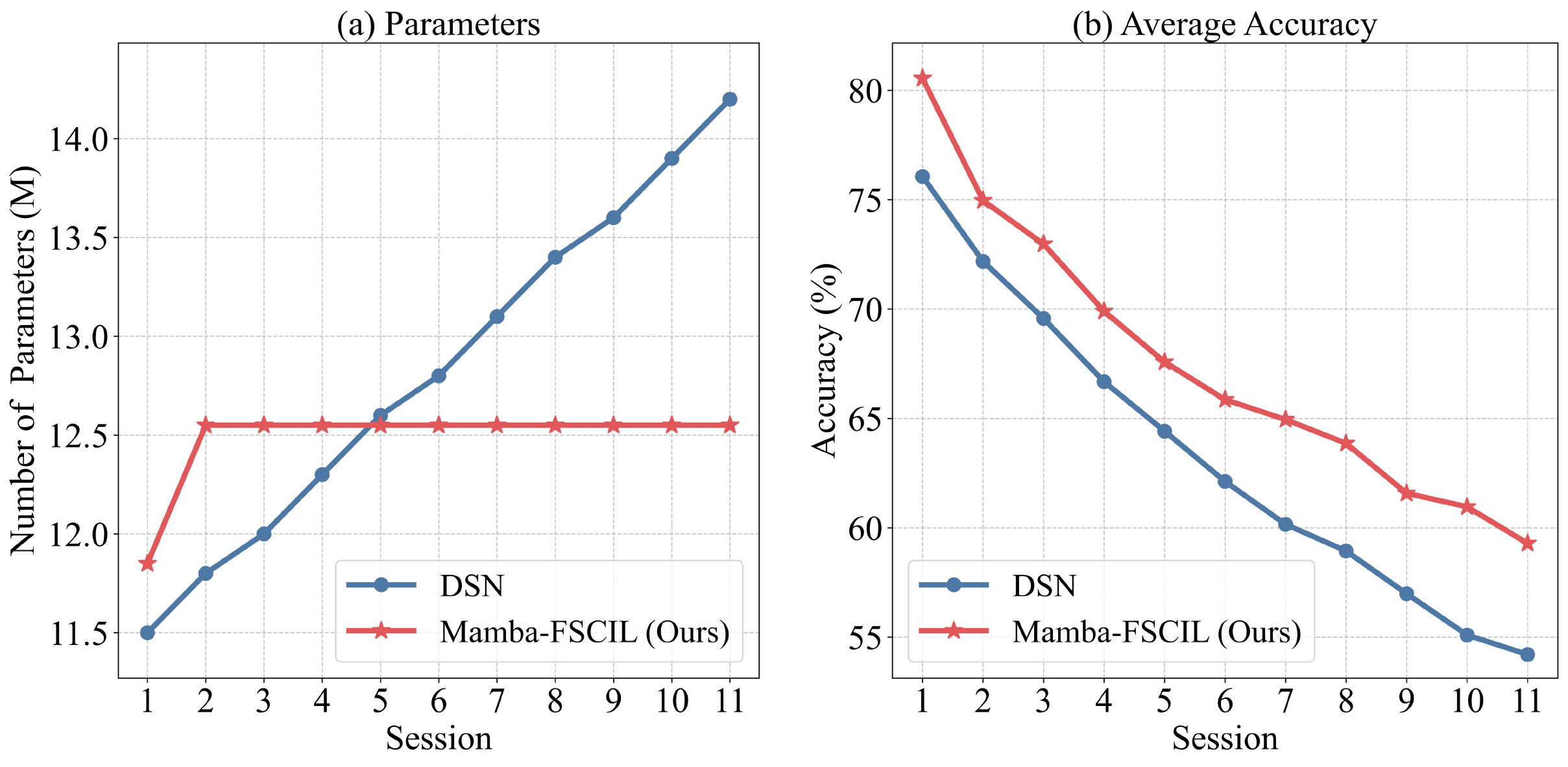}
    \vspace{-6mm}
    \caption{Comparison of Mamba-FSCIL and DSN \cite{yang2022dynamic} on CUB-200. (a) Parameter counts. (b) Accuracy across sessions.}
    \label{fig:dsn_compare}
    \vspace{-4mm}
\end{figure}

We compare Mamba-FSCIL with the expansion-based DSN~\cite{yang2022dynamic} on CUB-200 using a ResNet-18 backbone, focusing on session-wise parameter growth and accuracy. As shown in Fig.~\ref{fig:dsn_compare}(a), DSN increases its parameter count as new sessions arrive, whereas Mamba-FSCIL reuses the same dual selective SSM projector across sessions without adding new session-specific branches. Fig.~\ref{fig:dsn_compare}~(b) further shows that Mamba-FSCIL achieves higher accuracy across sessions, indicating a better accuracy-efficiency trade-off under this setting.
\vspace{-2mm}
\subsection{Ablation Studies}
\label{sec:exps_ablation}
\subsubsection{Impact of Individual Component}
\label{sec:exps_ablation_individual}
\begin{table*}[th!]
\begin{center}
\caption{Impact of dual selective SSM projector and class-sensitive selective scan on CIFAR-100 and miniImageNet.}
\vspace{-2mm}
\resizebox{\linewidth}{!}{
\begin{tabular}{l|cccc|cccc}
\toprule
\multicolumn{1}{l|}{\multirow{2}{*}{\textbf{Methods}}} & \multicolumn{4}{c}{\textbf{miniImageNet}}& \multicolumn{4}{c}{CIFAR-100}\\ 
 &  \textbf{\textsc{BASE}}& \textbf{\textsc{FINAL}}& \textbf{\textsc{AVG}}& \textbf{\textsc{PD}} &\textbf{\textsc{BASE}}& \textbf{\textsc{FINAL}}&\textbf{\textsc{AVG}}& \textbf{\textsc{PD}} \\ \midrule
Baseline &  83.83& 51.77& 63.09& 32.06&83.05&  49.50&63.51& 33.55\\
Single Selective SSM Projector &  84.35& 55.24& 66.79& 29.11&82.58&  55.01&66.90& 27.57\\
Dual Selective SSM Projector &  84.93& 58.92& 69.20& 26.01&82.80&  56.71&67.52& 26.09\\
Dual Selective SSM Projector + Class-sensitive Selective Scan  &  \textbf{84.93}& \textbf{59.36}& \textbf{69.81}& \textbf{25.57}&\textbf{82.80}&  \textbf{57.51}&\textbf{68.14}& \textbf{25.29}\\
\bottomrule
\end{tabular}
}
\label{tab:ablation}
\end{center}
\vspace{-4mm}
\end{table*}

We conduct ablation studies on miniImageNet and CIFAR-100. We compare four configurations with identical setups (same backbone, identity branch, memory module, classifier, and classification loss, with comparable parameter budgets to support a fair comparison):
(1) \textbf{Baseline} replaces the Selective Scan Mechanism in Fig.~\ref{fig:framework}~(b) with a 3-layer MLP, representing a static adaptation strategy commonly employed in existing methods.
(2) \textbf{Single Selective
SSM Projector} directly applies a single Mamba-based projector (with $p_\text{iden}$ and $g_\text{base}$) across all sessions. This setup helps us evaluate Mamba's inherent input-adaptive potential without structural separation or class-sensitive regularization.
(3) \textbf{Dual Selective
SSM Projector} introduces our dual-branch design, structurally decoupling base- and novel-class adaptation.
(4) \textbf{+ Class-Sensitive Selective Scan} adds our class-sensitive regularization via suppression and separation losses to further guide Mamba's dynamic processing.

As shown in Tab.~\ref{tab:ablation}, the \textbf{Baseline} obtains the lowest final and average accuracy, indicating that a static projection strategy is less effective under incremental updates. 
Replacing the MLP with a \textbf{Single Selective SSM Projector} improves performance, e.g., by +3.39 points in \textsc{AVG} and -5.98 points in \textsc{PD} on CIFAR-100, suggesting the benefit of input-dependent dynamic computation over static projection. 
The \textbf{Dual Selective SSM Projector} further improves the stability--plasticity trade-off; on miniImageNet, it increases \textsc{AVG} by 2.41 points and reduces \textsc{PD} by 3.10 points compared with the single-projector variant, supporting the usefulness of decoupling base- and novel-class processing. 
Finally, adding the \textbf{Class-Sensitive Selective Scan} yields the best results among the tested variants, reaching 59.36\% and 57.51\% \textsc{Final} accuracy on miniImageNet and CIFAR-100, respectively, with the lowest \textsc{PD} values of 25.57\% and 25.29\%. 
These results support the contribution of each component in Mamba-FSCIL.

\subsubsection{Effectiveness of Class-Sensitive Selective Scan}
\label{sec:exps_ablation_losses}
\begin{figure}[t!]
    \centering
    \includegraphics[width=1\linewidth]{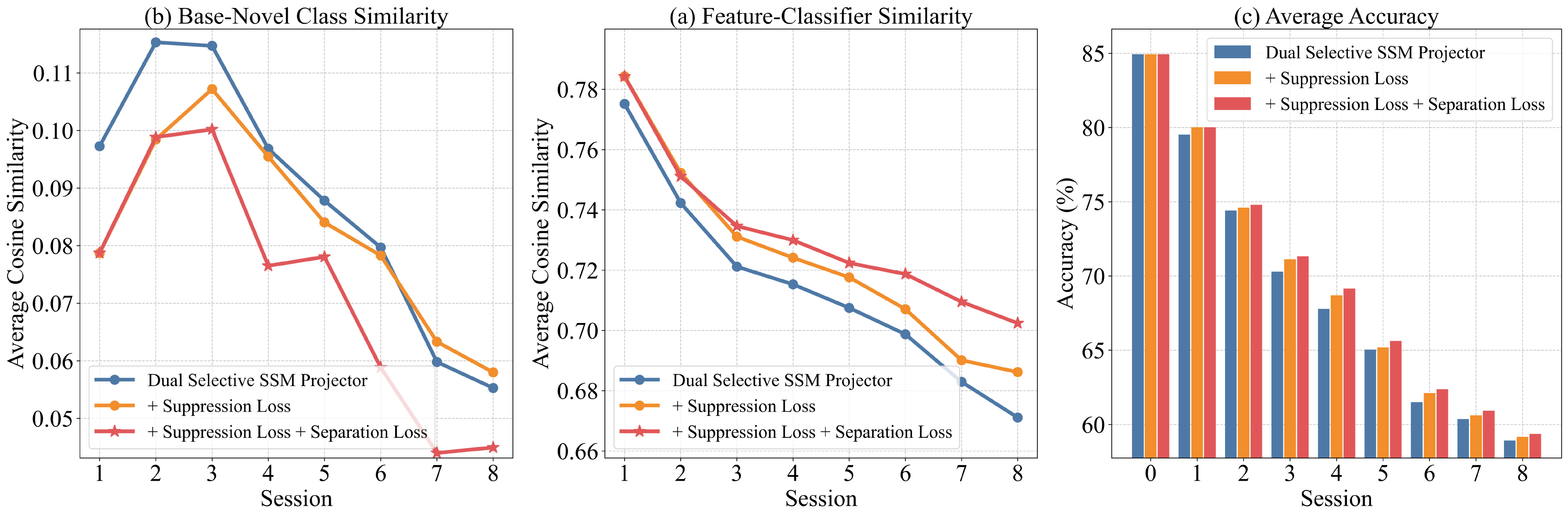}
    \vspace{-6mm}
    \caption{Effect of suppression ($\mathcal{L}_{\text{supp}}$) and separation ($\mathcal{L}_{\text{sep}}$) losses on miniImageNet test set features. (a) Feature–prototype alignment for base classes. (b) Base–novel feature similarity. (c) Accuracy trend across sessions.}
    \label{fig:selective_scan}
    \vspace{-4mm}
\end{figure}

To evaluate the effectiveness of the class-sensitive selective scan mechanism, we analyze the impact of the suppression loss ($\mathcal{L}_{\text{supp}}$) and separation loss ($\mathcal{L}_{\text{sep}}$) on (a) base-class stability, (b) base–novel discriminability, and (c) accuracy across sessions on the miniImageNet test set. Stability is quantified using the feature–prototype alignment (cosine similarity between base-class features and their fixed classifier prototypes), and discriminability is assessed by the base–novel feature similarity. Implementation details are provided in Appendix~\ref{supp:details_class-sensitive}.

As shown in Fig.~\ref{fig:selective_scan}~(a), $\mathcal{L}_{\text{supp}}$ substantially strengthens the alignment between base features and their prototypes across sessions. By reducing the incremental-branch response on base inputs, this loss encourages base representations to remain close to the frozen stability anchor, thereby mitigating feature drift caused by novel-class updates.

Fig.~\ref{fig:selective_scan}~(b) demonstrates that $\mathcal{L}_{\text{sep}}$ significantly reduces the representational overlap between base and novel classes. By reducing the absolute cosine similarity between generated parameters, $\mathcal{L}_{\text{sep}}$ encourages the incremental branch to produce more separated transformation patterns. This encourages more separated transformation patterns for novel concepts while reducing interference with base-class representations.

The accuracy trend in Fig.~\ref{fig:selective_scan}~(c) demonstrates the synergy between the objectives. The full model consistently outperforms its variants, supporting the role of the two objectives in improving the stability--plasticity balance.
For a detailed analysis of how Mamba-FSCIL achieves this stability--plasticity balance, we refer to Appendix~\ref{supp:granular_analysis}, where we provide in-depth visualizations and analyses of activation magnitudes, state-space parameters, and feature distributions.

\subsubsection{Impact of Freezing vs. Finetuning Base Projections}
\label{sec:exps_ablation_freeze}
\begin{figure}
    \centering
    \includegraphics[width=1\linewidth]{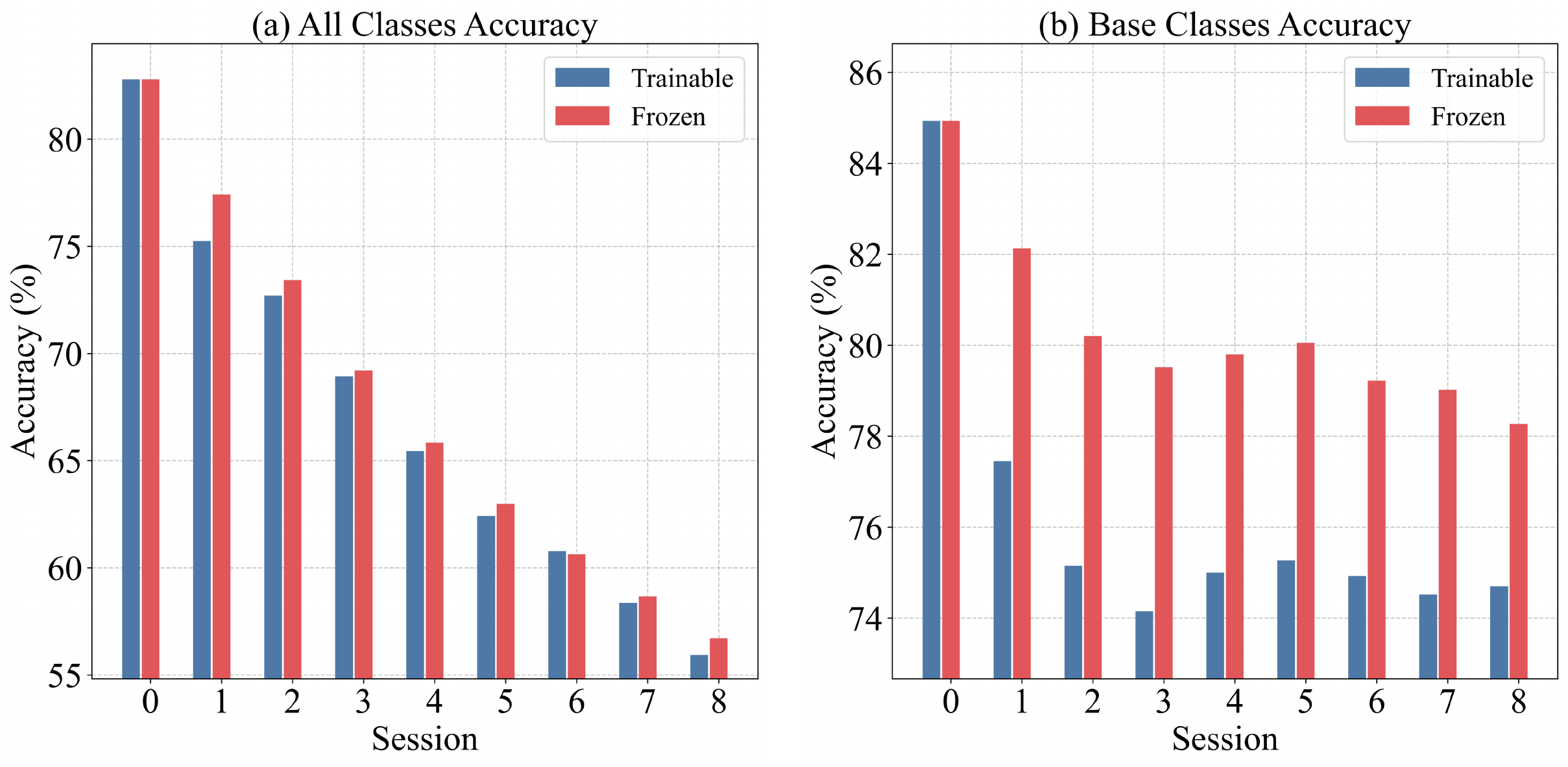}
    \vspace{-6mm}
    \caption{Impact of freezing vs. training the identity branch $p_\text{iden}$ and the base selective SSM branch $g_\text{base}$ during incremental sessions on miniImageNet: (a) Accuracy for all learned classes, (b) Accuracy for base classes.}
    \label{fig:trainable_frozen}
    \vspace{-4mm}
\end{figure}
We assess the impact of freezing the identity ($p_{\text{iden}}$) and base selective SSM ($g_{\text{base}}$) branches after base session training versus keeping them trainable throughout incremental sessions, using a ResNet-12 backbone on miniImageNet.
As shown in Fig.~\ref{fig:trainable_frozen}, the frozen configuration consistently achieves higher accuracy across all classes. The results support our design choice: freezing these branches effectively preserves base-session knowledge, crucial for mitigating forgetting. Allowing these branches to be trainable degrades performance by inadvertently adapting them to novel classes and disrupting the established knowledge.

\subsubsection{Impact of Number of Scan Paths}
\label{sec:exps_ablation_scanpaths}
\begin{table}[t!]
	\begin{center}
        \caption{Impact of the number of 2D Selective Scan (SS2D) paths (K) on CIFAR-100 and CUB-200.}
        \vspace{-2mm}
        \resizebox{\linewidth}{!}{
		\begin{tabular}{l|lcc|lcc}
			\toprule
			\multicolumn{1}{l|}{\multirow{2}{*}{\textbf{K}}} & \multicolumn{3}{c|}{\textbf{CIFAR-100}}& \multicolumn{3}{c}{\textbf{CUB-200}}\\ 
			& \textbf{\textsc{BASE}} &\textbf{\textsc{Final}} & \textbf{\textsc{AVG}}& \textbf{\textsc{BASE}} &\textbf{\textsc{Final}} & \textbf{\textsc{AVG}}\\ \midrule
            1&82.35&55.94&67.23&80.41&60.84&67.48\\
            2&82.65&57.13&67.53&80.49&61.08&67.86\\
            4&\textbf{82.80}&\textbf{57.51}&\textbf{68.14}&\textbf{80.90}&\textbf{61.65}&\textbf{68.13}\\ \bottomrule
		\end{tabular}}
\label{tab:influ_scan}
 \vspace{-2mm}
\end{center}
\end{table}

We evaluate the influence of the number of scan paths ($K$, Eq.~\ref{equ:ssm_branch_parameter_generation}) within our Selective SSM branches on CIFAR-100 and CUB-200 (Tab.~\ref{tab:influ_scan}). Increasing $K$ consistently enhances base, final, and average session accuracies. For instance, on CIFAR-100, extending from $K=1$ to $K=4$ boosts final session accuracy by +1.57\% (from 55.94\% to 57.51\%) and average accuracy by +0.91\% (from 67.23\% to 68.14\%), with similar gains on CUB-200. This highlights how more scan paths improve contextual information integration, crucial for complex visual tasks and adaptation. We employ four scan paths ($K=4$) by default.

\subsubsection{Comparison of Suppression Loss Variants}
\label{sec:exps_ablation_suppression}
\begin{table}[t!]
\centering
\caption{Comparison of different suppression losses on CIFAR-100 and CUB-200. Results are averaged over five random seeds.}
\vspace{-2mm}
    \label{tab:supp_comparison}
    \resizebox{\linewidth}{!}{
    \begin{tabular}{l|cc|cc}
    \toprule
    \multicolumn{1}{l|}{\multirow{2}{*}{\textbf{Loss Type}}}  & \multicolumn{2}{c}{\textbf{CIFAR-100}} & \multicolumn{2}{c}{\textbf{CUB-200}} \\
    & \textsc{\textbf{AVG}} & \textsc{\textbf{PD}} & \textsc{\textbf{AVG}} & \textsc{\textbf{PD}} \\
    \midrule
    $\|\mathbf{Z}_{\text{inc}}\|^2$ (\textbf{Ours}) & \textbf{68.08$\pm$0.007} & \textbf{26.08$\pm$0.035} & \textbf{68.05$\pm$0.001} & \textbf{20.01$\pm$0.062} \\
    $\|\mathrm{SiLU}(\mathbf{Z}_{\text{inc}})\|^2$ & 67.90$\pm$0.001 & 26.87$\pm$0.025 & 67.21$\pm$0.819 & 20.97$\pm$1.891 \\
    $\|\boldsymbol{\mu}_{\text{inc}}\|^2$& 67.80$\pm$0.001& 27.00$\pm$0.055 & 67.88$\pm$0.028 & 20.01$\pm$0.189 \\
    \bottomrule
    \end{tabular}
    }
\end{table}
We compare three suppression loss designs for controlling the gating signal $\mathbf{Z}_{\text{inc}}$ on CIFAR-100 and CUB-200. For all experiments, we set $\lambda_1 \in \{50,100\}$ and $\lambda_2=\lambda_3=0$, and report results averaged over five random seeds per dataset.
As shown in Tab.~\ref{tab:supp_comparison}, directly constraining $\mathbf{Z}_{\text{inc}}$ achieves the best \textsc{AVG} on both datasets and the lowest or tied-lowest \textsc{PD}. In contrast, $\|\mathrm{SiLU}(\mathbf{Z}_{\text{inc}})\|^2$ shows larger variance, while $\|\boldsymbol{\mu}_{\text{inc}}\|^2$ suppresses the full branch output after directional scanning and gating in Eq.~\ref{equ:ssm_branch_gated_output}, which can reduce novel-class adaptability. These results suggest that directly regulating the gate magnitude provides a more targeted way to control the incremental pathway while preserving the selective-scan outputs $\{\mathbf{Y}_k\}_{k=1}^{K}$.

\subsubsection{Hyperparameter Sensitivity Analysis}
\label{sec:exps_ablation_hyperparameter}
\begin{table}[t!]
    \centering
    \caption{Sensitivity analysis of hyperparameters on miniImageNet.}
    \vspace{-2mm}
    \begin{tabular}{c|ccccc}
        \toprule
        \textbf{$\lambda_1$} & 40 & 80 & 120 & 140 & 160 \\
        \textbf{\textsc{AVG}}  & 69.72 & \textbf{69.81} & \textbf{69.81} & 69.80 & 69.79 \\
        \midrule
        \textbf{$\lambda_2$} & 0.001 & 0.01 & 0.1 & 1 & 10 \\
        \textbf{\textsc{AVG}} & 69.75 & 69.67 & \textbf{69.81} & 69.76 & 69.23 \\
        \midrule
        \textbf{$\lambda_3$} & 0.05 & 0.1 & 0.3 & 0.5 & 0.7 \\
        \textbf{\textsc{AVG}} & 69.64 & 69.70 & 69.71 & \textbf{69.81} & 69.79 \\
        \bottomrule
    \end{tabular}
    \vspace{-4mm}
    \label{tab:influ_hyper}

\end{table}

We assess the impact of hyperparameters $\lambda_1$, $\lambda_2$, and $\lambda_3$ in Eq.~\ref{equ:optim_inc} on average accuracy using miniImageNet. As shown in Tab.~\ref{tab:influ_hyper}, varying each parameter in a proper range individually while holding others fixed yields only minor accuracy fluctuations, indicating that our model is robust to these hyperparameters. A joint hyperparameter study over a larger range is provided in Appendix~\ref{sec:appendix_hyperparameters}.

\subsubsection{Compatibility with Different Classifier Architectures}
\label{sec:exps_ablation_classifiers}

\begin{table}[t!]
\centering
\caption{Average accuracy across all sessions under different projector–classifier combinations.}
\vspace{-2mm}
\resizebox{\linewidth}{!}{
\begin{tabular}{llccc}
\toprule
 \textbf{Projector} &\textbf{Classifier}& \textbf{miniImageNet} & \textbf{CIFAR} & \textbf{CUB} \\
\midrule
 MLP &Linear (CE)& 61.30 & 62.68 & 59.58 \\
 Dual-SSM &Linear (CE)& \textbf{64.20} & \textbf{67.14} & \textbf{63.04} \\
\midrule
 MLP &ETF (DR)& 67.82 & 67.50 & 67.28 \\
 Dual-SSM &ETF (DR)& \textbf{69.81} & \textbf{68.14} & \textbf{68.13} \\
\bottomrule
\end{tabular}}
\label{tab:etf_ablation}
\end{table}

We assess the compatibility of our method with different classifier designs. While Mamba-FSCIL primarily utilizes a fixed ETF classifier trained with DR loss (Eq.~\ref{equ:loss_dr}), we also evaluate a standard learnable linear classifier with cross-entropy (CE) loss. Each classifier is paired with either a baseline MLP projector or our dual selective SSM projector.

As shown in Tab.~\ref{tab:etf_ablation}, our projection module consistently improves performance across all datasets, regardless of the classifier type. Notably, when paired with the standard classifier, our method yields notable gains (up to +4.46\% on CIFAR-100), suggesting its effectiveness even without geometric regularization from ETF. When combined with ETF, our approach further benefits from its ideal feature space arrangement, yielding the highest overall performance (e.g., 69.81\% on miniImageNet). These results suggest that the proposed projector is compatible with different classifier choices, rather than relying exclusively on the fixed ETF classifier.

\begin{figure*}[t!]
    \centering
    \includegraphics[width=1\linewidth]{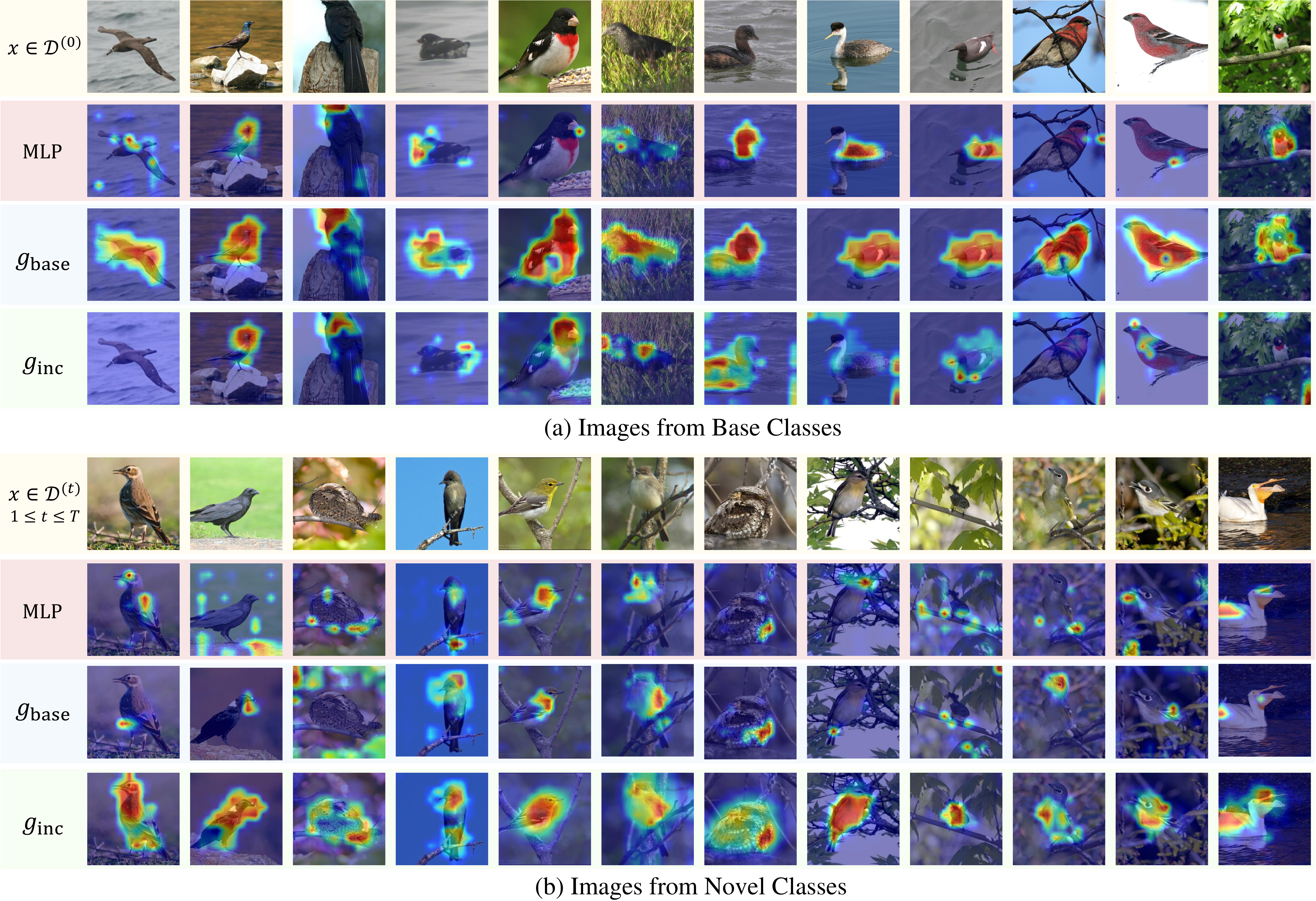}
    \vspace{-6mm}
     \caption{Activation map visualizations on CUB-200 after 10 incremental sessions. For the MLP baseline, we visualize activation maps derived from the backbone features (i.e., the input to the MLP projector). 
    For Mamba-FSCIL, we visualize activation maps computed from the input-dependent parameter matrix $\mathbf{B}$ inside each selective SSM branch ($g_{\text{base}}$ and $g_{\text{inc}}$). 
    These maps reflect how different projectors modulate input regions under base-class (a) and novel-class (b) images.}
    \label{fig:vis_cam}
\end{figure*}

\subsubsection{Comparison with Different Architectures}
\label{sec:exps_ablation_diff_projection}
\begin{table}[t!]
    \caption{Performance and computational complexity of different projection architectures on CIFAR-100. \textbf{Comp.} denotes the theoretical time complexity for processing a sequence of length $n$.}
\vspace{-2mm}
\centering
\resizebox{\linewidth}{!}{
\begin{tabular}{lcccc}
\toprule
    \textbf{Architecture} &\textbf{Comp.} &\textbf{\textsc{BASE}} &\textbf{\textsc{Final}} &\textbf{\textsc{AVG}} \\
    \midrule
    GRU~\cite{cho2014learning} &$\mathcal{O}(n)$ & 82.75 & 48.75 & 61.57 \\
    LSTM~\cite{hochreiter1997long} & $\mathcal{O}(n)$ & 83.13 & 49.80 & 63.58 \\
    S4~\cite{gu2021efficiently} & $\mathcal{O}(n \log n)$ & 82.25 & 54.43 & 65.24 \\
    MLP & $\mathcal{O}(n)$ & 82.73 & 54.23 & 66.35 \\
    Transformer~\cite{dosovitskiy2020image} &$\mathcal{O}(n^2)$ &82.45 & 55.05 & 66.46 \\
    \midrule
    \textbf{Selective SSM} & \textbf{$\mathcal{O}(n)$} & \textbf{82.80} & \textbf{57.24} & \textbf{68.08} \\
 \bottomrule
 \end{tabular}
}
 \vspace{-4mm}
\label{tab:diff_archs}
\end{table}

To evaluate the effect of the projection architecture, we compare Selective SSM with alternative projectors, including MLP, Transformer, S4, GRU, and LSTM, on CIFAR-100. For a fair comparison, all variants use the same ResNet-12 backbone, dual-branch design, and comparable parameter budgets (15M--17M). We also align their base-session accuracy to control for different initial performance levels, so that the comparison focuses on incremental adaptation and forgetting mitigation. Additional implementation details are provided in Appendix~\ref{supp:details_diff_projections}.

As shown in Tab.~\ref{tab:diff_archs}, the compared projectors have similar base-session accuracies, while their incremental performance differs notably. Static projectors (MLP/Transformer) and input-invariant recurrent models (S4/GRU/LSTM) obtain larger performance drops, with final-session accuracies ranging from 48.75\% to 55.05\%. In contrast, Selective SSM achieves the highest final accuracy (57.24\%) and average accuracy (68.08\%). These results suggest that input-conditioned state-space operators provide a suitable projector design for balancing novel-class adaptation and forgetting mitigation under the same FSCIL setting. A more detailed analysis of the parameter-sharing conflict and architectural differences is provided in Appendix~\ref{supp:theoretical_analysis}.

\vspace{-2mm}
\subsection{Qualitative Analysis}
\label{sec:exps_qualitative}

\subsubsection{Class-Sensitive Dynamic Adaptation Analysis}
\label{sec:exps_qualitative_gradcam}
To qualitatively examine how the dual-branch projector preserves base-class knowledge and adapts to novel classes, we visualize activation maps on CUB-200 after the final incremental session. All experiments use the same ImageNet-1K pretrained Swin-T backbone as in Sec.~\ref{sec:exps_overall_transformer}.  
For the static MLP projector, we visualize activation maps computed from the backbone output (the input features to the MLP head).  
Since this baseline uses a frozen backbone and a session-shared MLP projection, its activation patterns provide a reference for how the shared projected representation responds after incremental training. For Mamba-FSCIL, we visualize activation maps derived from the input-conditioned $\mathbf{B}$ matrices generated inside $g_{\text{base}}$ and $g_{\text{inc}}$. Since $\mathbf{B}$ influences how input tokens modulate the hidden-state evolution, these maps provide a qualitative view of how each branch emphasizes different image regions during selective scanning.

Fig.~\ref{fig:vis_cam} presents the activation patterns for both base and novel classes. The static MLP baseline, which relies on a shared session-agnostic projection, exhibits noticeable feature drift after incremental updates: base-class maps become less focused, while novel-class maps appear diffuse or misaligned. This pattern indicates limited ability to retain old knowledge and insufficient adaptability to new classes.

In contrast, Mamba-FSCIL shows clearly differentiated activation behaviors across its two branches. \textbf{For base-class images}, the frozen $g_{\text{base}}$ branch shows strong and coherent activations over the main object regions, demonstrating that its retained transformation pathway remains stable throughout incremental learning. Meanwhile, the $g_{\text{inc}}$ branch exhibits minimal activations for base inputs, which aligns with the effect of the suppression loss $\mathcal{L}_{\text{supp,base}}$ that reduces the incremental branch's contribution for known classes.
\textbf{For novel-class images}, the frozen $g_{\text{base}}$ activations become much weaker, as expected for a branch optimized only during the base session. In contrast, the $g_{\text{inc}}$ branch produces strong, well-localized activations driven by the input-conditioned parameters.  Its selective modulation is enhanced by the suppression loss $\mathcal{L}_{\text{supp,novel}}$ and the separation loss $\mathcal{L}_{\text{sep}}$, which encourage distinct activation patterns between base and novel classes, enabling adaptive and class-sensitive modulation for new categories while preserving stability for previously learned classes.
These visualizations qualitatively support that the dual-branch selective SSM projector achieves both stable preservation of base-class focus and dynamic adaptation to novel classes through input-conditioned operator generation.

\begin{figure}[t!]
\vspace{-2mm}
    \centering
    \includegraphics[width=1\linewidth]{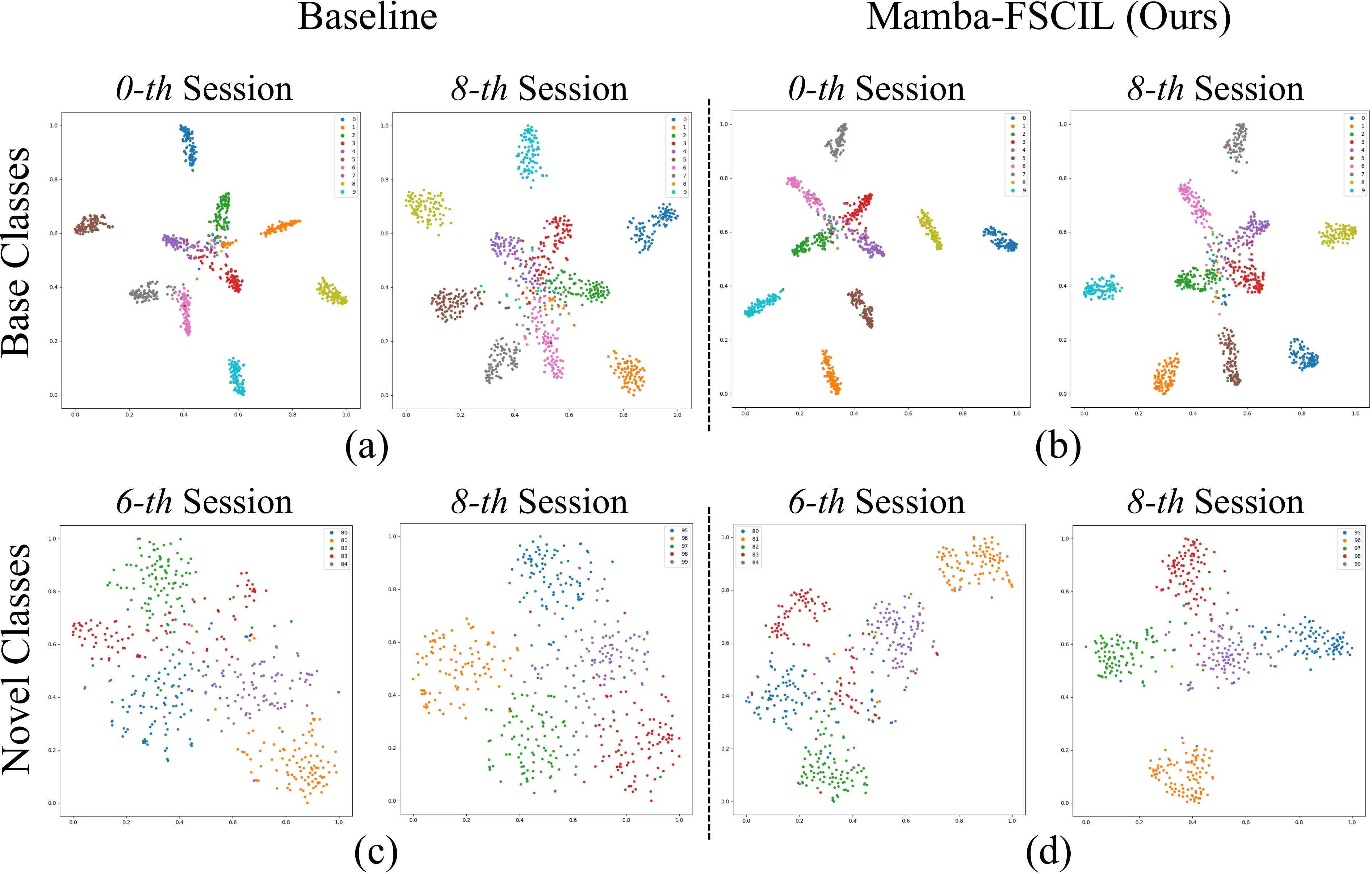}
    \vspace{-6mm}
    \caption{t-SNE visualization of feature embeddings on CIFAR-100. \textbf{(a, c):} Baseline; \textbf{(b, d):} Mamba-FSCIL. \textbf{Top:} Base classes at 0-th and 8-th sessions. \textbf{Bottom:} Novel classes at 6-th and 8-th sessions. Colors indicate class labels.}
    \label{fig:vis_features}
    \vspace{-4mm}
\end{figure}
\subsubsection{Feature Embedding Visualization Analysis}
\label{sec:exps_qualitative_tsne}

We analyze feature embeddings using t-SNE visualizations for both base and novel classes from the test set of the CIFAR-100 dataset, comparing the strong baseline NC-FSCIL with our Mamba-FSCIL. Fig.~\ref{fig:vis_features}~(a) and Fig.~\ref{fig:vis_features}~(b) show the results of some base classes during the first and last sessions. The visualization shows that the base-class features of the baseline method are much more scattered in the last session. As a comparison, Mamba-FSCIL demonstrates consistent intra-class compactness and distinct inter-class separation, indicating enhanced stability for base classes that better maintain their performance in incremental training. 
Fig.~\ref{fig:vis_features} (c) and Fig.~\ref{fig:vis_features} (d) depict the results of some novel classes in the \textit{6-th} and \textit{8-th} sessions. In both sessions, Mamba-FSCIL shows significantly more compact clusters with clearer margins among classes, highlighting its capacity to accommodate new classes. These results provide qualitative evidence consistent with the benchmark improvements of Mamba-FSCIL.

\vspace{-2mm}
\section{Conclusion}
\label{sec:conclusion}
In this study, we propose the Mamba-FSCIL framework, leveraging selective state space models to address the challenges of FSCIL. Mamba-FSCIL utilizes a dual selective SSM projector to achieve dynamic adaptation without the need to continually expand the parameter space of the model. The dual design dynamically adjusts its parameters to integrate new classes while preserving the integrity of previously learned information. Furthermore, the incorporation of the class-sensitive selective scan mechanism, composed of the suppression loss and the separation loss, guides the dynamic adaptation process, enhancing both stability and adaptability in incremental sessions' training. Empirical results demonstrate that Mamba-FSCIL outperforms the state-of-the-art static and dynamic methods across three benchmark datasets. 

\bibliographystyle{IEEEtran}
\bibliography{main}

\clearpage
\twocolumn
\captionsetup[figure]{labelformat=default}
\captionsetup[table]{labelformat=default}
\setcounter{page}{1}
\appendices
\clearpage
\section{Implementation Details}
\label{supp:details}
\subsection{Initialization of the Incremental Branch}
In incremental sessions, the identity branch $p_{\text{iden}}$ and the base selective SSM branch $g_\text{base}$ are frozen, while the incremental selective SSM branch $g_\text{inc}$ is optimized and reused across all incremental sessions. To avoid disrupting the base-session representation at the beginning of incremental training, we \emph{zero-initialize} $g_\text{inc}$ by setting the projection parameters of $p_z$ (which generate the gating features $\mathbf{Z}_\text{inc}$) to zero. Consequently, the initial incremental output $\boldsymbol{\mu}_\text{inc}$ is zero, and the fused representation reduces to $\boldsymbol{\mu} \approx \boldsymbol{\mu}_\text{iden} + \boldsymbol{\mu}_\text{base}$ at the start of the incremental phase. During training, $g_\text{inc}$ is gradually optimized to model feature shifts for novel classes, while the frozen branches preserve knowledge learned in base session.

\subsection{Training Details for CNN Backbones}
\label{supp:details_cnn}
On \textbf{miniImageNet}, ResNet-12 and ResNet-18 are trained for 500 epochs in the base session, followed by 100--170 iterations per incremental session. The initial learning rates are set to 0.25 (base) and 0.01 (incremental).

On \textbf{CIFAR-100}, ResNet-12 and ResNet-18 are used with 200 base training epochs and 200 iterations per incremental session. The learning rate is 0.25 for both stages.

On \textbf{CUB-200}, ResNet-18, pre-trained on ImageNet-1K~\cite{russakovsky2015imagenet} following~\cite{yang2023neural}, is trained for 80 base epochs and 200--290 incremental iterations. The learning rate is set to 0.02 for backbone and 0.2 for projector during base training, and reduced to 0.05 in incremental sessions.

\subsection{Training Details for Transformer Backbones}
\label{supp:details_vit}
We evaluate the scalability of Mamba-FSCIL on CUB-200 using three transformer backbones:

\textbf{Swin-T (IN1K).} Following Comp-FSCIL~\cite{zou2024compositional}, we use Swin Transformer-Tiny~\cite{liu2021swin} pretrained on ImageNet-1K~\cite{russakovsky2015imagenet}, with $384 \times 384$ input resolution. The model is trained for 80 epochs in the base session and 600 iterations per incremental session. Learning rate settings are the same as those used for the CNN backbone.

\textbf{ViT-B/16 (CLIP).} Following CPE-CLIP~\cite{d2023multimodal}, we adopt the CLIP's ViT-B/16 vision encoder~\cite{radford2021learning}, with $224 \times 224$ input resolution. Only the last transformer block is fine-tuned during base training. The model is trained for 200 base epochs and 1000 iterations per incremental session, with learning rates of 0.2 and 0.05, respectively.

\textbf{ViT-B/16 (IN21K).} Following PriViLege~\cite{park2024pre}, we use ViT-B/16~\cite{dosovitskiy2020image} pretrained on ImageNet-21K, utilizing the same training setup as ViT-B/16 (CLIP).

\subsection{Evaluation metrics for Class-Sensitive Selective Scan}
\label{supp:details_class-sensitive}
To support the analysis in Sec.~\ref{sec:exps_ablation_losses}, we detail the diagnostics used to quantify stability and discriminability induced by the class-sensitive selective scan mechanism:

\textbf{Feature–Classifier Alignment (Stability Proxy).}
This metric measures how well the learned feature representations remain aligned with the fixed ETF classifier prototypes across incremental sessions. For each base class $k$, let $\boldsymbol{\mu}_{k,i}$ denote the final feature representation of the $i$-th sample and $\mathbf{w}_k$ denote the corresponding classifier prototype. The metric is:
\begin{equation}
\text{Align} = 
\frac{1}{\sum_{k=1}^{C_{\text{base}}} N_k}
\sum_{k=1}^{C_{\text{base}}}
\sum_{i=1}^{N_k}
\cos(\boldsymbol{\mu}_{k,i}, \mathbf{w}_k),
\end{equation}
where $C_{\text{base}}$ is the number of base classes and $N_k$ is the number of test samples from class $k$. Higher values indicate stronger stability and less drift in the base-class manifold.

\textbf{Base–Novel Class Separability (Discriminability Proxy).}
This metric evaluates how distinctly the model represents base and novel classes. Let $\boldsymbol{\mu}_{\text{base},i}$ and $\boldsymbol{\mu}_{\text{novel},j}$ denote features from base-class and novel-class test samples, with $N_{\text{base}}$ and $N_{\text{novel}}$ samples, respectively. The metric is:
\begin{equation}
\text{Sep} =
\frac{1}{N_{\text{base}} N_{\text{novel}}}
\sum_{i=1}^{N_{\text{base}}}
\sum_{j=1}^{N_{\text{novel}}}
\cos( \boldsymbol{\mu}_{\text{base},i}, \boldsymbol{\mu}_{\text{novel},j} ).
\end{equation}

Lower values correspond to a clearer separation between base and novel classes and indicate stronger discriminability induced by the selective scan mechanism.

These two metrics jointly quantify how the proposed regularization transforms Mamba’s input-dependent dynamics into stable, class-sensitive adaptation.

\subsection{Implementation Details for Different Projections}
\label{supp:details_diff_projections}
To validate the effectiveness of the selective SSM projection branches within our Dual Selective SSM Projector for FSCIL, we compare its performance against several alternative projection architectures on CIFAR-100. All models share a ResNet-12 backbone within the same dual-branch framework and have comparable total parameter counts (15M–17M, including backbone and projector) to ensure a fair comparison. Specifically, we replace the Selective Scan Mechanism in our selective SSM branch architecture (green in Fig.~\ref{fig:framework}~(b)) with: a 3-layer MLP, an 8-head Transformer followed by a feed-forward network, a static SSM (S4) with input-independent parameters, and RNN baselines (GRU and LSTM with hidden size 640). Suppression loss ($\lambda_1 \in [50,100], \lambda_2=0$) is applied to all configurations. For architectures without an inherent gating mechanism (e.g., MLP, Transformer, GRU, LSTM), the suppression loss is applied to the output of the incremental branch ($g_\text{inc}$). For gating-based designs (Selective/Static SSM), it is applied to the gating features $\mathbf{Z}_\text{inc}$, as detailed in Eq.~\ref{equ:ssm_branch_scan_and_gate}.

\subsection{Hyperparameter Selection Protocol}
\label{supp:hyperparameter_protocol}
To avoid inadvertent tuning-to-test, we determine architectural choices and loss weights using only the base-session training data. Specifically, we split the base-session training classes into a pseudo-base session and several few-shot pseudo-incremental sessions to simulate the FSCIL process. All validation scores used for model-design decisions are computed on these pseudo-incremental tasks with seed 0. The actual novel classes used in benchmark evaluation are not used in this process.

The selection procedure follows the order of the method design. We first determine the architectural choices, and then choose the loss weights from a small set of predefined discrete candidate values under the selected architecture.

\textbf{Stage 1: Architectural design selection.}
We first determine the categorical architectural choices using the pseudo-incremental validation protocol. These choices include the freezing strategy for the base projection branch, the direct gating-feature suppression variant, and the number of SS2D scan paths. For each architectural factor, we compare the corresponding candidate designs on the pseudo-incremental validation tasks and select the one with the highest average accuracy. Once determined, these architectural choices are fixed for the subsequent loss-weight selection and for the actual benchmark evaluation within each corresponding setting.

 \textbf{Stage 2: Loss-weight selection.}
After fixing the architectural choices, we select the loss weights $\lambda_1$, $\lambda_2$, and $\lambda_3$ from predefined discrete candidate sets. This procedure is not a continuous hyperparameter optimization process; it only compares a small number of candidate values.

Specifically, we first choose $\lambda_1$ from $\{50,100,150,200\}$ with $\lambda_2=0$ and $\lambda_3=0$. After fixing $\lambda_1$, we choose $\lambda_2$ from $\{0.001,0.01,0.1,1.0\}$ with $\lambda_3=0$. Finally, with $\lambda_1$ and $\lambda_2$ fixed, we choose $\lambda_3$ from $\{0.05,0.1, 0.3, 0.5\}$. The configuration with the highest average accuracy on the pseudo-incremental validation tasks is selected. Once determined, the loss weights are fixed for all actual incremental sessions and all random seeds within each corresponding benchmark setting.

\section{Additional Experimental Analysis}
\label{supp:results}
\subsection{Performance Drop Comparison}
\label{supp:pd_comparison}
To further analyze forgetting behavior, we summarize the performance drop (PD) across the three benchmarks in Tables~\ref{table:imgnet}, \ref{table:cifar}, and \ref{table:cub}.  For the main benchmark settings where three-seed results are reported, we use the corresponding mean PD values. The results show that the improvements in average and final accuracy are achieved without sacrificing stability.

\begin{itemize}
    \item \textbf{miniImageNet:} Mamba-FSCIL achieves a favorable balance between accuracy and forgetting. In the moderate base-accuracy regime, its PD (22.61\%) is comparable to strong baselines such as FACT (22.07\%), while achieving higher average accuracy. In the high base-accuracy regime ($\sim$80--84\%), Mamba-FSCIL obtains PD values of 24.82\% and $25.25{\pm}0.29\%$, which are comparable to or lower than competitive methods such as ALICE (24.90\%) and NC-FSCIL (25.71\%), while maintaining higher final accuracy.
    \item \textbf{CIFAR-100:} Mamba-FSCIL shows stable performance across different base-accuracy regimes. In the low-to-mid base regimes ($\sim$73--75\%), it achieves lower PD values (21.08\%--22.18\%) than DSN (23.65\%) and CEC (23.93\%). In the strongest base regime ($\sim$82\%), its PD is $25.43{\pm}0.24\%$, which remains lower than NC-FSCIL (26.41\%) and YourSelf (26.30\%). This indicates that the method improves final-session performance while maintaining competitive forgetting behavior.
    \item \textbf{CUB-200:} Mamba-FSCIL also maintains strong stability on the fine-grained CUB-200 benchmark. In the mid-range base-accuracy regime ($\sim$77\%), it obtains the lowest PD among the compared ResNet-18 methods (16.83\%). In the high base-accuracy regime ($\sim$80\%), its PD is $19.40{\pm}0.14\%$, which is lower than NC-FSCIL (21.01\%) and FeSSSS (26.62\%). These results suggest that the class-sensitive selective scan mechanism helps reduce interference with previously learned classes in fine-grained incremental learning.
\end{itemize}

\subsection{Performance Trends Across Sessions}
\label{supp:trend}

\begin{figure*}[t!]
\centering
\includegraphics[width=1\linewidth]{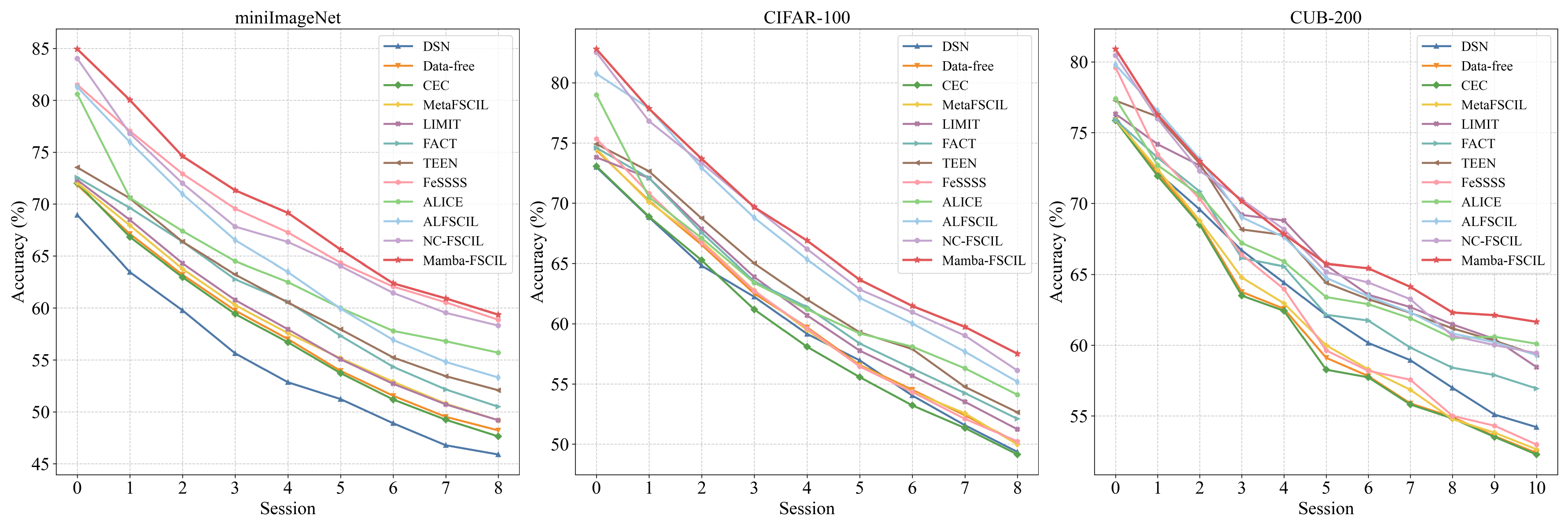}
\vspace{-6mm}
\caption{Performance trends across incremental sessions on miniImageNet, CIFAR-100, and CUB-200.}
\label{fig:trend_full}
\end{figure*}
To provide a more intuitive understanding of the long-term behavior of different FSCIL methods, we visualize the accuracy trends across incremental sessions on \textbf{miniImageNet}, \textbf{CIFAR-100}, and \textbf{CUB-200}. The performance curves are shown in Fig.~\ref{fig:trend_full}, generated directly from the numerical results in Tables~\ref{table:imgnet}--\ref{table:cub}. We compare Mamba-FSCIL with varying competitive baselines including comparisons with dynamic methods (e.g., DSN~\cite{yang2022dynamic}, FeSSSS~\cite{ahmad2022few}) and static strong baselines (e.g., NC-FSCIL~\cite{yang2023neural}, ALICE~\cite{peng2022few}). 

Across all three benchmarks, Mamba-FSCIL shows favorable long-term stability trends:
On \textbf{miniImageNet}, Mamba-FSCIL exhibits one of the slowest decay rates among all methods. Even when compared to methods with similar base accuracy (e.g., YourSelf, NC-FSCIL), our method maintains a clear advantage in the later sessions, reflecting its better resistance to forgetting.
On \textbf{CIFAR-100}, Mamba-FSCIL not only remains competitive in early sessions but surpasses most competitors in the final session, consistent with our leading \textbf{AVG} scores and highly competitive \textbf{PD} values.
On \textbf{CUB-200}, Mamba-FSCIL shows a stable trend among all baselines, and consistently ranks at or near the top in every incremental stage, further supporting its robustness.

The performance curves highlight the effectiveness of our design: The \textbf{frozen base branch} anchors stable representations throughout all sessions. The \textbf{input-dependent Mamba incremental branch} adapts to novel classes while reducing disturbance to the base feature space. The \textbf{class-sensitive losses} suppress harmful updates on base samples and enforce discriminative operator divergence for novel samples.

\subsection{Impact of Training Shots}
\label{sec:exps_ablation_trainingshots}
\begin{figure*}[t!]
    \centering
    \includegraphics[width=1\linewidth]{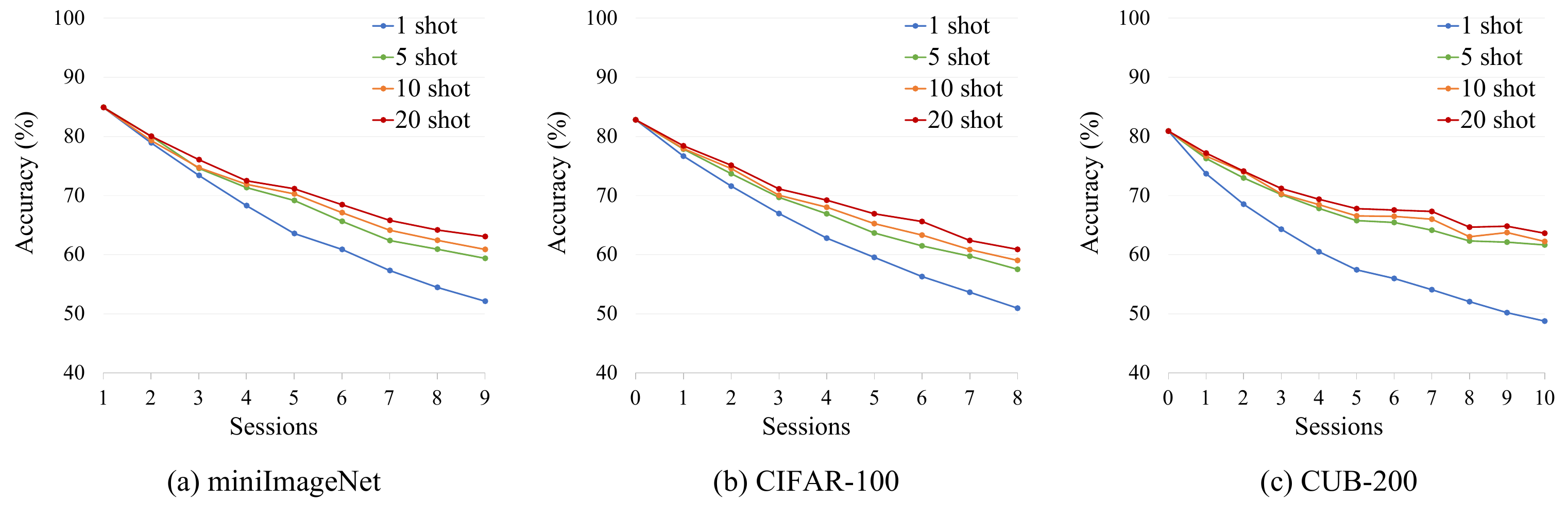}
    \vspace{-6mm}
    \caption{Performance across different sessions for 1-shot, 5-shot, 10-shot, and 20-shot learning scenarios on miniImageNet, CIFAR-100, and CUB-200 datasets.}
    \label{fig:multishot}
\end{figure*}

We explore the effect of varying the number of training samples per class on Mamba-FSCIL's performance. As shown in Fig.~\ref{fig:multishot}, increasing the number of training examples from 1-shot to 20-shot consistently improves accuracy across all sessions and datasets. This demonstrates Mamba-FSCIL's ability to leverage additional novel class samples while preserving base knowledge and mitigating forgetting.

\subsection{Joint Hyperparameter Ablation}
\label{sec:appendix_hyperparameters}
\begin{figure*}[t!]
    \centering
    \includegraphics[width=1\textwidth]{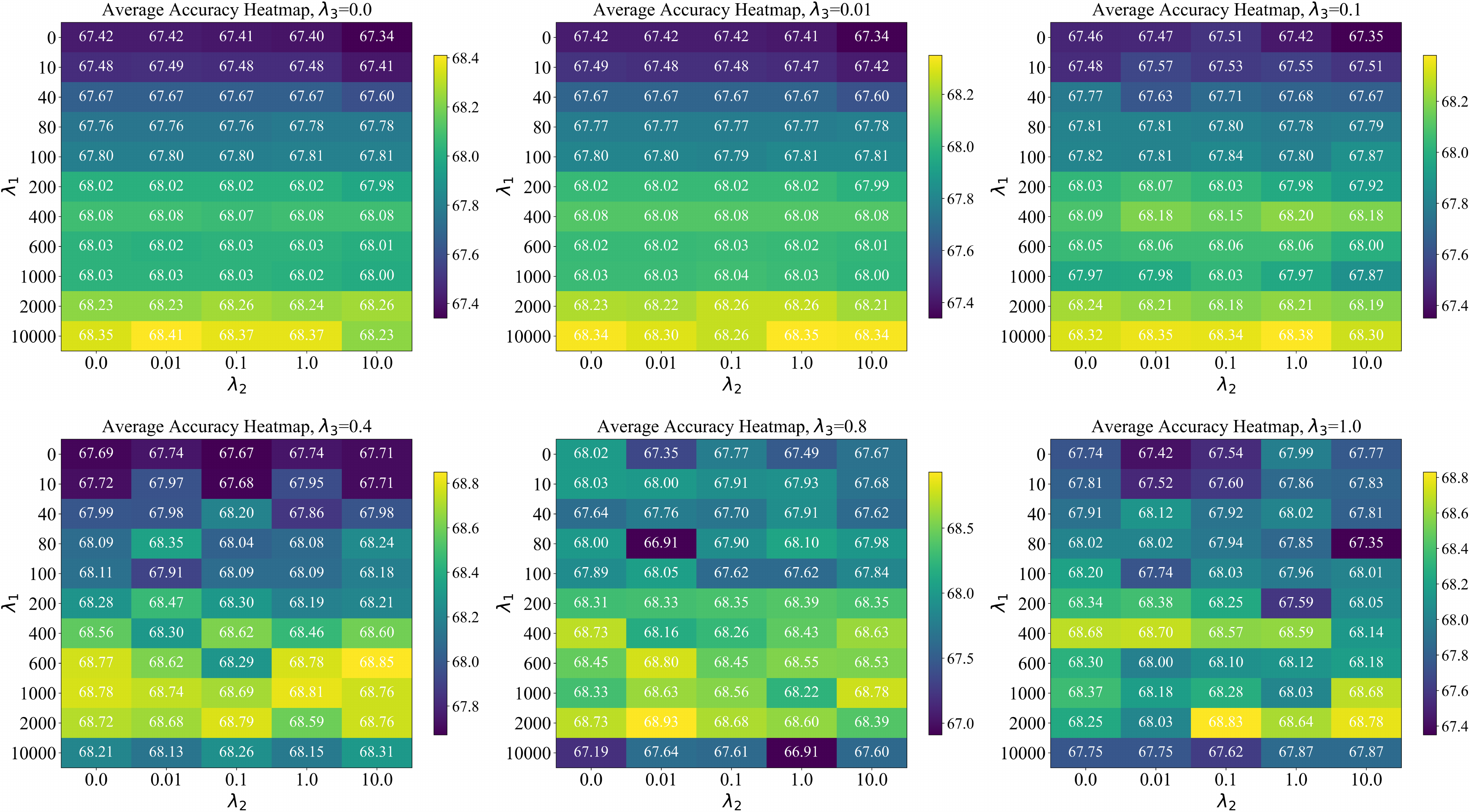}
    \vspace{-6mm}
    \caption{\textbf{Joint Hyperparameter Ablation on miniImageNet.} The heatmaps display the Average Accuracy (\%) across different combinations of $\lambda_1$ (Y-axis), $\lambda_2$ (X-axis), and $\lambda_3$ (Subplots). Warmer colors (yellow) indicate better performance.}
    \label{fig:rebuttal_heatmap}
    \vspace{-4mm}
\end{figure*}

To assess the robustness of Mamba-FSCIL and understand how the three regularization terms interact, we jointly vary the loss weights $\lambda_1$ (base-class suppression $\mathcal{L}_\text{supp, base}$ in Eq.~\ref{equ:loss_supp}), $\lambda_2$ (novel-class activation $\mathcal{L}_\text{supp, novel}$ in Eq.~\ref{equ:loss_supp}), and $\lambda_3$ (parameter separation $\mathcal{L}_\text{sep}$ in Eq.~\ref{equ:loss_sep}) as defined in Eq.~\ref{equ:optim_inc}.
To accelerate evaluation, we reduce the dimensionality of the generated state-space parameters to $D_B = D_C = D_\Delta = 64$ and train each incremental session for 100 iterations with batch size 32. The base session strictly follows the same settings as Tab.~\ref{table:imgnet} to ensure consistency.
We sweep $\lambda_1\in[0,10{,}000]$, $\lambda_2\in\{0,0.01,0.1,1,10\}$, and repeat the grid for $\lambda_3\in\{0,0.01,0.1,0.4,0.8,1.0\}$.
This joint hyperparameter ablation is used only for robustness analysis and is not used to select the final hyperparameters. The final selection protocol is described in Appendix~\ref{supp:hyperparameter_protocol}.
The resulting average (\textbf{\textsc{AVG}}) accuracies on miniImageNet are shown in Fig.~\ref{fig:rebuttal_heatmap}.

\begin{table*}[t!]
\renewcommand\arraystretch{1.1}
\begin{center}
\centering
\caption{Effect of feature dimension on CIFAR-100 and CUB-200 across sessions.}
\vspace{-2mm}
\begin{tabular}{l c c c c c c c c c c c c c c}
\toprule
\multicolumn{1}{l}{\multirow{2}{*}{\bf Dataset}} &
\multicolumn{1}{c}{\multirow{2}{*}{\bf Dimension}} &
\multicolumn{11}{c}{\bf Accuracy in each session (\%) } &
\multirow{2}{*}{\bf \textsc{AVG}} &
\multirow{2}{*}{\bf \textsc{PD}} \\
\cmidrule{3-13}
& & 
\bf 0 & \bf 1 & \bf 2 & \bf 3 & \bf 4 & \bf 5 & \bf 6 & \bf 7 & \bf 8 & \bf 9 & \bf 10 & & \\
\midrule
\multirow{4}{*}{\bf CIFAR-100}
& 32  & 82.73& 77.62& 73.57& 69.23& 66.25& 63.39& 61.43& 59.53& 56.86& --- & --- & 67.85& 25.87\\
& 64  & 83.10 & 78.02 & 73.64 & 69.23 & 66.22 & 63.14 & 60.79 & 59.17 & 56.96 & --- & --- & 67.81 & 26.14 \\
& 128 & 83.00& 78.18& 73.79& 70.13& 66.74& 63.88& 61.89& 59.44& 57.59& --- & --- & 68.29& 25.41\\
& 256 & 82.80 & 77.85 & 73.69 & 69.67 & 66.89 & 63.66 & 61.48 & 59.74 & 57.51 & --- & --- & 68.14 & 25.29 \\
\midrule
\multirow{4}{*}{\bf CUB-200}
& 32  & 80.27 & 75.76 & 73.12 & 69.85 & 68.00 & 65.53 & 64.85 & 64.08 & 62.24 & 61.59 & 61.05 & 67.85 & 19.22 \\
& 64  & 80.38 & 76.11 & 72.54 & 69.58 & 67.85 & 65.09 & 64.54 & 64.16 & 61.76 & 61.50 & 61.03 & 67.69 & 19.35 \\
& 128 & 80.97& 76.23& 72.94& 70.09& 67.83& 65.67& 65.26& 64.14& 62.22& 61.99& 61.60& 68.09& 19.37\\
& 256 & 80.90 & 76.26 & 72.97 & 70.14 & 67.83 & 65.74 & 65.43 & 64.12 & 62.31 & 62.12 & 61.65 & 68.13 & 19.25 \\
\bottomrule
\end{tabular}
\label{tab:ablation_dimension}
\end{center}
\vspace{-4mm}
\end{table*}

The joint hyperparameter heatmaps reveal three clear trends that align well with the intended roles of our Class-Sensitive Selective Scan mechanism (Sec.~\ref{sec:method_selective_scan_loss_suppression}). 
\textbf{First}, increasing the base-class suppression weight $\lambda_{1}$ consistently improves performance across all settings. A broad high-accuracy region emerges once $\lambda_{1} \ge 200$, and accuracy remains stable even for large values. This confirms that strongly suppressing $g_{\text{inc}}$ on base inputs is essential for preserving the base-class manifold.

\textbf{Second}, the novel-class activation weight $\lambda_{2}$ exhibits an optimal middle range. Moderate values ($0.01$--$1.0$) yield the highest accuracies, while excessively large values (e.g., $\lambda_{2}=10$) introduce slight degradation due to over-activation. Since $g_{\text{inc}}$ is the only learnable branch in incremental sessions, $\lambda_{2}$ provides a complementary signal to the classification loss, ensuring sufficiently strong feature responses for novel classes under few-shot conditions.

\textbf{Third}, the separation weight $\lambda_{3}$ improves performance when applied with moderate strength. Without separation ($\lambda_{3}=0$), accuracy plateaus around $68.4\%$. Introducing $\lambda_{3}=0.4$--$0.8$ consistently raises the peak accuracy to $68.8\%\sim 68.9\%$, indicating that encouraging divergence between base and novel operator parameters enhances novel-class discrimination. Performance slightly declines at $\lambda_3 = 1.0$ but remains strong, suggesting that overly strong separation may distort the representation geometry.

Overall, these findings show that Mamba-FSCIL remains robust across a wide hyperparameter space and that the three regularization terms interact cohesively to maintain the stability--plasticity balance central to our Class-Sensitive Selective Scan mechanism.

\subsection{Impact of Generated Parameter Dimensions}
\label{sec:ablation_dimension}

In the selective SSM branch, the input-dependent parameters $\mathbf{B}, \mathbf{C}, \mathbf{\Delta}$ are generated via linear projections (Eq.~\ref{equ:ssm_parameters}). The dimension of these parameters ($D_B=D_C=D_\Delta$, denoted as $D_\text{gen}$) controls the expressive capacity of Mamba’s dynamic modulation. To evaluate how this capacity influences FSCIL performance, we conduct experiments on CIFAR-100 and CUB-200 using the ResNet-12 backbone, varying $D_\text{gen}$ from 32 to 256. All other settings remain consistent with the main experiments.

As shown in Table~\ref{tab:ablation_dimension}, Mamba-FSCIL is relatively robust to the generated-parameter dimension. Reducing $D_\text{gen}$ from 256 (the default setting in the main paper) to 32 causes only a small decrease in average accuracy. For instance, on \textbf{CUB-200}, the average accuracy decreases by 0.28 points (from $68.13\%$ to $67.85\%$), and on \textbf{CIFAR-100}, it decreases by 0.29 points (from $68.14\%$ to $67.85\%$). Even with the lightweight configuration ($D_\text{gen}=32$), Mamba-FSCIL remains competitive with strong FSCIL baselines. On \textbf{CIFAR-100}, it achieves an average accuracy of \textbf{67.85\%}, exceeding \textbf{NC-FSCIL} (67.50\%) and \textbf{ALFSCIL} (66.75\%). On \textbf{CUB-200}, it also achieves \textbf{67.85\%}, exceeding \textbf{NC-FSCIL} (67.28\%) and \textbf{ALICE} (65.75\%). The best average accuracy on CIFAR-100 is obtained at $D_\text{gen}=128$, while $D_\text{gen}=256$ gives the best result on CUB-200. Therefore, we use $D_{\text{gen}}=256$ as the default setting in the main paper to maintain a strong and stable configuration across datasets.

\subsection{Impact of Backbone Adaptation Strategies}
\label{supp:results_backbone_ablation}
\begin{table*}[t!]
\renewcommand\arraystretch{1.1}
\begin{center}
\centering
\caption{Effect of backbone adaptation strategies on CIFAR-100 and CUB-200 using ViT-B/16 (CLIP).}
\vspace{-2mm}
\resizebox{1\textwidth}{!}{
\begin{tabular}{l c c c c c c c c c c c c c c}
\toprule
\multicolumn{1}{l}{\multirow{2}{*}{\bf Dataset}} &
\multicolumn{1}{c}{\multirow{2}{*}{\bf Methods}} &
\multicolumn{11}{c}{\bf Accuracy in each session (\%) } &
\multirow{2}{*}{\bf \textsc{AVG}} &
\multirow{2}{*}{\bf \textsc{PD}} \\
\cmidrule{3-13}
& & 
\bf 0 & \bf 1 & \bf 2 & \bf 3 & \bf 4 & \bf 5 & \bf 6 & \bf 7 & \bf 8 & \bf 9 & \bf 10 & & \\
\midrule
\multirow{3}{*}{\bf CIFAR-100}& Finetune Backbone& 86.77 & 82.25 & 78.39 & 73.40 & 71.68 & 68.24 & 67.77 & 65.96 & 63.81 & --& --& 73.14 & 22.96 \\
& \textbf{Frozen Backbone (Ours)}& 86.77 & 82.58 & 79.89 & 75.36 & 73.26 & 70.46 & 70.17 & 68.16 & 66.20 & --& --& 74.76 & 20.57 \\
& \textbf{Add LoRA Adapter (Ours + LoRA)}& 86.77 & 82.60 & 79.79 & 75.64 & 74.09 & 70.82 & 70.90 & 68.86 & 67.44 & --& --& 75.21 & 19.33 \\
\midrule
\multirow{3}{*}{\bf CUB-200}& Finetune Backbone& 87.05 & 78.91 & 80.41 & 76.77 & 68.35 & 71.24 & 70.35 & 70.23 & 68.06 & 66.02 & 67.73 & 73.19 & 19.32 
\\
& \textbf{Frozen Backbone (Ours)}& 87.05 & 83.01 & 81.34 & 77.90 & 74.83 & 75.57 & 75.77 & 76.32 & 74.70 & 74.94 & 74.11 & 77.77 & 12.94 
\\
& \textbf{Add LoRA Adapter (Ours + LoRA)}& 87.05 & 82.63 & 81.20 & 78.62 & 75.07 & 75.91 & 76.72 & 77.42 & 75.93 & 76.17 & 75.09 & 78.35 & 11.96 \\ \bottomrule
\end{tabular}
}
\label{tab:backbone_ablation}
\end{center}
\end{table*}

To validate our design choice of freezing the backbone and assess the extensibility of Mamba-FSCIL, we compare three adaptation strategies on CIFAR-100 and CUB-200 using the \textbf{ViT-B/16 (CLIP)} backbone. All configurations utilize the same Dual Selective SSM Projector and Class-Sensitive Selective Scan Mechanism; they differ only in how the backbone is updated during incremental sessions. We compare:
\begin{itemize}
    \item \textbf{Frozen Backbone (Ours):} The backbone is fixed; only the projector is trained.
    \item \textbf{Finetune Backbone:} The last two Transformer blocks are unfrozen and fine-tuned.
    \item \textbf{Add LoRA Adapter (Ours + LoRA):} The backbone remains frozen, but $6$ Low-Rank Adapters (LoRA)~\cite{hu2022lora} are inserted into the last two blocks (rank $r=256$ for CIFAR-100, $r=128$ for CUB-200).
\end{itemize}

Incremental sessions are trained for 1000 iterations with a batch size of 32. As shown in Tab.~\ref{tab:backbone_ablation}, the results support two key conclusions:

\vspace{1mm}\textbf{(1) Frozen backbone performs better than finetuning on scarce FSCIL data.}
Directly fine-tuning the backbone on scarce incremental data degrades performance. On CUB-200, fine-tuning increases the Performance Drop (\textsc{PD}) from 12.94\% (Frozen) to 19.32\% and lowers Average Accuracy by over 4.5\%. CIFAR-100 shows the same trend (22.96\% vs.\ 20.57\% \textsc{PD}). These results indicate that our input-dependent Mamba-based projector combined with the class-sensitive losses can dynamically adapt to novel classes without modifying the backbone. This confirms that freezing the backbone remains a principled and effective choice in FSCIL for preserving pretrained representations, consistent with standard FSCIL practice~\cite{zhang2021few,yang2023neural,zou2022margin,zou2024compositional}.

\vspace{1mm} \textbf{(2) LoRA can further complement the frozen-backbone setting.}
The results also show that Mamba-FSCIL is compatible with parameter-efficient backbone adaptation. Adding LoRA adapters improves the frozen-backbone variant, increasing the CUB-200 \textsc{AVG} from 77.77\% to 78.35\% and reducing \textsc{PD} from 12.94\% to 11.96\%. This suggests that the Dual Selective SSM Projector and Class-Sensitive Scan can be combined with lightweight backbone adaptation when additional adaptation capacity is desired.

\subsection{Memory Cost and Design Analysis}
\label{supp:memory_analysis}

In this section, we clarify the memory settings of Mamba-FSCIL, quantify its storage footprint, and evaluate its performance under matched and expanded memory budgets.

\paragraph{Memory Cost Quantification.}
Mamba-FSCIL operates under an \textit{image-exemplar-free} setting. It avoids storing viewable raw images, but retains a lightweight feature-level memory $\mathcal{M}^{(s)}$ consisting of one mean spatial feature map per old class. The total memory cost is calculated as $|\mathcal{C}_{\mathrm{old}}^{(s)}| \times D \times H \times W \times 4$ bytes and reported in MiB. 
\begin{table}[t!]
\centering
\caption{Memory cost of storing one class-mean feature per class using FP32 storage.}
\resizebox{1\linewidth}{!}{
\begin{tabular}{lccc}
\toprule
\textbf{Setting} & \textbf{Classes} & \textbf{Feature Shape} & \textbf{Memory Cost}\\
\midrule
CIFAR-100 / ResNet-12 
& 100 & $640{\times}2{\times}2$ & 0.98 MiB \\
miniImageNet / ResNet-12 
& 100 & $640{\times}5{\times}5$ & 6.10 MiB \\
CUB-200 / ResNet-18 
& 200 & $512{\times}7{\times}7$ & 19.14 MiB \\
CUB-200 / ViT/Swin 
& 200 & $768{\times}14{\times}14$ & 114.84 MiB \\
\bottomrule
\end{tabular}
}
\label{tab:response_memory_cost}
\end{table}

As shown in Tab.~\ref{tab:response_memory_cost}, the memory cost grows linearly with the number of learned classes and the spatial feature size. For CNN-based settings, the required memory is modest: 0.98 MiB on CIFAR-100, 6.10 MiB on miniImageNet, and 19.14 MiB on CUB-200 with ResNet features. For transformer-scale $14{\times}14$ features, the cost increases to 114.84 MiB because more spatial tokens are retained. This result makes the memory requirement explicit and shows that the additional storage remains manageable in our experimental settings.

\paragraph{Memory Design and Size Ablation.} 
To evaluate how performance depends on memory design and size, we conduct an ablation on CUB-200 with ResNet-18. The ablation focuses on two factors: the representativeness of the stored feature and the number of stored features per class. Specifically, we compare three settings: (1) one randomly sampled feature per class, (2) one class-mean feature per class, which is our default design, and (3) multiple sampled features per class.

\begin{table*}[t!]
\centering
\caption{Memory design and size ablation on CUB-200 with ResNet-18. The memory cost is reported relative to storing one class-mean feature per class.}
\resizebox{\linewidth}{!}{
\begin{tabular}{lcccccc}
\toprule
\textbf{Memory Design} & \textbf{Stored Feature / Class} & \textbf{Memory Mult.} & \textbf{Memory Cost} & \textbf{\textsc{Base}} & \textbf{\textsc{Final}} & \textbf{\textsc{Avg} / \textsc{PD}} \\
\midrule
One random feature
& $1\times D\times H\times W$ 
& $1\times$ & 19.14 MiB 
& 80.90 & 52.74 & 64.50 / 28.16 \\
One class-mean feature (Ours)
& $1\times D\times H\times W$ 
& $1\times$ & 19.14 MiB 
& 80.90 & 61.30 & 68.15 / 19.60 \\
Two sampled features
& $2\times D\times H\times W$ 
& $2\times$ & 38.28 MiB 
& 80.90 & 55.30 & 64.81 / 25.60 \\
Three sampled features
& $3\times D\times H\times W$ 
& $3\times$ & 57.42 MiB 
& 80.90 & 56.01 & 64.89 / 24.89 \\
Five sampled features
& $5\times D\times H\times W$ 
& $5\times$ & 95.70 MiB 
& 80.90 & 58.49 & 65.84 / 22.41 \\
\bottomrule
\end{tabular}}
\label{tab:supp_memory_design}
\end{table*}

As shown in Tab.~\ref{tab:supp_memory_design}, the class-mean design achieves the best accuracy-memory trade-off in this ablation. Under the same $1{\times}$ memory budget, replacing the class-mean feature with one randomly sampled feature reduces the average accuracy from 68.15\% to 64.50\% and the final accuracy from 61.30\% to 52.74\%. This matched-budget comparison indicates that the class-mean representation provides a more stable old-class anchor than a single sampled feature.

Expanding the budget to store multiple sampled features ($2{\times}$ to $5{\times}$) increases the storage cost from 19.14 MiB to 95.70 MiB, but does not outperform the single class-mean design in this setting. This suggests that the representativeness of the stored memory is more important than simply increasing the number of sampled features. Therefore, we adopt one class-mean feature per class as the default memory design.

\subsection{Efficiency Analysis}
\label{supp:efficiency_analysis}

\subsubsection{Training Time Evaluation}
To quantify the practical cost, we first measure the total training time over all 10 incremental sessions on CUB-200 using the ViT-B/16 (CLIP) backbone. All runs use a batch size of 32 and 2000 iterations per session on a single NVIDIA A100. For a fair comparison, the baseline MLP projector is matched in parameter size to our Dual Selective SSM Projector. The Class-Sensitive Scan mechanism introduces additional regularization terms only during training and does not add inference-time modules.

\begin{table}[t!]
\centering
\caption{Total training time over 10 incremental sessions on CUB-200 with the ViT-B/16 backbone. ``Full'' denotes the Dual Selective SSM Projector with Class-Sensitive losses.}
\begin{tabular}{l|c|c}
\toprule
\textbf{Method} & \textbf{Training Time}& \textbf{Overhead} \\
\midrule
Simple MLP Projector & 1.9 hours& $1.00\times$ \\
Dual Selective SSM Projector &  2.1 hours& $1.1\times$\\
\textbf{Mamba-FSCIL} & 2.2 hours& \textbf{$1.15\times$} \\
\bottomrule
\end{tabular}
\label{tab:efficiency_response}
\end{table}

As shown in Tab.~\ref{tab:efficiency_response}, the full Mamba-FSCIL model introduces a moderate training-time overhead in this setting. Compared with the MLP baseline, the total training time over 10 incremental sessions increases from 1.9 hours to 2.2 hours. The additional cost from the Class-Sensitive losses is limited, since these losses involve simple tensor operations and are used only during training. These results indicate that the proposed projector and training objectives add a relatively small training-time cost compared with the MLP baseline in this experimental setting.

\subsubsection{Projector-Level Profiling}
The above training-time comparison measures the total incremental training cost under one representative setting. To further analyze the practical efficiency of the projector itself, we conduct projector-level profiling under identical backbone feature settings. This analysis reports wall-clock inference latency, training latency, and peak training memory.

All profiling experiments are conducted under the same hardware and measurement protocol. We use a single NVIDIA A100-SXM4-40GB GPU and set the batch size to 8. Inference and training latency are measured using CUDA-event timing after warm-up, with explicit CUDA synchronization to avoid reporting only CPU-side kernel dispatch time. Each measurement is averaged over 500 iterations after 100 warm-up iterations, and the reported mean/std are computed over 5 repeated measurements.

The profiling is conducted at the projector level because the proposed selective SSM is placed after the visual backbone rather than inside the backbone. Thus, all compared methods share the same backbone feature extractor, and their differences only appear in the projector applied to the extracted features. We therefore feed identical backbone features to all projectors and report only the projector-side latency and peak training memory. This controlled setting isolates the computational cost introduced by different projector designs.

\begin{table*}[t!]
\centering
\caption{Projector-level efficiency under practical backbone feature settings. Dual-MLP and Expanding-MLP are initialized with matched first-session projector capacity. Inference and training latency are reported as mean $\pm$ standard deviation over repeated measurements.}
\begin{tabular}{llrrr}
\toprule
\textbf{Setting} & \textbf{Method} & \textbf{Infer Lat. (ms)} & \textbf{Train Lat. (ms)} & \textbf{Peak Mem. (MB)} \\
\midrule
\multirow{3}{*}{\makecell[l]{CIFAR, ResNet-12, $640{\times}2{\times}2$}}
& Dual-MLP & $0.5523{\pm}0.0046$ & $3.2920{\pm}0.0554$ & 49.73 \\
& Mamba-FSCIL & $1.3393{\pm}0.0052$ & $7.8944{\pm}0.1040$ & 51.72 \\
& Expanding-MLP-S8 & $2.8722{\pm}0.0005$ & $20.1323{\pm}0.8374$ & 221.20 \\
\midrule
\multirow{3}{*}{\makecell[l]{miniImageNet, ResNet-12, $640{\times}5{\times}5$}}
& Dual-MLP & $0.6103{\pm}0.0007$ & $3.5772{\pm}0.0313$ & 52.10 \\
& Mamba-FSCIL & $1.4692{\pm}0.0803$ & $8.0009{\pm}0.0940$ & 82.50 \\
& Expanding-MLP-S8 & $2.6871{\pm}0.0114$ & $22.0062{\pm}0.3349$ & 223.58 \\
\midrule
\multirow{3}{*}{\makecell[l]{CUB, ResNet-18, $512{\times}7{\times}7$}}
& Dual-MLP & $0.6133{\pm}0.0008$ & $3.4437{\pm}0.0384$ & 41.32 \\
& Mamba-FSCIL & $1.5182{\pm}0.0381$ & $7.7552{\pm}0.1460$ & 89.51 \\
& Expanding-MLP-S10 & $3.2992{\pm}0.0116$ & $23.9534{\pm}0.0700$ & 198.77 \\
\midrule
\multirow{3}{*}{\makecell[l]{CUB, ViT/Swin, $768{\times}14{\times}14$}}
& Dual-MLP & $0.6764{\pm}0.0011$ & $3.9239{\pm}0.0479$ & 150.08 \\
& Mamba-FSCIL & $2.7473{\pm}0.0085$ & $9.9209{\pm}0.0124$ & 407.90 \\
& Expanding-MLP-S10 & $3.6331{\pm}0.0129$ & $24.3157{\pm}0.1979$ & 695.79 \\
\bottomrule
\end{tabular}
\label{tab:response_backbone_settings}
\end{table*}

\begin{figure*}[t!]
\centering
\subfloat[Params]{
    \includegraphics[width=0.24\linewidth]{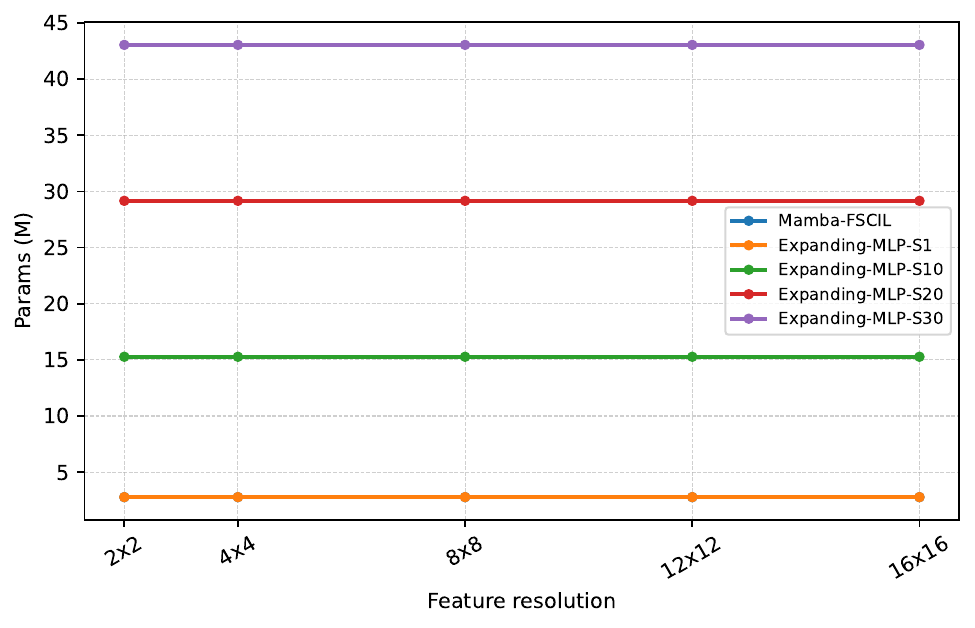}
}
\subfloat[Inference latency]{
    \includegraphics[width=0.24\linewidth]{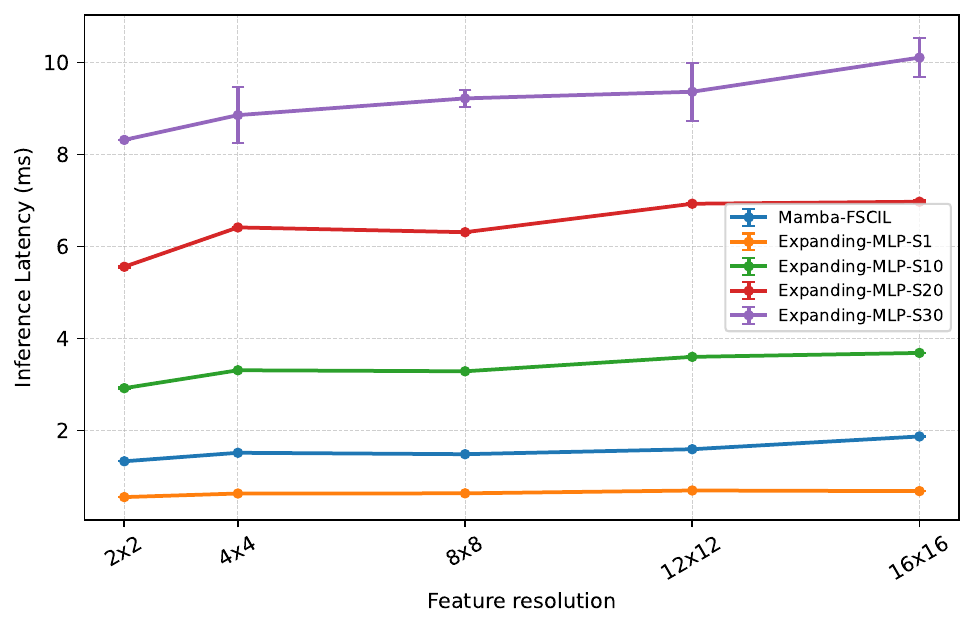}
}
\subfloat[Training latency]{
    \includegraphics[width=0.24\linewidth]{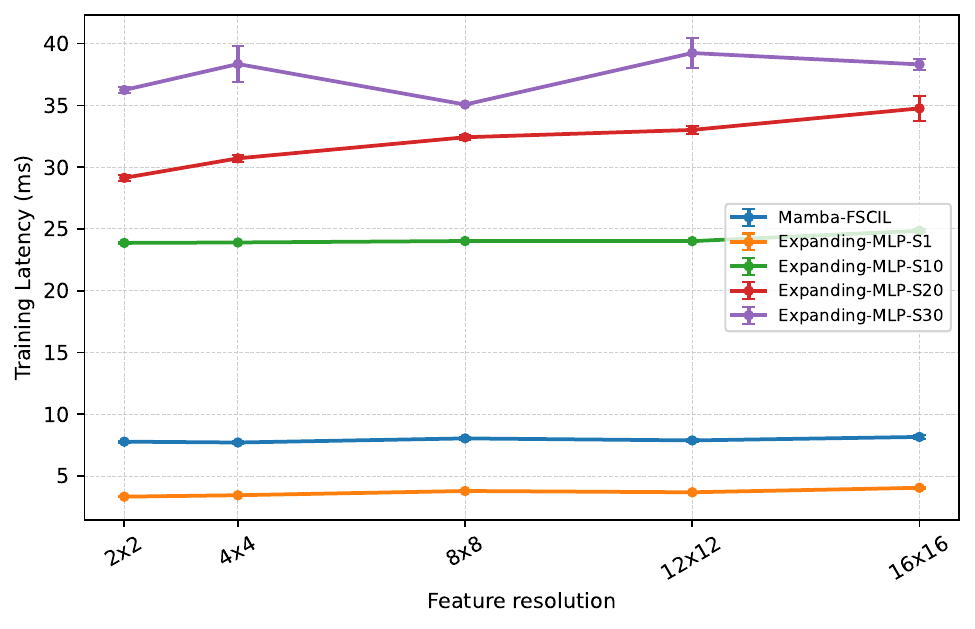}
}
\subfloat[Peak training memory]{
    \includegraphics[width=0.24\linewidth]{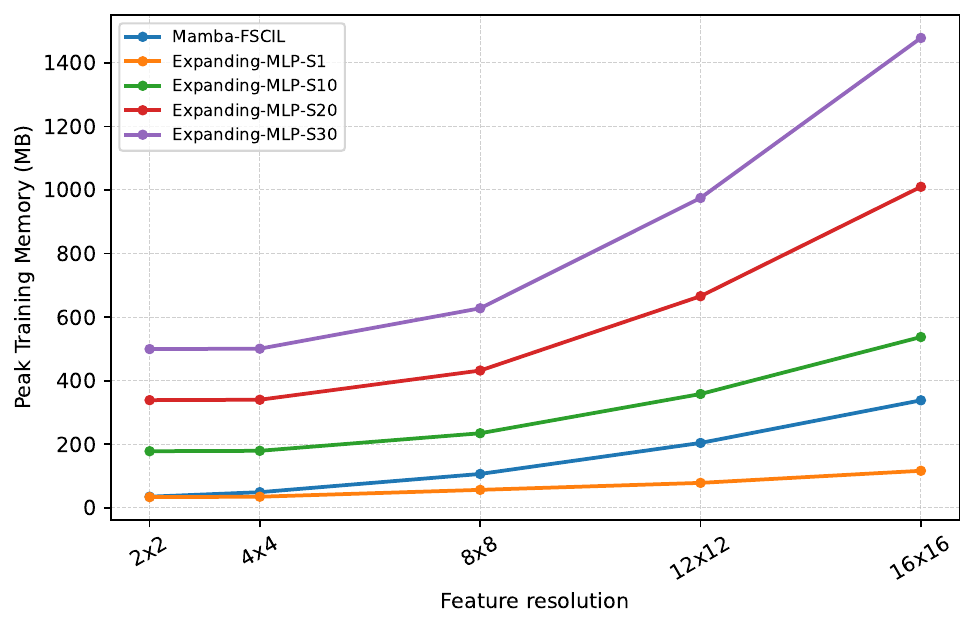}
}
\caption{Resolution/token-length scaling curves of different projectors.}
\label{fig:response_resolution_scaling}
\end{figure*}

\begin{figure*}[t!]
\centering
\subfloat[Params]{
    \includegraphics[width=0.24\linewidth]{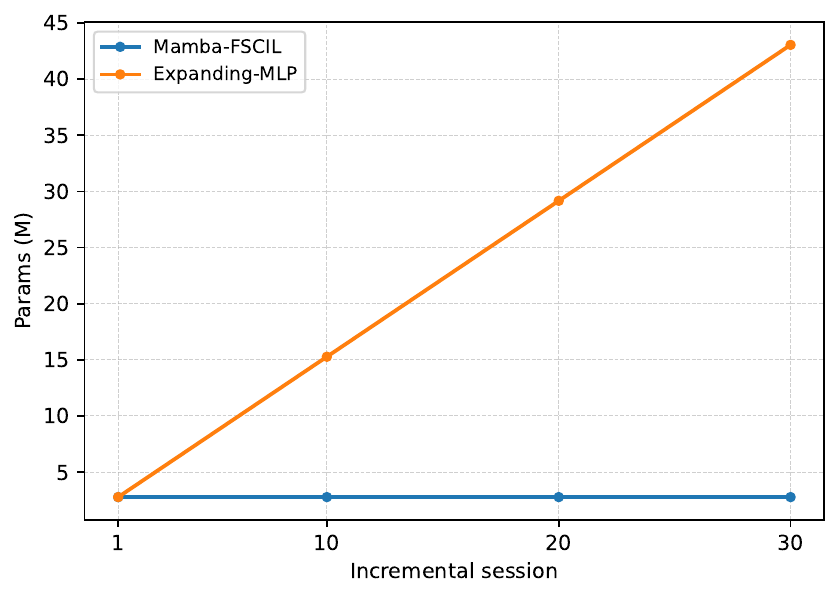}
}
\subfloat[Inference latency]{
    \includegraphics[width=0.24\linewidth]{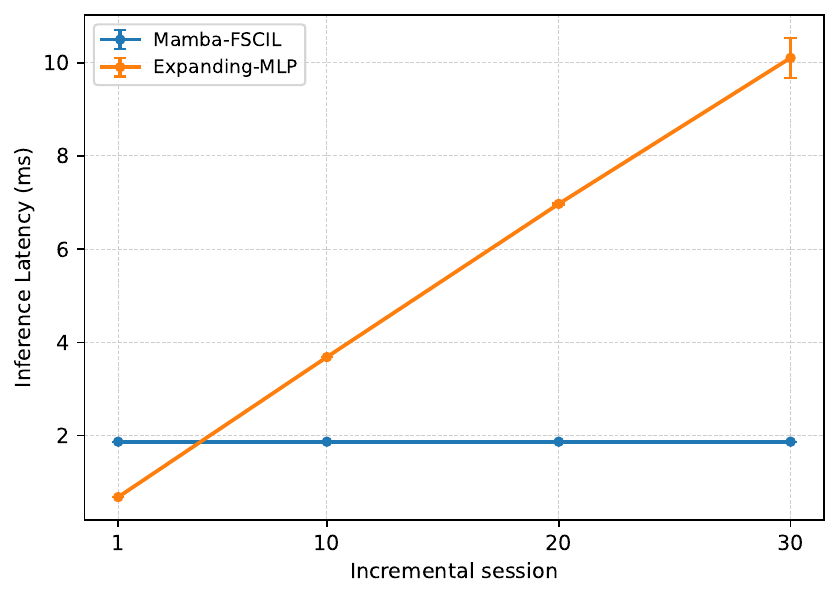}
}
\subfloat[Training latency]{
    \includegraphics[width=0.24\linewidth]{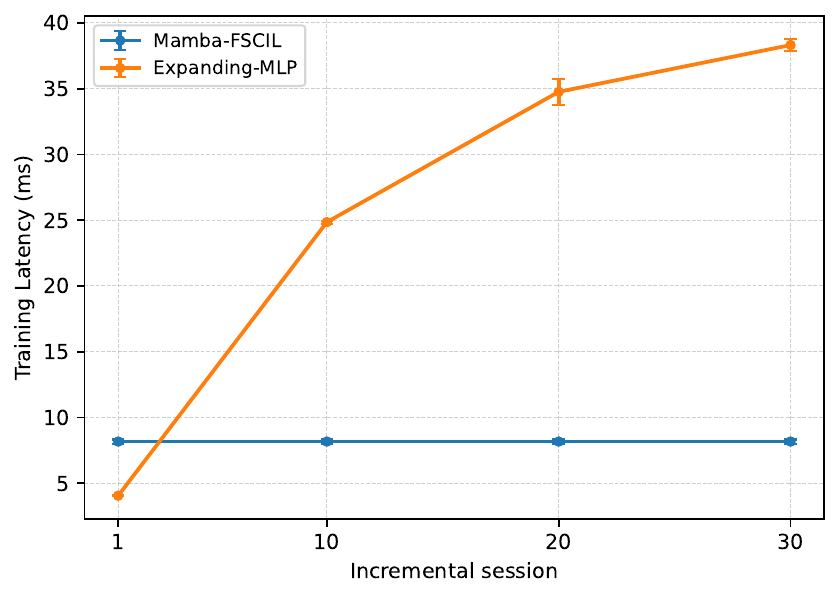}
}
\subfloat[Peak training memory]{
    \includegraphics[width=0.24\linewidth]{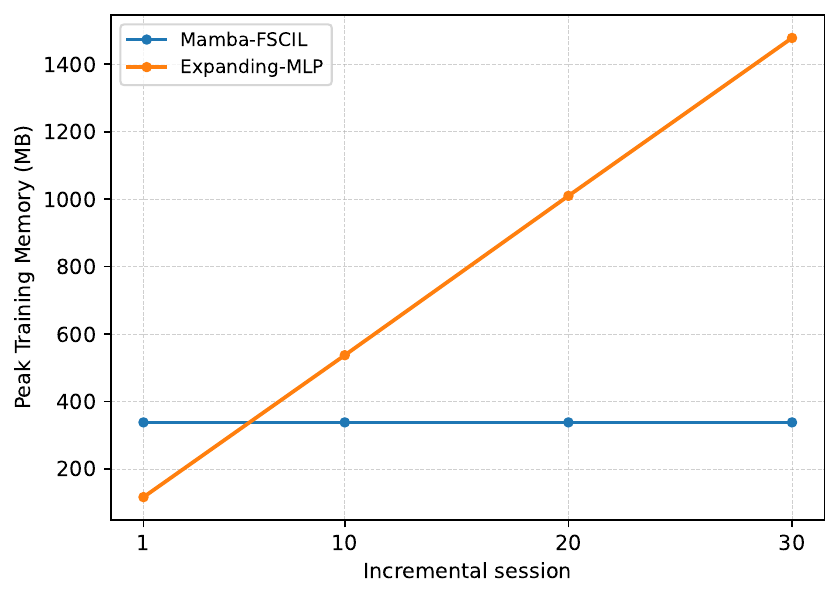}
}
\caption{Session scaling curves at $16{\times}16$ feature resolution.}
\label{fig:response_session_scaling}
\end{figure*}

We use two baselines to separate different sources of computational cost:
\begin{itemize}
    \item \textbf{Dual-MLP (static baseline)} is the fixed-capacity MLP counterpart of Mamba-FSCIL. It keeps the same spatial feature-map input, dual-branch configuration, and session protocol, but replaces the Selective SSM blocks with MLP blocks under a matched projector capacity. This comparison measures the practical overhead of input-dependent selective scanning relative to an MLP projector with the same overall structure.
    \item \textbf{Expanding-MLP (dynamic baseline)} is an expansion-based projector. It starts from the same matched first-session projector capacity as Dual-MLP and Mamba-FSCIL, and then adds one same-width incremental branch for each new session. This baseline is not intended to reproduce a specific prior method. Instead, it isolates the computational effect of session-wise branch accumulation under the same backbone features, feature resolution, branch width, and profiling protocol.
\end{itemize}

We first profile projector efficiency under the feature configurations used in our benchmark experiments. Specifically, we evaluate CIFAR / ResNet-12 with 640-channel $2{\times}2$ features, miniImageNet / ResNet-12 with 640-channel $5{\times}5$ features, CUB / ResNet-18 with 512-channel $7{\times}7$ features, and ViT/Swin-style features with 768-channel $14{\times}14$ resolution. For CIFAR and miniImageNet, where the incremental protocol contains 8 sessions, Mamba-FSCIL is compared with Expanding-MLP-S8. For CUB and ViT/Swin settings, where the protocol contains 10 sessions, Mamba-FSCIL is compared with Expanding-MLP-S10.

As shown in Tab.~\ref{tab:response_backbone_settings}, Mamba-FSCIL has higher projector-level latency than Dual-MLP. This is expected because selective scanning performs input-dependent token processing, whereas MLP blocks are input-agnostic feed-forward operators. This comparison quantifies the overhead introduced by selective scanning under a matched projector structure.

Compared with the corresponding last-session Expanding-MLP baseline, Mamba-FSCIL consistently reduces inference latency, training latency, and peak training memory across all evaluated settings. Across the four settings, the corresponding last-session Expanding-MLP baseline has $1.32{\times}$--$2.17{\times}$ higher inference latency, $2.45{\times}$--$3.09{\times}$ higher training latency, and $1.71{\times}$--$4.28{\times}$ higher peak training memory than Mamba-FSCIL, where each ratio is computed as the Expanding-MLP cost divided by the Mamba-FSCIL cost under the same feature setting. These results indicate that reusing a selective SSM projector avoids the cumulative latency and memory growth caused by session-wise branch expansion.

\subsubsection{Scaling with Feature Resolution and Session Count}
We analyze how projector cost changes with feature resolution/token length and with the number of incremental sessions.

First, we vary the feature resolution from $2{\times}2$ to $16{\times}16$, corresponding to token lengths from $4$ to $256$, and compare Mamba-FSCIL with Dual-MLP and Expanding-MLP under different expansion stages. Expanding-MLP-S20 and Expanding-MLP-S30 are included as long-horizon stress-test settings beyond the standard benchmark protocols to illustrate the effect of session-wise branch accumulation.

As shown in Fig.~\ref{fig:response_resolution_scaling}~(a), when the channel dimension and branch width are fixed, Mamba-FSCIL keeps a constant parameter count as the feature resolution increases. Figs.~\ref{fig:response_resolution_scaling}~(b)--(d) show that its latency and memory increase with longer token sequences because selective scanning processes more spatial tokens. In contrast, Expanding-MLP has larger parameter counts when more session branches are accumulated, and the S20/S30 stress-test settings further illustrate how branch accumulation raises the overall cost under different token lengths.

Second, we fix the feature resolution to $16{\times}16$ and vary the number of incremental branches in Expanding-MLP up to Session 30. This setting isolates the effect of session count from the effect of feature resolution.

As shown in Fig.~\ref{fig:response_session_scaling}~(a), when the feature resolution is fixed, Mamba-FSCIL keeps a constant parameter count across sessions by reusing the same dynamic selective-SSM branch. Its input-dependent selective scanning adapts the projector to different session features without introducing session-specific branches. In contrast, Expanding-MLP adds one branch per session; although it can be lightweight in early sessions, its cost accumulates in later sessions such as S10, S20, and S30. Figs.~\ref{fig:response_session_scaling}~(b)--(d) show that this accumulation increases inference latency, training latency, and peak training memory.

\subsection{Theoretical Analysis of Architectural Differences}
\label{supp:theoretical_analysis}
In this section, we provide a detailed mathematical analysis comparing the parameter update mechanisms of static architectures (MLP, Transformer, S4, RNN) versus our proposed Dynamic Selective SSM. We discuss a parameter-sharing interference that may arise in static architectures during incremental learning.

\vspace{1mm}
\textbf{(1) Limitations of Static Architectures: Parameter-Sharing Interference.}
Static architectures (MLPs, Transformers, S4, RNNs) rely on \textbf{static weights} shared across all sessions. Mathematically, for a base input $\mathbf{x}_\text{b}$, the output is determined by shared parameters $\theta$. When the model updates $\theta \rightarrow \theta + \Delta\theta$ to minimize the loss on novel classes, the mapping for base classes undergoes a shift:
\[
    f_{\text{static}}~(\mathbf{x}_\text{b}; \theta + \Delta\theta) \neq f_{\text{static}}~(\mathbf{x}_\text{b}; \theta).
\]
Consequently, such updates can introduce interference into established base-class representations and may contribute to catastrophic forgetting. This limitation manifests differently across architectures:

\begin{itemize}
    \item \textbf{MLP Projector:} A standard MLP defines a fixed mapping $\mathbf{y} = f_{\text{MLP}}~(\mathbf{x}; \theta) = W_L \sigma(\dots \sigma(W_1 \mathbf{x})\dots)$. The weights $\theta = \{W_1, \dots, W_L\}$ are static matrices. Optimizing them to activate for new classes creates a geometric conflict if one simultaneously attempts to suppress activations for base classes within the same feature space.
    \item \textbf{Transformer Projector:} Self-attention relies on projections $\mathbf{Q}=\mathbf{X}W_Q, \mathbf{K}=\mathbf{X}W_K, \mathbf{V}=\mathbf{X}W_V$. Since the weight matrices $W_Q, W_K, W_V$ are \textbf{static and shared}, updating them to capture novel classes may also alter the feature mapping for base classes. Because the Attention map is computed via token-to-token correlations ($A = \text{softmax}(QK^\top)$), even a small drift in the projected base features ($Q, K$) can affect the learned attention patterns.
    \item \textbf{Static SSM (S4):} S4 relies on \textbf{fixed, input-invariant} system matrices $(\mathbf{A}, \mathbf{B}, \mathbf{C})$. This means the same state evolution rules ($h_{t+1} = \mathbf{A}h_t + \mathbf{B}x_t$) are applied to every sample regardless of its class. Consequently, updating these matrices to capture novel-class dynamics may change the global processing logic. Since the model cannot dynamically switch its parameters based on the input, modifications for novel classes may interfere with the behavior learned for base classes.
    \item \textbf{GRU / LSTM:} Although the hidden state evolves over time  (e.g., $\Theta_{\text{GRU}}$ in $h_t = \mathrm{GRUCell}~(\mathbf{x}_t, h_{t-1}; \Theta_{\text{GRU}})$), the mechanisms governing state transitions (i.e., the weights $\Theta_{\text{GRU}}$) are static and globally shared. This imposes a single set of temporal processing rules for all inputs. Consequently, updating these gating weights to capture novel-class dependencies may also affect the state-transition logic for base-class inputs, potentially disturbing the memory dynamics established during the base session.
\end{itemize}

\vspace{1mm}
\noindent \textbf{(2) Selective SSM: Input-Dependent Operator Generation.}
In contrast, Selective SSM uses shared parameter-generation functions to produce input-conditioned operators. Its key property is \textbf{input-dependent operator generation}, where the system matrices are generated as functions of the input: $(\mathbf{B}, \mathbf{C}, \mathbf{\Delta}) = (f_B(\mathbf{x}), f_C(\mathbf{x}), f_\Delta(\mathbf{x}))$.
This allows the incremental branch to generate different state-space operators for different inputs using the same parameter-generation functions:
\[
(\mathbf{B}_\text{b}, \mathbf{C}_\text{b}, \mathbf{\Delta}_\text{b}) \neq (\mathbf{B}_\text{n}, \mathbf{C}_\text{n}, \mathbf{\Delta}_\text{n}),
\]
where the subscript $b$ denotes base inputs and $n$ denotes novel inputs. This structure provides additional flexibility for differentiating the transformations applied to base and novel inputs. Together with our \textbf{Class-Sensitive Regularization}, this input-conditioned operator generation encourages two desirable behaviors:
\begin{itemize}
    \item \textbf{Reduced base-class interference:} The suppression loss encourages limited incremental-branch responses for base inputs, reducing perturbation to the frozen stability anchor.
    \item \textbf{Novel-class adaptation:} The separation loss encourages more distinct generated operators for base and novel inputs, helping the incremental branch form discriminative transformations for novel classes.
\end{itemize}
This analysis provides an explanatory perspective for why input-conditioned operator generation can improve the stability--plasticity trade-off observed in our experiments.

\subsection{Granular Analysis of Dynamic Adaptation}
\label{supp:granular_analysis}
\begin{figure}[t!]
    \centering
    \includegraphics[width=1\linewidth]{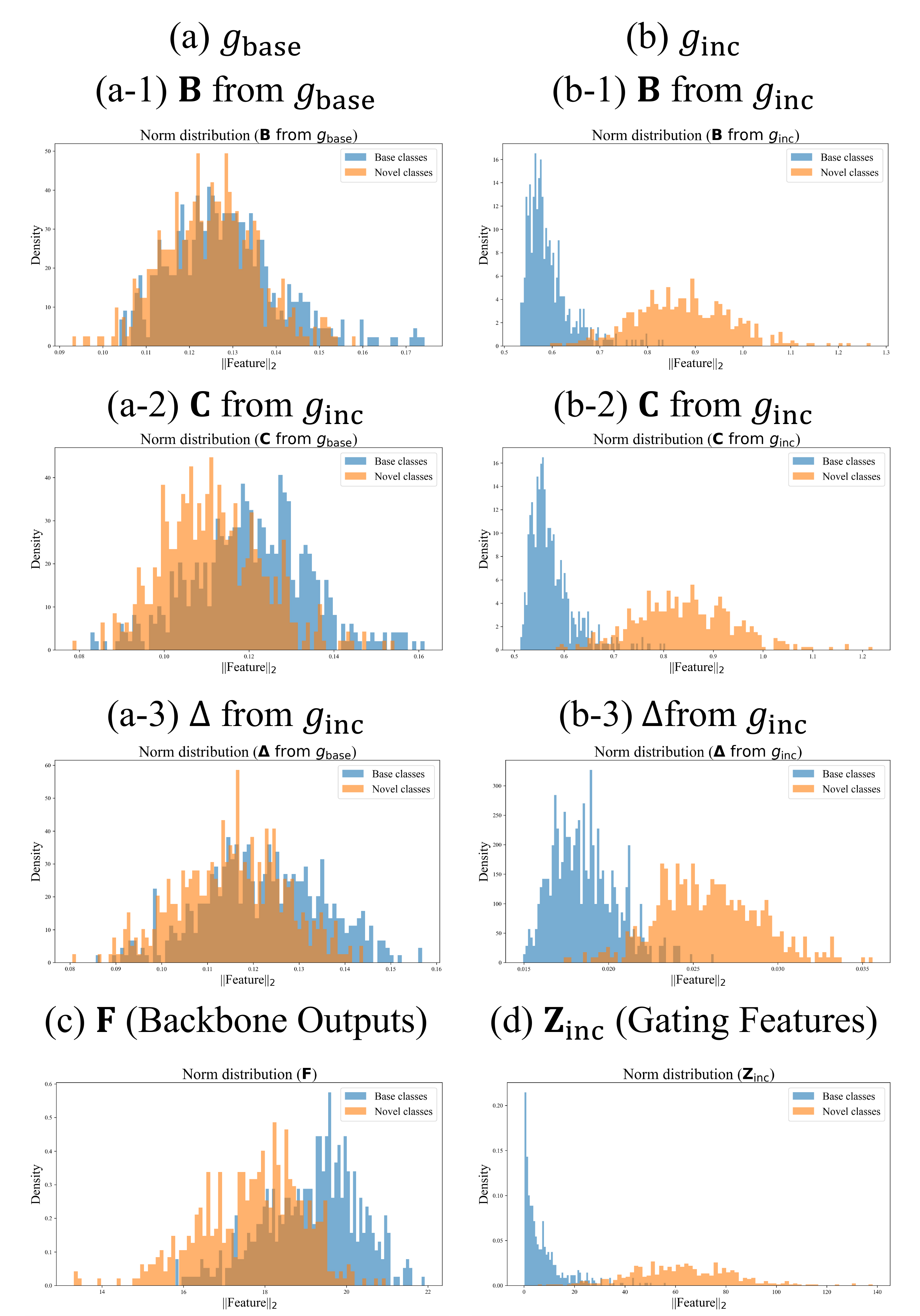}
    \vspace{-4mm}
    \caption{Norm distributions of dynamic SSM parameters and intermediate features. Comparison of distributions for base-class inputs (black) and novel-class inputs (orange). (a) Parameters from the frozen Base Branch ($g_{\text{base}}$). (b) Parameters from the Incremental Branch ($g_{\text{inc}}$). (c) Backbone output features $\mathbf{F}$. (d) Gating features $\mathbf{Z}_{\text{inc}}$.
    \label{fig:vis_norm}}
\end{figure}

\begin{figure}[t!]
    \centering
    \includegraphics[width=1\linewidth]{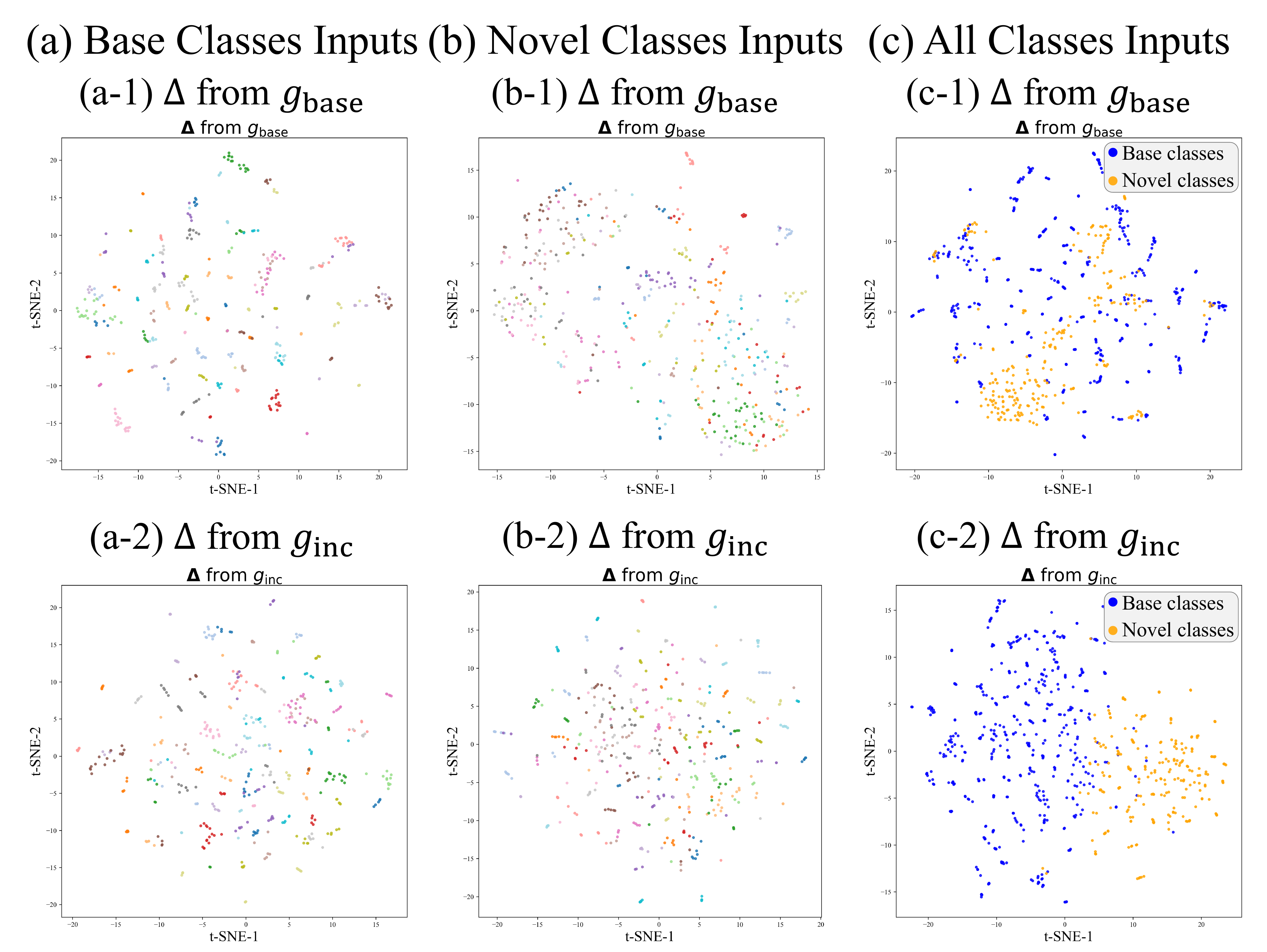}
    \vspace{-4mm}
    \caption{Distribution of input-conditioned SSM parameters $\mathbf{\Delta}$ in the final session. Row 1: parameters generated by the frozen $g_{\text{base}}$.  
    Row 2: parameters generated by the trainable $g_{\text{inc}}$.  
    Columns show: (a) base-class inputs, (b) novel-class inputs, (c) all inputs.
    }
    \label{fig:vis_delta}
\end{figure}

\begin{figure}[t!]
    \centering
    \includegraphics[width=1\linewidth]{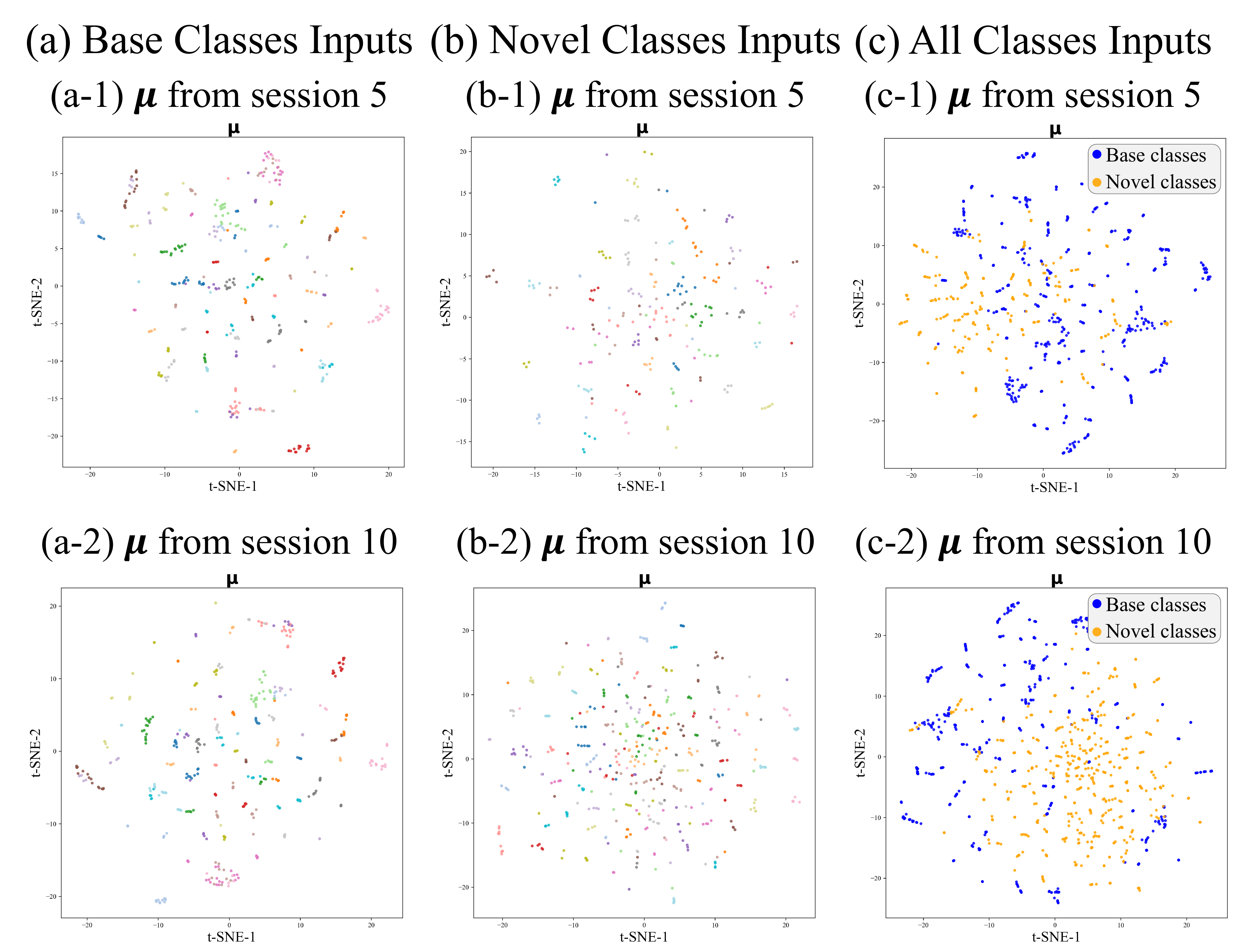}
    \vspace{-4mm}
    \caption{\textbf{Visualization of the final output features ($\boldsymbol{\mu}$)} on CUB-200. Comparison between Session 5 (Top) and Session 10 (Bottom). \textbf{(a)} Base class inputs. \textbf{(b)} Novel class inputs. \textbf{(c)} All classes combined.}
    \label{fig:vis_output}
\end{figure}

Unlike static architectures such as MLPs or CNNs, which apply a fixed transformation to all inputs during inference, Mamba enables dynamic adaptation via \textit{input-dependent parameter generation}. To better explain how Mamba-FSCIL maintains a balance between stability and plasticity, we perform a detailed granular analysis of the Dual Selective SSM Projector. This analysis involves visualizing activation magnitudes (Norm Distributions, Fig.~\ref{fig:vis_norm}), state-space parameters (Parameter Manifold, Fig.~\ref{fig:vis_delta}), and feature distributions (Fig.~\ref{fig:vis_output}). These visualizations provide qualitative evidence for the class-sensitive dynamic adaptation behavior of Mamba-FSCIL.

\vspace{1mm} \noindent \textbf{(1) Norm Distribution in $g_{\text{base}}$ and $g_{\text{inc}}$: Selective Activation for Plasticity.}
We examine how the Class-Sensitive Selective Scan Mechanism governs the activation of the incremental branch ($g_{\text{inc}}$) using the CUB test set and a ViT-B (CLIP) backbone. The $L_2$ norm densities for various components are plotted as follows:
\begin{itemize}
    \item \textbf{Backbone $f$ and $g_{\text{base}}$ provide stable representations with limited base--novel differentiation:} The norm distributions of both the backbone features (Fig.~\ref{fig:vis_norm}~(c)) and the frozen $g_{\text{base}}$ parameters (Fig.~\ref{fig:vis_norm}~(a)) show significant overlap between base and novel inputs. This suggests that the frozen base branch applies relatively similar transformations to base and novel inputs, which is consistent with its role as a stable representation anchor.
    \item \textbf{Incremental branch ($g_{\text{inc}}$) selectively activates for novel classes:} In contrast, the incremental branch shows more separated norm distributions for parameters ($\mathbf{B}, \mathbf{C}, \mathbf{\Delta}$) (Fig.~\ref{fig:vis_norm}~(b)) and gating features $\mathbf{Z}_{\text{inc}}$ (Fig.~\ref{fig:vis_norm}~(d)), with significantly higher norms for novel-class inputs (orange) and lower norms for base-class inputs (black). This observation supports the effect of the suppression loss ($\mathcal{L}_{\text{supp}}$), which reduces the response of $g_{\text{inc}}$ for base-class inputs to preserve stability ($\mathcal{L}_{\text{supp, base}}$) while allowing stronger responses for novel-class inputs ($\mathcal{L}_{\text{supp, novel}}$).
\end{itemize}

\vspace{1mm} \noindent \textbf{(2) Parameter-Space Comparison: Input-Dependent Operator Generation.}
To better understand how Mamba adapts its processing logic, we visualize the t-SNE embeddings of the input-conditioned time-scale parameter $\mathbf{\Delta}$ (Fig.~\ref{fig:vis_delta}). The parameter $\mathbf{\Delta}$ plays a critical role in Selective SSMs by controlling the discretization of continuous system parameters ($\mathbf{A}, \mathbf{B} \rightarrow \overline{\mathbf{A}}, \overline{\mathbf{B}}$ in Eq.~\ref{equ:ssm_param_discretization}). As an input-dependent time-scale parameter, $\mathbf{\Delta}$ affects the discretization and hidden-state update dynamics, thereby influencing how current inputs are integrated into the state. Unlike static weights in MLPs or transformers, $\mathbf{\Delta}_t = f_\Delta(x_t)$ is dynamically generated, and its distribution provides a useful diagnostic for how the model adapts its state-space dynamics to different inputs:

\begin{itemize}
\item \textbf{Frozen $g_{\text{base}}$ remains computationally dynamic:} \textbf{For base-class inputs} (Fig.~\ref{fig:vis_delta}~(a-1)), the generated operators form relatively compact clusters, indicating stable transformation patterns learned during the base session.
\textbf{For novel-class inputs} (Fig.~\ref{fig:vis_delta}~(b-1)), the generated parameters of $g_{\text{base}}$ still reflect visual similarities among novel-class inputs. However, as shown in the combined view (c-1), the parameter manifolds for base and novel classes overlap, indicating that while $g_{\text{base}}$ captures general features, the base branch alone provides limited discrimination between base and novel concepts.
\item \textbf{Incremental branch ($g_{\text{inc}}$) adapts for novel classes:}  Guided by $\mathcal{L}_{\text{supp}}$ and $\mathcal{L}_{\text{sep}}$, the incremental branch produces more class-sensitive operator distributions (Fig.~\ref{fig:vis_delta}~(b-2 and c-2)):  
\textbf{For novel-class inputs} (Fig.~\ref{fig:vis_delta}~(b-2)), $g_{\text{inc}}$ generates \textbf{more compact and separated parameter clusters} compared to $g_{\text{base}}$ in (b-1), suggesting that it learns input-conditioned operators that are more adapted to novel-class features.
\textbf{For base-class inputs} (Fig.~\ref{fig:vis_delta}~(c-2)), the generated operators (black) are located separately from novel-class operators (orange).  This separation is consistent with reduced interference between base and novel inputs, supporting the role of the suppression loss in reducing the response of $g_{\text{inc}}$ for base-class inputs, maintaining stability while allowing activation for novel classes, capturing new feature shifts. Additionally, the separation loss ($\mathcal{L}_{\text{sep}}$) encourages novel-class operators to be more separated from base-class operators, promoting more distinct transformation patterns for novel classes while reducing interference with base-class representations.
\end{itemize}

\vspace{1mm} \noindent \textbf{(3) Representation-Space Dynamics: Evolution of Final Outputs Across Sessions.}  
Finally, we visualize the t-SNE embeddings of the fused representation $\boldsymbol{\mu} = \boldsymbol{\mu}_\text{iden} + \boldsymbol{\mu}_\text{base} + \boldsymbol{\mu}_\text{inc}$ in Fig.~\ref{fig:vis_output} to show how internal mechanisms affect the final feature space:
\begin{itemize}
    \item \textbf{Discriminative Structure (Plasticity):} As shown Fig.~\ref{fig:vis_output}~(c-1, c-2), the novel classes (orange) form relatively compact clusters with visible separation from the base classes (black). The adaptation for novel classes is accompanied by relatively stable base-class clusters, consistent with the suppressed response of $g_{\text{inc}}$ on base-class inputs.
    \item \textbf{Topological Consistency (Stability):} Comparing Session 5 (Fig.~\ref{fig:vis_output}~c-1) and Session 10 (Fig.~\ref{fig:vis_output}~c-2), the structure of base-class clusters remains relatively stable as new classes are added. This trend is consistent with the frozen $g_{\text{base}}$ anchor and the suppression of $g_{\text{inc}}$ for base-class inputs.
\end{itemize}

In summary, these visualizations suggest that class-sensitive regularization encourages more separated operator patterns and feature representations for base and novel classes. This provides qualitative evidence that Mamba-FSCIL improves the stability--plasticity balance by modulating both the magnitude and direction of input-conditioned state-space operators.

\end{document}